\documentclass{article}
\usepackage[dvips]{epsfig,graphicx}
\usepackage{times,subfigure,longtable,lscape,amssymb,amsbsy,times,fancyhdr,color,amsmath,latexsym,rotating,hhline,url,multirow} 
\usepackage{float,amsmath,graphicx,latexsym,amsfonts,amssymb,soul,wrapfig,algorithmic,algorithm,dsfont}
\newcommand{\ie}{i.e.,\,}

\newtheorem{theorem}{Theorem}

\newtheorem{remark}{Remark}

\begin{document}
\title{Structural bias in population-based algorithms}
\author{Anna V.~Kononova\thanks{anna.kononova@gmail.com, Department of Computer Science, Heriot-Watt University, Edinburgh, EH14 4AS, UK} \and David W.~Corne\thanks{d.w.corne@hw.ac.uk, Department of Computer Science, Heriot-Watt University, Edinburgh, EH14 4AS, UK} \and Philippe De Wilde\thanks{p.dewilde@kent.ac.uk, School of Computing, University of Kent, Canterbury, CT2 7NZ, UK} \and Vsevolod Shneer\thanks{v.shneer@hw.ac.uk, School of Mathematical and Computer Sciences, Heriot-Watt University, Edinburgh, EH14 4AS, UK} \and Fabio Caraffini\thanks{fabio.caraffini@gmail.com, Centre for Computational Intelligence, School of Computer Science and Informatics, De Montfort University, Leicester LE1 9BH, UK}}
\maketitle

\begin{abstract} Challenging optimisation problems are abundant in all areas of science and industry. Since the 1950s, scientists have responded to this by developing ever-diversifying families of 'black box' optimisation algorithms. The latter are designed to be able to address any optimisation problem, requiring only that the quality of any candidate solution can be calculated via a 'fitness function' specific to the problem. For such algorithms to be successful, at least three properties are required: (i) an effective informed sampling strategy, that guides the generation of new candidates on the basis of the fitnesses and locations of previously visited candidates; (ii) mechanisms to ensure efficiency, so that (for example) the same candidates are not repeatedly visited; (iii) the absence of structural bias, which, if present, would predispose the algorithm towards limiting its search to specific regions of the solution space. The first two of these properties have been extensively investigated, however the third is little understood and rarely explored. In this article we provide theoretical and empirical analyses that contribute to the understanding of structural bias. In particular, we state and prove a theorem concerning the dynamics of population variance in the case of real-valued search spaces and a 'flat' fitness landscape. This reveals how structural bias can arise and manifest as non-uniform clustering of the population over time. Critically, theory predicts that structural bias is exacerbated with (independently) increasing population size, and increasing problem difficulty. These predictions, supported by our empirical analyses, reveal two previously unrecognised aspects of structural bias that would seem vital for algorithm designers and practitioners. Respectively, (i) increasing the population size, though ostensibly promoting diversity, will magnify any inherent structural bias, and (ii) the effects of structural bias are more apparent when faced with (many classes of) 'difficult' problems. Our theoretical result also contributes to the 'exploitation/exploration' conundrum in optimization algorithm design, by suggesting that two commonly used approaches to enhancing exploration -- increasing the population size, and increasing the disruptiveness of search operators -- have quite distinct implications in terms of structural bias. \end{abstract}

\section{Introduction}\label{sect:intro}
Successful implementation of any randomised population-based optimisation algorithm depends on the efficiency of both its sampling component and exploitation of previously sampled information. Among other fields, Evolutionary Computation (EC) \cite{cit:EibenSmith} provides various examples of randomised population-based search strategies. Greatly simplified, any evolutionary computation algorithm is a guided re-sampling strategy where movement of points is directed by its operators assisted by selection criteria based on currently attained values of the objective function. A vast body of research in the field of Evolutionary Computation deals with efficient exploitation of information already contained within the population \cite{cit:Crepinsek2013} while little attention has been paid to investigation of whether or not a specific combination of algorithmic operators is actually capable of reaching all parts of the search space efficiently. This paper attempts to draw attention to this issue and starts to investigate the latter question.

Inspection of recent literature \cite{cit:Qin2009}, \cite{cit:deOca2009}, \cite{cit:Peng2010}, \cite{cit:Zhang2014} confirms the presence of a tendency to (over-)complicate both the design of individual algorithmic operators and the logic of their assembly, counter to the rationale of the well-known Occam's razor\footnote{Originally attributed to William of Occam, reformulated by Betrand Russell as "Whenever possible, substitute constructions out of known entities for inferences to unknown entities." \cite{cit:Russell1924}}, sometimes to such a degree that the end result turns out to be intractable. Researchers seem regularly to be swayed by an attraction towards "multiplying entities beyond necessity". We suggest that a materially greater contribution to the understanding of population-based algorithms and their design can be obtained via 'going back to basics'. More specifically, the great majority of optimisation algorithms fall within a class of generate-and-test methods, iteratively alternating between these two components until a termination criterion is met. Ideally, the generating/sampling component of such methods should have the following characteristics \cite{cit:Winston1984}:
\begin{enumerate}
\item future samples should be biassed by information obtained from previously visited points, \ie the algorithm should be \textit{informed}, 
\item future samples should be previously unvisited samples \ie the algorithm should be \textit{non-redundant},
\item every solution in the search space should be equally accessible \ie the algorithm should be \textit{complete}.
\end{enumerate}

It is worth noting our use of the phrase \textit{equally accessible} within the 'completeness' characteristic. Often, for example, algorithm designers may be subconsciously swayed by the fact that the randomness of the initialisation process means that every part of the search space is \textit{reachable}, and hence feel no further need to consider this characteristic. Reachability and completeness (the way we define it here) are however very different. For example, if a stochastic hill-climbing algorithm includes a uniform random initialisation in $\mathbb{R}^n$, then all points are reachable, however if the perturbation operator is designed to add only integer-valued vectors, then there are extreme variations in the accessibilities of different points in the space.  

Clearly, evolutionary computation methods build richly upon their 'generate-and-test' backbone architecture. However the above guidelines remain valid, and, in practice, they translate well into rules for algorithm design. The first two properties -- informedness and non-redundancy in the sampling process -- have been extensively researched, each from a variety of viewpoints. To some extent, however, contributions related to these two properties have appeared in diverse and unconnected literature, using varying terminology, and there remains a need to creatively assimilate their findings.

For example, with sufficient imagination one can see that the \textit{informedness} property is closely linked to the concepts of exploration, exploitation, and their balance, which is considered to be primary in the behaviour of evolutionary algorithms (EAs), as examples of stochastic "generate-and-test" methods \cite{cit:Eiben1998}. Exploration and exploitation are fundamental for evolutionary optimisation \cite{cit:EibenSmith} but surprisingly, several decades after the first examples of EAs have been proposed, they still lacked even a proper definition. Over the following years, a lot of research has been carried out in this direction - the latest survey of results can be found in \cite{cit:Crepinsek2013}. The current consensus definitions consider \textit{exploration} as the process of visiting entirely new regions of a search space whilst \textit{exploitation} as the process of visiting those regions of a search space within the neighbourhood of previously visited points \cite{cit:Crepinsek2013}.

The second characteristic, \textit{non-redundancy}, has been investigated under the guise of 'non-revisiting' algorithms. Inspired by ideas from Tabu search \cite{cit:Glover1989}, \cite{cit:Glover1990}, basic evolutionary algorithms have been extended to ensure the non-revisiting property \cite{cit:Yuen2007}, \cite{cit:Yuen2009_1}, \cite{cit:Yuen2009_2}. Another direction of research into the non-redundancy property is the study of diversity management in evolving populations. Diversity in populations can refer to differences in solutions in either the values of coordinates ('genotypic' diversity) or the objective function ('phenotypic' diversity). To date, no single measure exists which can suitably characterise diversity in the face of all kinds of problems and search logics \cite{cit:Neri2012}. The situation is further complicated by the fact that a diverse population offers benefits at some stages of evolutionary process (helps avoid premature convergence to local optima) and creates obstacles in others (impedes exploitation) \cite{cit:Burke2004}. The most popular diversity-preserving mechanisms include \cite{cit:Gupta2012} niching, crowding, restricted mating, sharing, multiploidy, elitism, injection, alternative replacement strategies \cite{cit:Lozano2008} and fitness uniform selection \cite{cit:FUSS}.

Much promising research is also carried out that tries to explore the connections between \textit{informedness} and \textit{non-redundancy}, stemming from the fact that exploration of the search space is only possible if populations are diverse enough \cite{cit:Crepinsek2013}. However, different amounts of exploration and exploitation are needed for different optimisation problems. Currently there are no accepted techniques for direct measurement of this balance; it can only be noisily sensed via a proxy (such as 'level of diversity'). Feedback from online monitoring of such a proxy, if it is suitably computationally efficient, can then be used to dynamically tilt the exploration-exploitation balance \cite{cit:Crepinsek2013}, \cite{cit:Kononova2007}, \cite{cit:Briscoe2011}.

Meanwhile, in recent decades, the third important property of the generating component, which we can call \textit{completeness}, has been treated as an obscure topic and largely ignored by modern research efforts - the latest reference to this issue, \cite{cit:Eiben1991}, briefly deals with the accessibility of a solution through evolution as a necessary condition for reaching the optimum by a genetic algorithm in the discrete case. To the best of our knowledge, no research related to this property in the real-valued case has been published until a rather recent revival of interest to a closely related topic. A popular belief \cite{cit:Gehlhaar1996}, \cite{cit:Angeline1998} -- that many population-based algorithms tend to perform better when the true optimum is located at or near the centre of the initialisation region -- has recently sparked the interest of a small group of researchers who have explored this phenomenon in the context of specific variants of PSO. When the initialisation region is centred around the origin, this phenomenon is referred to as 'origin-seeking bias'. To date, there is little or no evidence to support the presence of such a bias in the general case. Having investigated the effects of modifying the search domains of three benchmark problems on results produced by an unusual variant of PSO, the authors of \cite{cit:Monson2005} concluded the presence of origin-seeking bias in their specific algorithm/problem scenarios, and suggested that their results could be generalised towards all population-based methods. However these results were later disputed and have been largely dismissed \cite{cit:Kennedy2007}. As regards theoretical analysis of the movement of particles in a PSO swarm, a study of particle trajectories \cite{cit:vandenBergh2006} reveals that under certain conditions every particle converges to a stable point defined by its personal and global best positions, with weights determined by the acceleration coefficients. Experimental results also suggest that, for the well-known \textit{sphere} objective function, the movement of particles is influenced by the direction of the coordinate axis which potentially makes the algorithm sensitive to rotation of the objective function \cite{cit:Janson2007}. Further theoretical analysis \cite{cit:Spears2010} indicates that there is an angular bias in the core PSO algorithm which consists of two parts. The first part, skew, pushes particles towards bearings parallel with the diagonals, meanwhile the second part, spread, indicates that diagonal directions are highly unstable. The combination of the two parts creates a PSO bias that favours particle bearings that are aligned with the coordinate axes. The latest publication on this topic \cite{cit:Davarynejad2014} extends the work from \cite{cit:Monson2005} and proposes a metric for centre-seeking and initialisation biases based on multiple re-runs of the algorithm in modified domains. 

The majority of authors implicitly suppose their algorithms fare well in the ability to potentially cover the whole search domain. Put another way, researchers tend to take for granted the property of 'completeness'. Such quiet confidence about this property probably stems from the perception that, given the stochastic nature of standard initialisation methods and standard operators, all parts of the search space are \textit{reachable}. However, this ignores the prospect that reachability may actually be highly non-uniform across the search space.  Results presented in this paper demonstrate that even the most commonly used algorithms exhibit inherent preferences towards certain parts of the search domain. We refer to this preference as the \textit{structural bias} of the algorithm.

\begin{figure*} \centering
\includegraphics[width=0.4\textwidth]{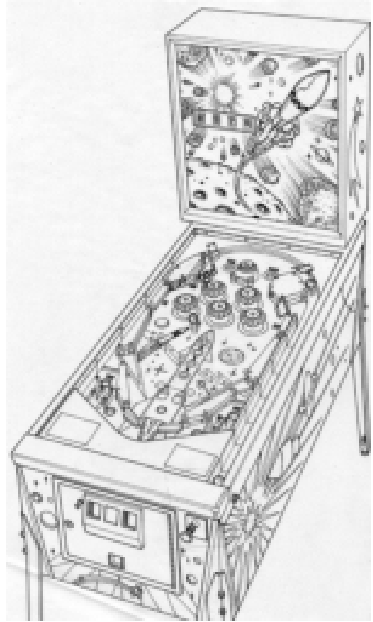}
\caption{A typical pinball machine}\label{fig:pbm}
\end{figure*}%

To help illustrate the concept of structural bias, it may be helpful to imagine a pinball machine where a player has to operate a system of mechanical devices to allow a ball to stay on the game surface as long as possible before hitting the drain, see Fig \ref{fig:pbm}. We can consider the whole system -- the pinball machine and the actions of the player -- to represent an algorithm, while the ball represents a solution travelling around the search space. We can conceptualise multiple games on such a machine, overlapped in time, to represent the case of an algorithm that maintains a population of solutions. In the ideal case, the population should be able to access the entire game surface. A population-based algorithm exhibiting structural bias is then replicated by an overlapping in time of such machines which are unfairly tilted at some angle. Clearly, even in this case the actions of the player have a certain effect on the movement of the balls. However, due to gravity, the ball ends up constantly rolling to the lower side of the machine \ie exhibiting a certain preference and limiting the overall coverage of the game surface.

The remainder of the paper is arranged as follows. In Section \ref{sect:sbias} we argue that studying the structural bias that may be inherent in an algorithm can be facilitated by decoupling the effects of the search landscape from the algorithmic operations. This section then introduces and analyses a test function $f_0$, which allows such decoupling. Section \ref{sect:numres} takes the function $f_0$ and uses it as a 'structural bias probe', presenting several experiments in which we investigate whether structural bias seems to be present in typical designs of a genetic algorithm and a particle swarm optimisation algorithm. Visualisations of the results of our experiments in this section offer evidence that structural bias is indeed present in these algorithms, and sensitive to parameters such as population size. This section ends with remarks concerning pseudorandom number generators, in particular those used in our implementation, and offers evidence -- following further empirical investigation -- that supports the view that our observations in the previous section were uncontaminated by artefacts of the pseudorandom generator. In Section \ref{sect:sbiasGA} we turn to a theoretical investigation, which considers a simplified genetic algorithm process (nevertheless capturing well the behaviour of a typical genetic algorithm on $f_0$). The theorem proved in this section shows that the considered (simplified but otherwise standard) algorithm design will, under certain but unexceptional conditions, induce a continual reduction in sample variance over time; this means that the population will increasingly cluster around certain areas of the domain while avoiding others. Reasoning based on the theorem leads to expectations of the relationship between a genetic algorithm's population size and the occurrence of structural bias, which match our empirical findings from Section \ref{sect:numres}; further reasoning predicts a relationship between structural bias and problem difficulty, which is tested in the experiments of the next section. Section \ref{sect:conseq} begins by outlining and demonstrating approaches to visually investigate, and then to quantify, the levels of structural bias inherent in the design of an optimization algorithm. This is followed by an examination of how structural bias seems to manifest differently when we apply our standard genetic algorithm to variety of functions from a well-known test suite. The findings from these experiments again match with theory-grounded expectations arising from our arguments in Section \ref{sect:sbiasGA}. Finally, Section \ref{sect:conc} provides a summary of the paper, and brief discussions of its wider relevance, such as the manifestation of structural bias in combinatorial spaces.

\section{Structural bias}\label{sect:sbias}
When faced with the task of optimising a given function, the amount of information usually available regarding its features is highly limited. Therefore, one wishes to design an algorithm capable of locating the optima no matter where exactly they are in the search space. This implies that the generating operators of the algorithm must be able to, first, reach every region of the search space and second, ideally, do so without imposing any preferences for some regions of the domain over others. Clearly, different functions and domains give rise to different situations, greatly complicating the prospects for a general theoretical analysis. In addition, such an analysis cannot be tackled directly due to the apparent coupling between the landscape of the objective function and artefacts from the iterative application of algorithmic operators \ie structural bias. It is therefore highly desirable to be able to separate these effects. A closer inspection reveals that, in almost all cases, the action of a selection operator actually can be characterised as the imposition of a stochastic rank-ordering over a specific set of values of the objective function in the current population. If we replaced the objective function with uniform random noise, over a series of statistically significant number of independent runs, this would enable us to separate 'landscape effects' from 'algorithm design effects', eliminating the influence of the ("geographical") position of selected points, but retaining algorithmic artefacts.

Therefore, one way to overcome this coupling issue is to use "the most random" test function, such that its value at any point does depends neither on the values within its neighbourhood nor on the past (independent) evaluations at this point \ie be independent and identically distributed (i.i.d.). For the sake of simplicity, and without loss of generality, we can consider an artificial objective function
\begin{equation} \label{eq:f0}
f_0:D\subset \mathbb{R}^n \to [0,1] \mbox{, where } x\in Uniform(D), f_0(x)\in Uniform[0,1], x \mbox{ and } f_0(x) \mbox{ are i.i.d. } 
\end{equation}
as this "most random" function. Again, without loss of generality, we can consider $D=[0,1]^n$. Such an $f_0$ contains no structure stable over different runs, therefore an \textit{ideal optimisation method} will arrive at different regions of the search space over a series of runs. Indeed, over multiple runs, it will cover the entire search space with uniform probability.

\begin{figure*} \centering
\subfigure[beginning]{\includegraphics[width=.32\linewidth]{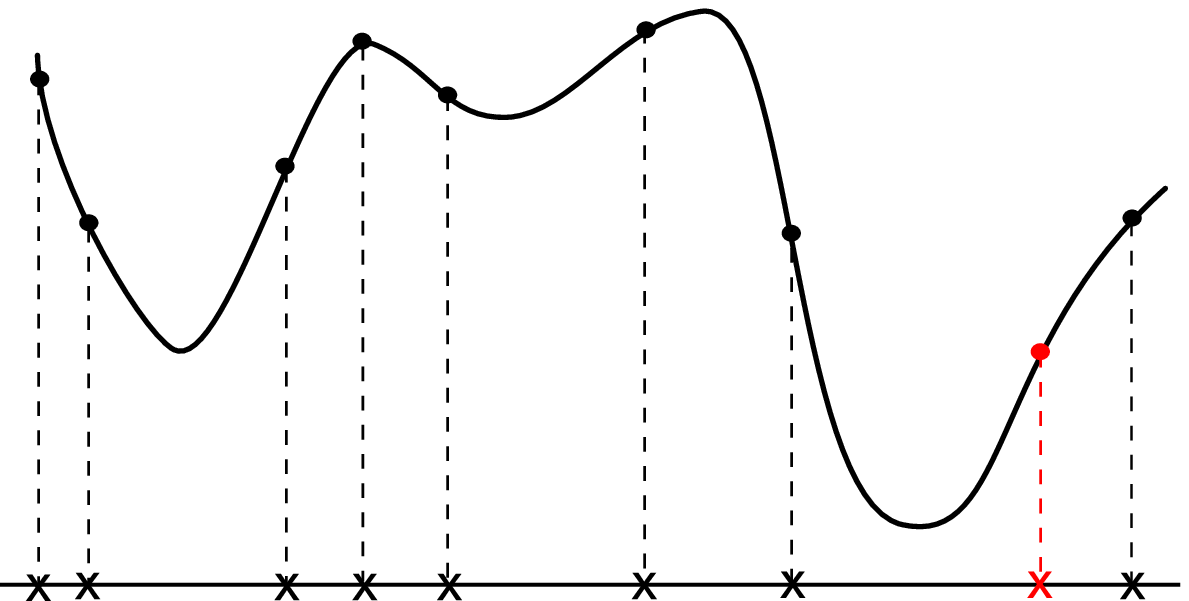}}
\subfigure[halfway through]{\includegraphics[width=.32\linewidth]{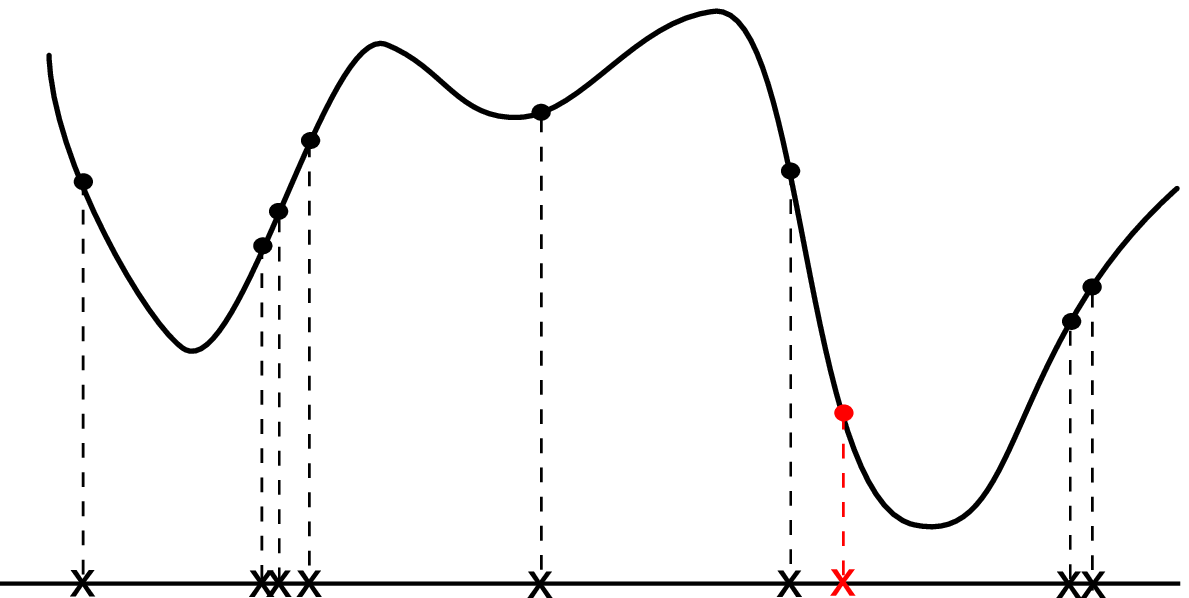}}
\subfigure[end]{\includegraphics[width=.32\linewidth]{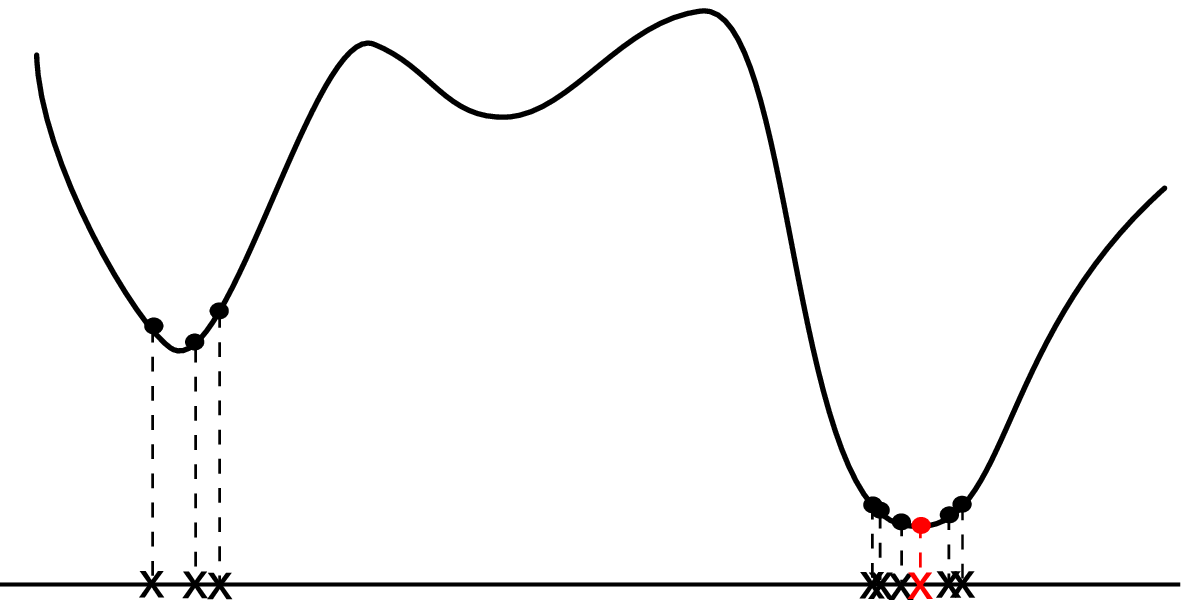}}
\caption{Sketch of a typical progress of an evolutionary algorithm on a minimisation problem in terms of population distribution with projections of points' coordinate, adapted from \cite{cit:EibenSmith}. Red points mark best points in the population.}\label{fig:EAprogress}
\end{figure*}%
As shown in Fig. \ref{fig:EAprogress}, the typical progress of a capable evolutionary algorithm consists of three stages \cite{cit:EibenSmith}: in the beginning, the population is spread randomly over the domain, roughly halfway through the optimisation the population starts rolling down the hill, and in the final stages of optimisation the whole population is concentrated around the minima. Thanks to the construction of $f_0$, independent runs of the algorithm provide different landscapes, all of identical difficulty (due to the i.i.d. property), where populations move/converge towards minima located at different parts of the domain. In other words, over a series of runs of the algorithm, the situation shown in Fig. \ref{fig:EAprogress} is replicated for different landscapes where the optimisation process arrives at different parts of the domain - that is, red points will be distributed all over the interval. In the following section we show that minima of $f_0$ are in fact distributed uniformly over $D$. This implies that the distribution of minima found by an ideal unbiased algorithm across different runs should be uniform as well.

\subsection{Distribution of minima of $f_0$}
Assume that points $Z_1,..Z_N$ are independent and identically distributed. Assume that each of these points (say, $Z_i$) is assigned a mark $X_i$ and assume that $X_1, X_2,..,X_N$ is a collection of i.i.d. random variables with an absolutely continuous distribution. Assume also that the sets $X_1, X_2,..,X_N$ and $Z_1,..Z_N$ are mutually independent. Let $I = \arg\min_i X_i$ be the index of the point with the lowest mark.
\begin{remark}
We only assume that the distribution is absolutely continuous for convenience here. This ensures that $P(X_i=X_j) = 0$ for any $i \neq j$. This makes our proofs shorter and more transparent but is not essential for our statements to hold.
\end{remark}
\begin{remark}
In the notation above, $Z_1,..Z_N$ represent the vector of points' coordinates and $X_1,..X_N$ represent values of the objective function at these points.
\end{remark}

{\bf Proposition.} The distribution of $Z_I$ is the same as that of $Z_1$ (or the same as the distribution of any of the initial points).\\
{\bf Proof.} Note first that $P(I=i) =1/N$ for any $i$. This is evident as
$$
\sum_{i=1}^N P(I=i) = 1,
$$
and all probabilities are equal to each other due to the identical distribution of $X$.

We can now calculate, for any set of points $A$
\begin{eqnarray*}
P(Z_I \in A) = \sum_{i=1}^N P(I=i)P(Z_I \in A|I=i) =\sum_{i=1}^N \frac{1}{N} P(Z_i \in A) = P(Z_1 \in A) {}_{\square}
\end{eqnarray*}
This shows that minima of $f_0$ defined in (\ref{eq:f0}) are distributed uniformly over $D$.

\subsection{Further comments on $f_0$}
Since, by construction, $f_0$ is in effect a noisy signal with zero smoothness \ie no correlation between neighbouring points, it is clearly not suited for testing the quality of fitness improvement or as a direct guide in assembling the algorithm. As explained previously, the rationale behind using $f_0$ is solely to elucidate the underlying structural bias of the tested algorithm. More comments on this issue are given in Section \ref{sect:further_numerical}.

\section{Numerical results}\label{sect:numres}
In this section we illustrate the use of $f_0$ in investigating the occurrence of structural bias in different algorithm configurations. We apply this 'structural bias probe' to two algorithms that are frequently deployed in optimization practice and research, namely: a genetic algorithm (GA), and particle swarm optimisation (PSO). In both cases, our instantiations of the algorithms (and the subsequent further analyses), are in the context of optimization in a continuous decision space (i.e. the optimization of vectors of real-valued parameters). Combinatorial optimization is certainly also of interest, and we later briefly speculate on structural bias in that scenario. However, our focus here on real-valued decision spaces is consistent with the observation that real-valued optimization (particularly via PSO variants) is the most rampant breeding ground for the publication of new algorithms. As such, real-valued optimization can be considered in relatively more need for techniques that can help researchers or practitioners discern performance-related properties of new algorithm designs.

\subsection{Typical genetic algorithm}\label{sect:GA}
As the first example of a randomised population-based algorithm used to solve the problem of minimisation of $f_0:[0,1]^n \to [0,1]$, we consider the most straight-forward example - a typical steady-state genetic algorithm (GA) where solutions are encoded as strings of real values of length $n$ and are subject to the following transformations:\\ 
\textit{1. initialize and evaluate a population of $N$ solutions within the boundaries of problem domain\\
2. continue until the maximum number of fitness evaluations is reached (300000)\\
2.1. select $parent_1$ from the population via tournament selection \cite{cit:Goldberg1989} of size $n_t$ (here, $n_t=2$) in the following manner: \\
2.1.1. select at random $n_t$ solutions;\\
2.1.2. based on their fitness values, choose the best solution to become a parent; \\
2.2. similarly, select $parent_2$ via tournament selection of size $n_t$, independently on the choice of another parent\\
2.3. generate child solution as $parent_1 +\alpha * (parent_2 - parent_1)$, $\alpha \sim Uniform(-d,1+d)$ re-sampled for each dimension, $d=0.25$\\
2.4. with probability $p=1$, mutate child solution via Gaussian mutation -- perturb every coordinate independently with $\delta \sim N(0,m_d* r)$, $m_d=0.01$, $r$ is the width of search domain in this coordinate\\
2.5. evaluate child solution\\
2.6. if child solution is better or equal to current worst solution in population then child solution replaces it\\
2.7 end of loop, go to step 2. }${}_{\square}$

All specified parameters represent standard choices in the field of evolutionary computation. If a result of an operator, in some dimension, goes outside the domain, it is corrected in a saturation manner where it is forced to the closest domain boundary in this dimension. The dimensionality of the problem is set to $n=30$ as a value high enough to be relevant for the field but low enough to allow clear visualisation. This also dictates the choice of the termination criterion as 300000 fitness evaluations, following \cite{cit:Suganthan2005}. We consider three settings for population size: $N=5$, $N=20$, $N=100$. To provide enough statistical power for the results, $50$ independent runs are considered for each parameter setting. According to\cite{cit:Rudolph1996}, in the limit, this algorithm converges to the global minimum of any real-valued function $f:M\to\textbf{R}$ whose values are bounded below and $M$ is an arbitrary domain. 

\subsection{Numerical results for a typical genetic algorithm}\label{sect:numresGA}
\begin{figure*} \centering
\subfigure[GA $N=5$ start]{\includegraphics[width=.34\linewidth,angle=270]{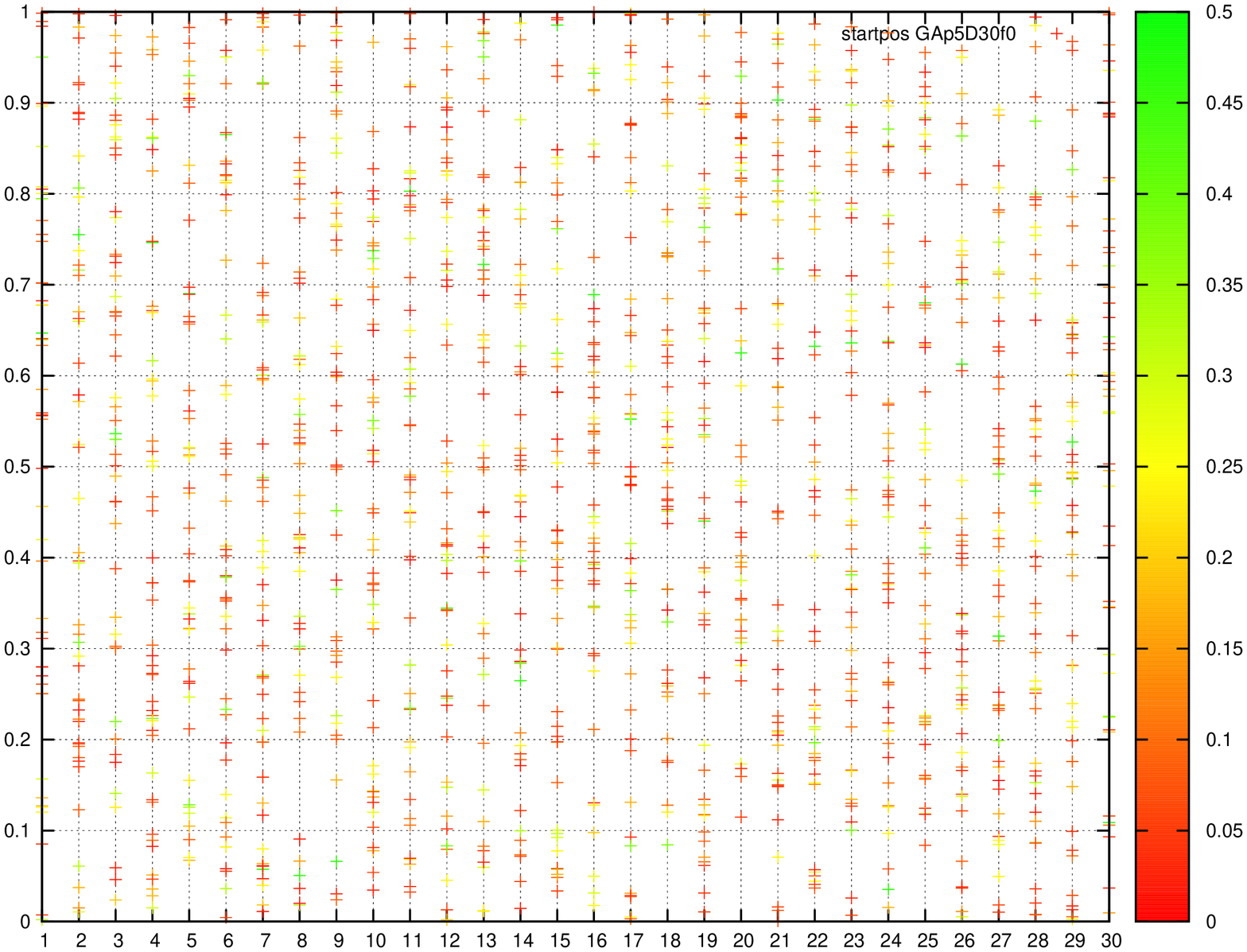}}
\subfigure[GA $N=5$ end]{\includegraphics[width=.34\linewidth,angle=270]{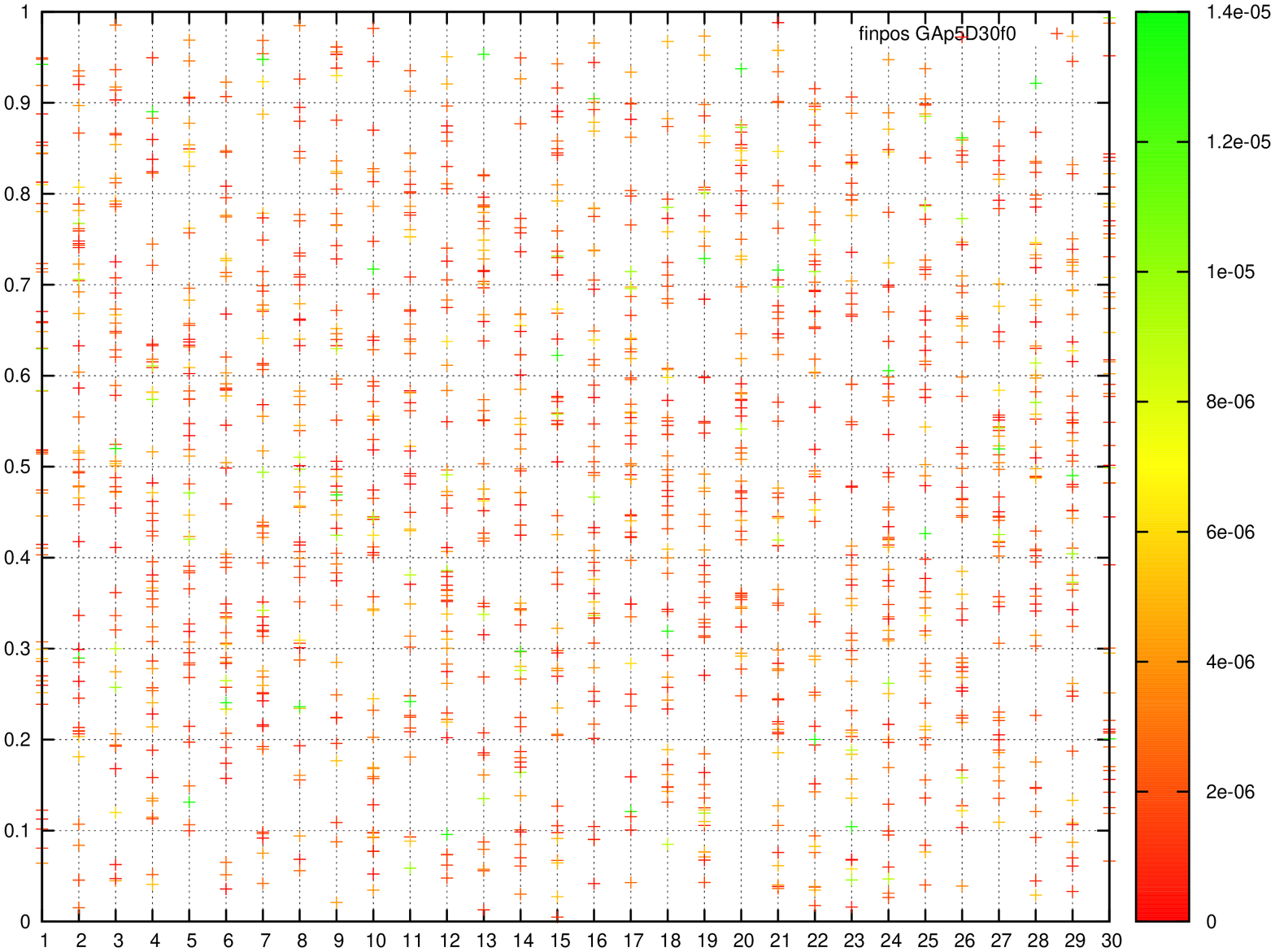}\label{fig:GAcoors_finp5}}\\
\subfigure[GA $N=20$ start]{\includegraphics[width=.34\linewidth,angle=270]{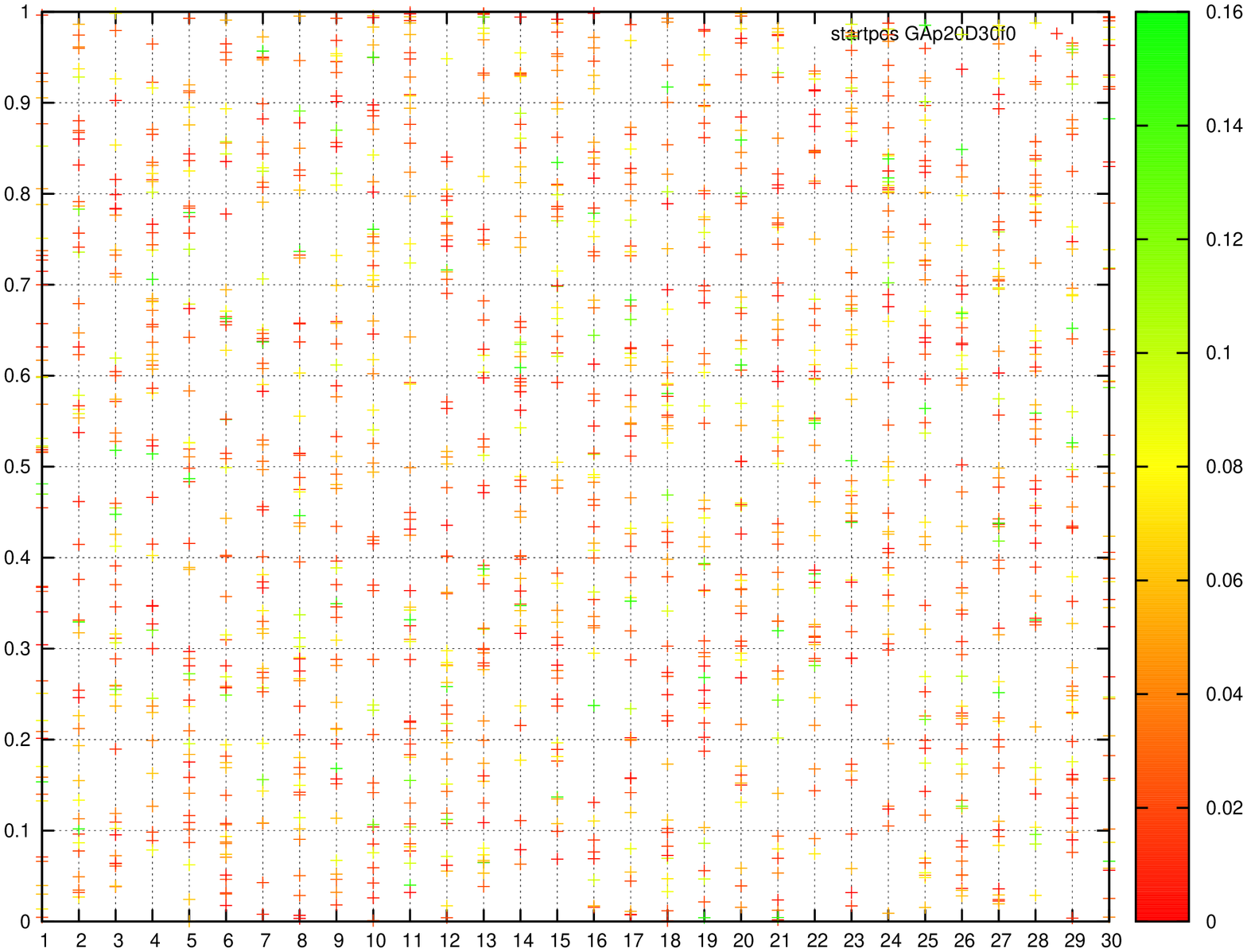}}
\subfigure[GA $N=20$ end]{\includegraphics[width=.34\linewidth,angle=270]{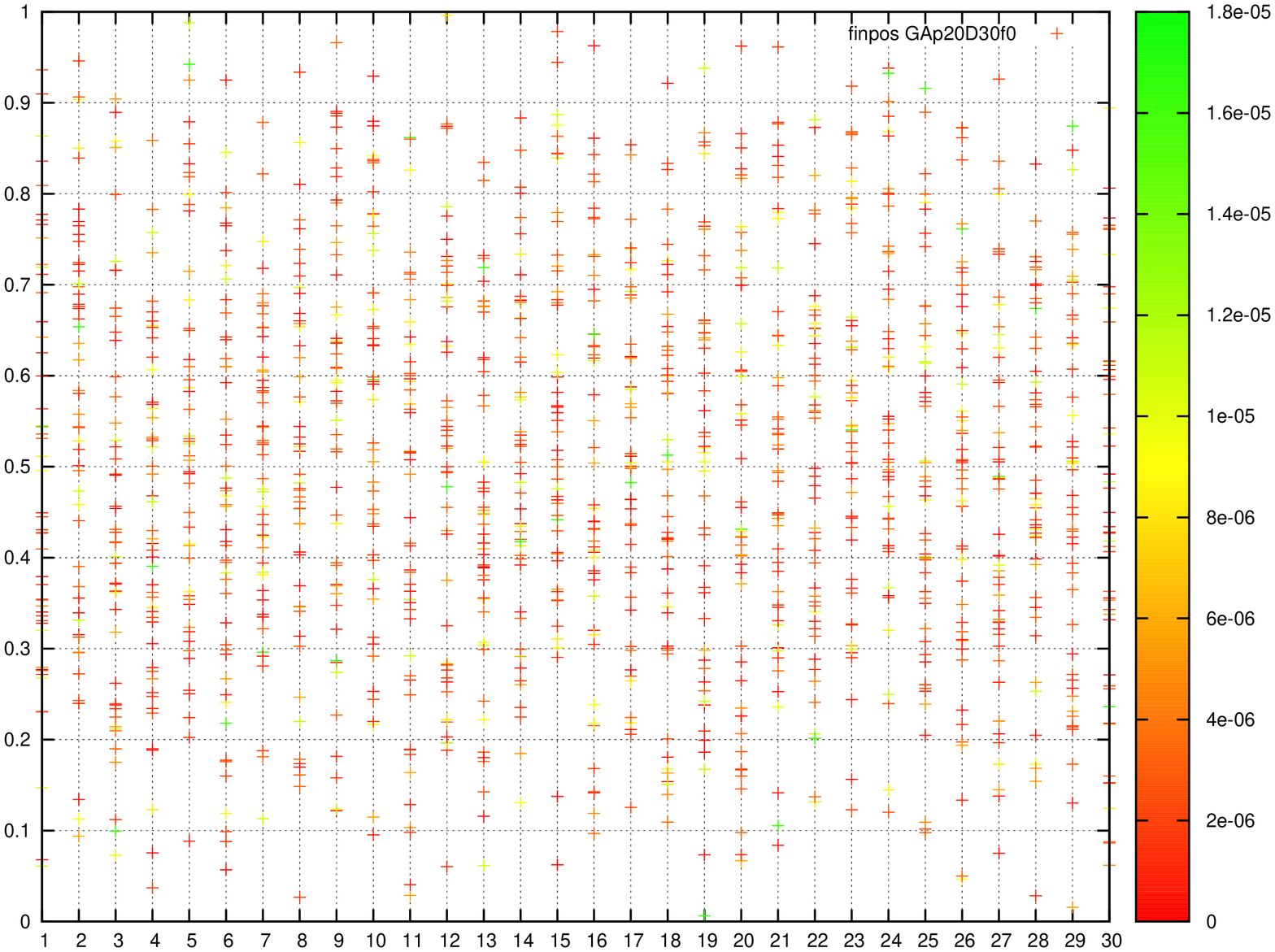}\label{fig:GAcoors_finp20}}\\
\subfigure[GA $N=100$ start]{\includegraphics[width=.34\linewidth,angle=270]{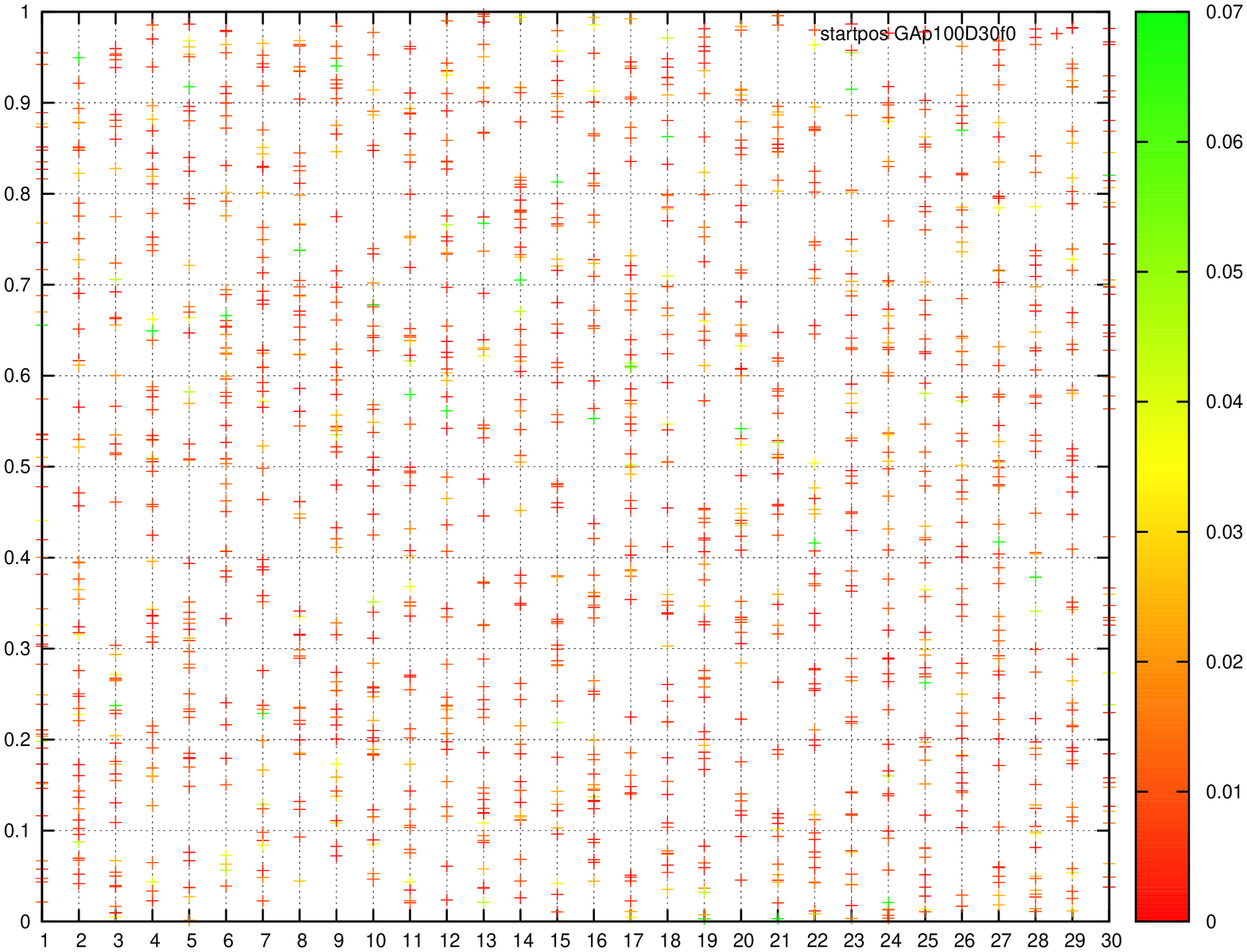}}
\subfigure[GA $N=100$ end]{\includegraphics[width=.34\linewidth,angle=270]{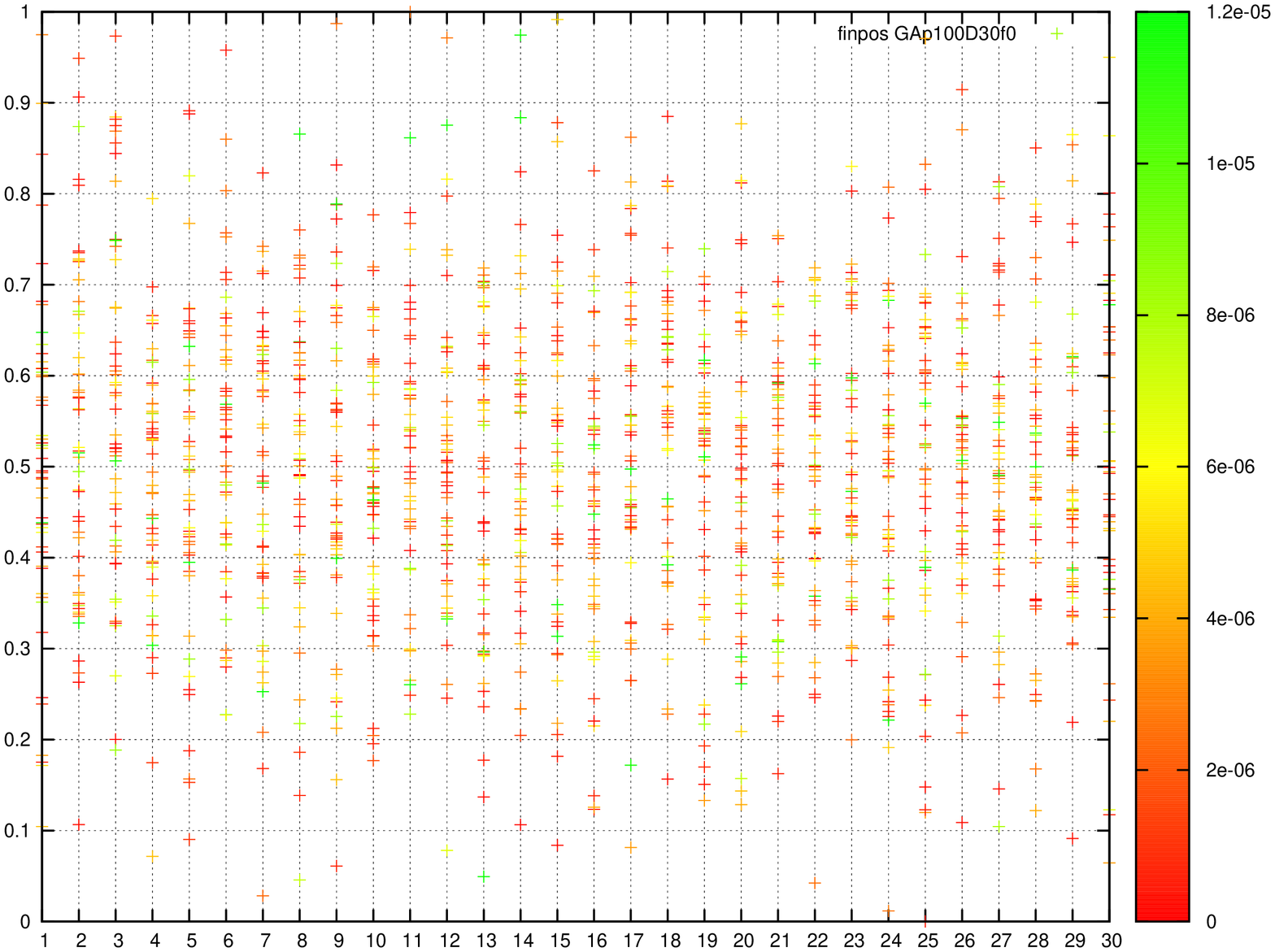}\label{fig:GAcoors_finp100}}
\caption{Positions of points with the best fitness values in the first (left column) and the last (right column) populations of 50 runs of the considered GA for different population sizes in parallel coordinates; horizontal axis shows the ordinal number of the coordinate, vertical axis shows the range of this coordinate; fitness value of each point is shown in colour. A clear bias towards the centre of the search space is visible in the last populations as population size increases.}\label{fig:GApcoors}
\end{figure*}
A convenient way of visualising multidimensional data is by the method of 'parallel coordinates' \cite{cit:dOcagne1885}, \cite{cit:Inselberg1985} which allows an insight into the space regardless of its dimensionality. Using this technique to visualise an $n$-dimensional point, a backdrop consisting of $n$ vertical equally spaced parallel lines is drawn and a point in $n$-dimensional space is represented as a collection of markers on each of these $n$ lines, each matching a value of the corresponding coordinate. Traditionally, these markers are connected to form polylines (piecewise linear curves) which can reveal 2-dimensional patterns for certain high dimensional properties \cite{cit:Inselberg2008}. Unfortunately, finding the correct layout for each dataset to facilitate data exploration is a problem on its own \cite{cit:Zhou2008}, especially for high values of $n$ \cite{cit:Hurley2010}. Such investigation is currently beyond the scope of our interests, however it may become of interest for algorithms that use highly correlated search strategies. Since the focus of this paper is solely the movement of the population of points in the search domain, unconnected markers suffice. Using colour allows us to visualise the additional dimension - the value of the objective function at the point.

\begin{figure*} \centering
\includegraphics[width=56mm]{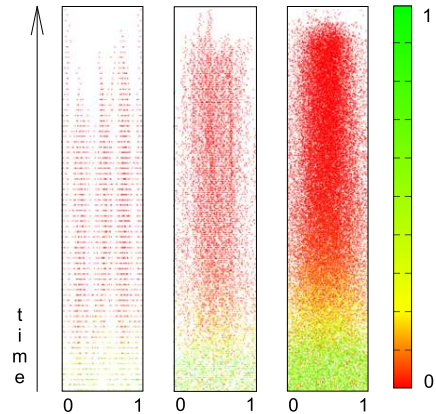}
\caption{Evolution in time of $50$ populations of a typical genetic algorithm in selected dimension for $N=5$, $N=20$ and $N=100$. Horizontal axis shows values of coordinate, vertical axis represents time. Colour of the dots corresponds to values of objective function}\label{fig:allposf0}
\end{figure*}%
Following the technique described above, Fig. \ref{fig:GApcoors} shows, in parallel coordinates, positions of points with the best fitness values in the first (left column) and the last (right column) populations for $50$ runs of the considered genetic algorithm, for different population sizes $N=5$, $N=20$, $N=100$. Clearly, for all population sizes, the initial distribution of positions in the left columns of the figures is close enough to uniform. However, in the right column, instead of seeing a near-uniform distribution of points, a clear bias towards the centre of the search space becomes more evident in the final populations as population size increases. In other words, a genetic algorithm with bigger populations tends to avoid the corners of the search domain and concentrates more on sampling points closer to the middle of the interval, for no obviously apparent reason. Such behaviour is barely noticeable for $N=5$, more pronounced for $N=20$ and is very clear for $N=100$. These anomalies represent structural bias. Our numerical results also suggest that this behaviour is consistent throughout time and does not depend on termination criterion - consecutive populations spread out less and less from the middle of the search domain, see Fig. \ref{fig:allposf0} which shows the evolution in time of positions of all points of $50$ populations in a selected dimension for all three population sizes. \textit{We therefore conclude that, owing to structural bias, a typical genetic algorithm with a large population potentially wastes the fitness evaluations budget via oversampling a region of the search domain, to the detriment of overall performance.}

\subsection{Typical Particle Swarm Optimisation algorithm}\label{sect:PSO}
Particle swarm optimisation (PSO) is another example of a population-based optimisation algorithm introduced by Kennedy and Eberhart in \cite{cit:Kennedy}, and then developed in various variants for test problems and applications. The main metaphor employed in particle swarm optimisation is that a group of particles makes use of their personal and social experience in order to explore a decision space and detect solutions with high performance.\\
More specifically, to minimise $f_0:[0,1]^{30} \to [0,1]$, the following steps are taken:\\
\textit{1.1. initialise a population of $N$ solutions within the boundaries of the problem at $t=0$\\
1.2. evaluate every solution in the population based on the objective function\\
1.3. for each solution, assign its personal best position $p_t^{pb}=p_0$\\
1.4. assign the global best position to $p_t^{gb}$\\
1.5. for each solution, initialise a speed vector $v_0=(v_0^1,...,v_0^n)$ such that $v_0^i\sim Uniform[0,0.1]$\\
2. continue until the maximum number of fitness evaluations is reached (300000)\\
2.1. update the speed vector for every solution in the population as $v_{t+1}=c_0v_t+c_1\alpha_1(p_t^{pb}-p_t)+c_2\alpha_2(p_t^{gb}-p_t)$, where $\alpha_1,\alpha_2\sim Uniform[0,1]$ are re-sampled independently for each solution and $c_0=1$, $c_1=2$, $c_2=2$\\
2.2. if $\left \| v_{t+1} \right \|_2 > 0.2$ substitute it coordinate-wise with $\frac{0.2 v_{t+1}^i}{\left \| v_{t+1} \right \|_2}$, $i=1,...,n$\\
2.3. update the position of every solution in the population as $p_{t+1}=p_t+v_{t+1}$, evaluate the new solution\\
2.4. if needed, update personal best position $p_t^{pb}$ for each solution\\
2.5. if needed, update the global best position $p_t^{gb}$\\
2.6. $t=t+1$, end of loop, go to step 2}${}_{\square}$

As well as in the previous section, the algorithm and specified parameters represent standard choices in the field of evolutionary computation. To allow fair comparison, the termination criterion is kept as $300000$ fitness evaluations. Finally, echoing the experiments done with a typical genetic algorithm, here we also use the three settings for population size $N=5$, $N=20$, $N=100$, and we perform $50$ independent runs for each.

\subsection{Numerical results for PSO}
\begin{figure*} \centering
\subfigure[PSO $N=5$ start]{\includegraphics[width=.34\linewidth,angle=270]{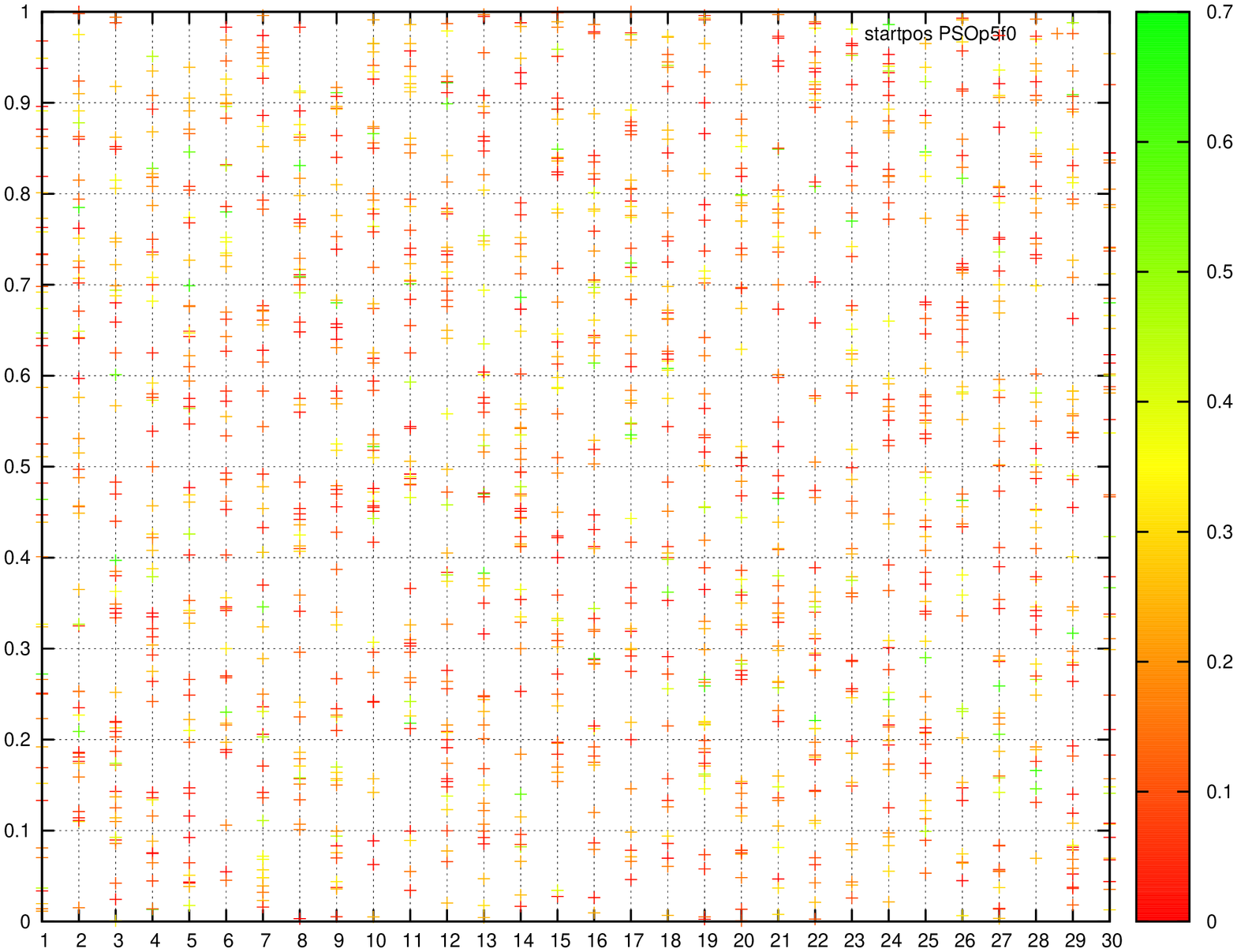}}
\subfigure[PSO $N=5$ end]{\includegraphics[width=.34\linewidth,angle=270]{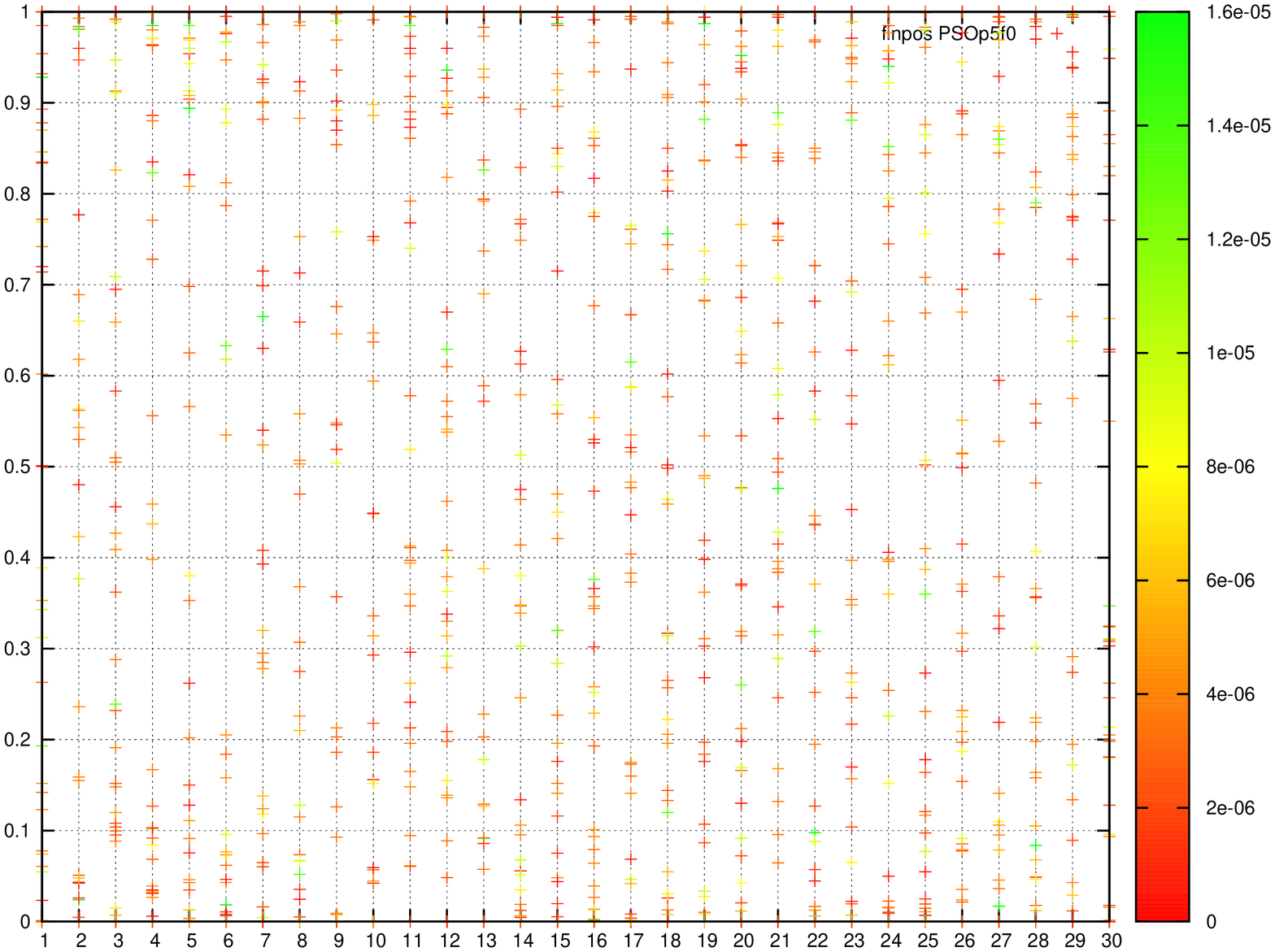}\label{fig:PSOcoors_finp5}}\\
\subfigure[PSO $N=20$ start]{\includegraphics[width=.34\linewidth,angle=270]{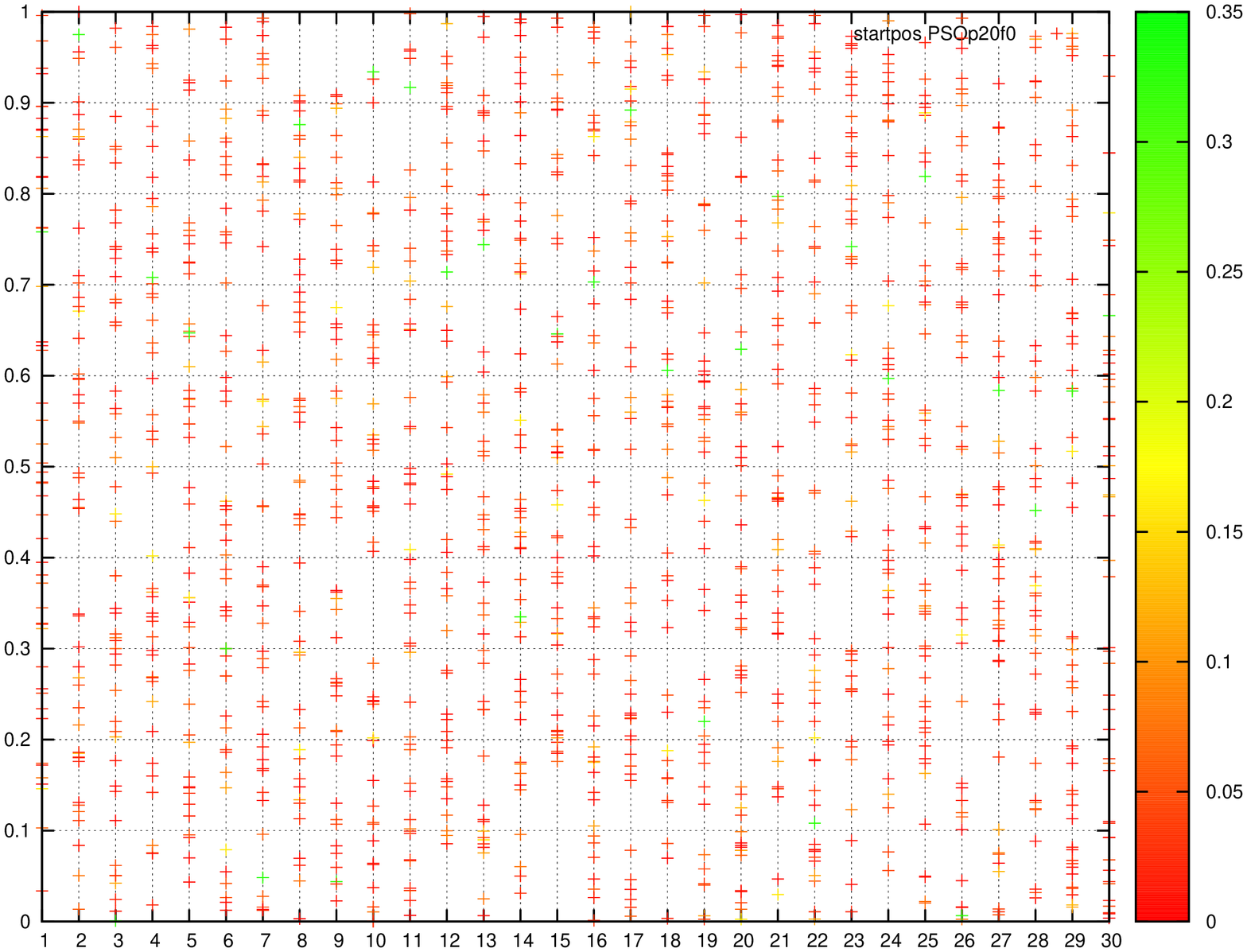}}
\subfigure[PSO $N=20$ end]{\includegraphics[width=.34\linewidth,angle=270]{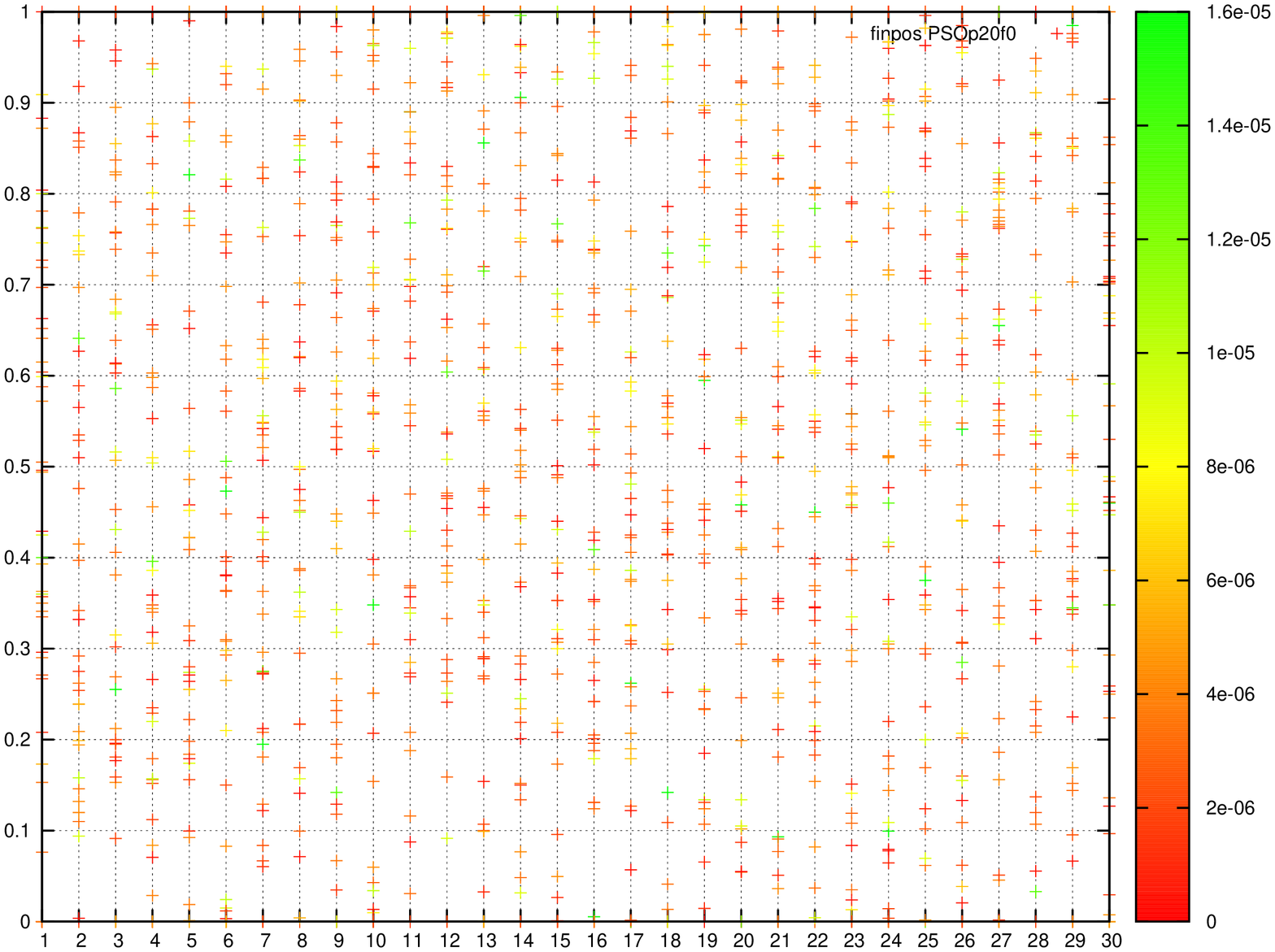}\label{fig:PSOcoors_finp20}}\\
\subfigure[PSO $N=100$ start]{\includegraphics[width=.34\linewidth,angle=270]{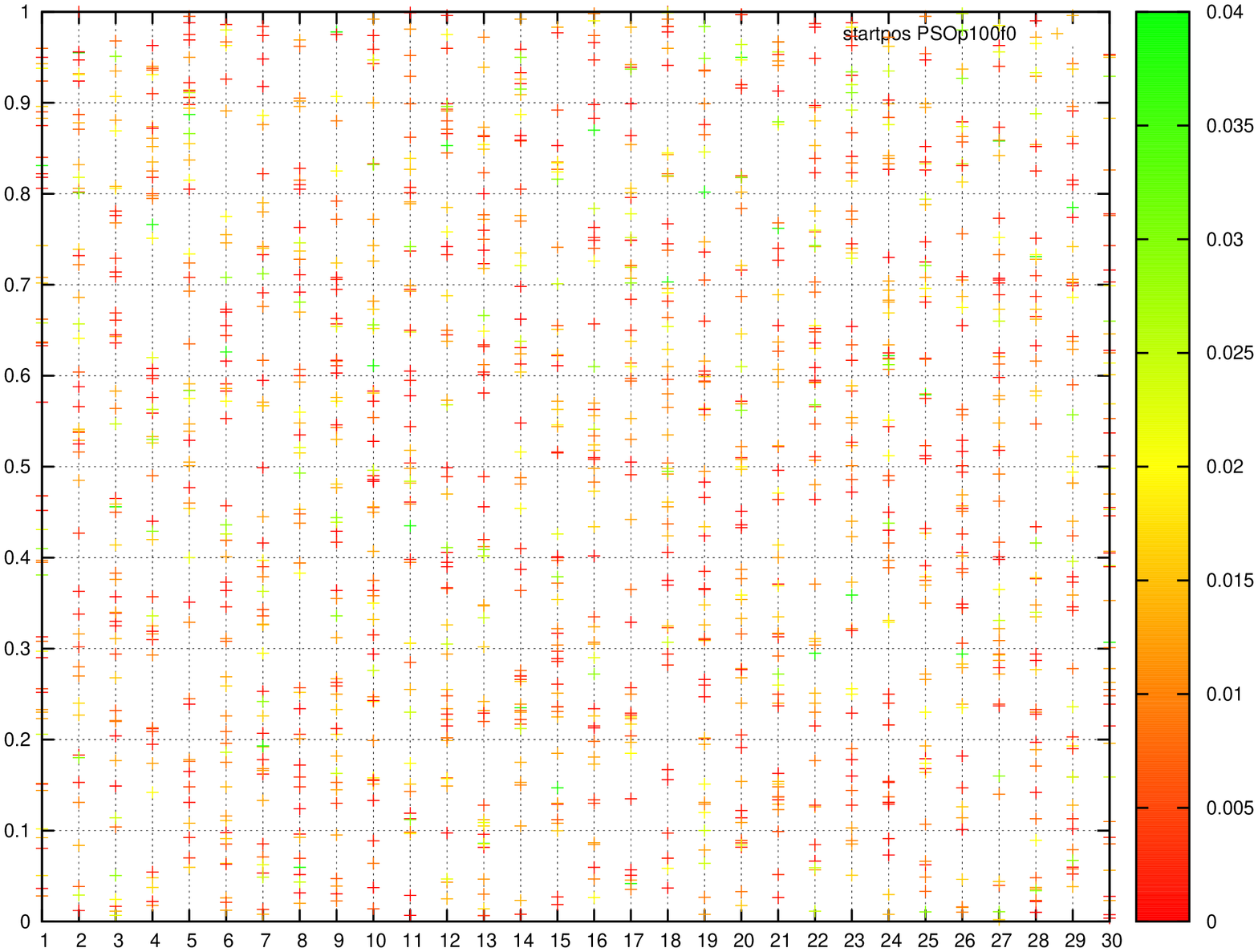}}
\subfigure[PSO $N=100$ end]{\includegraphics[width=.34\linewidth,angle=270]{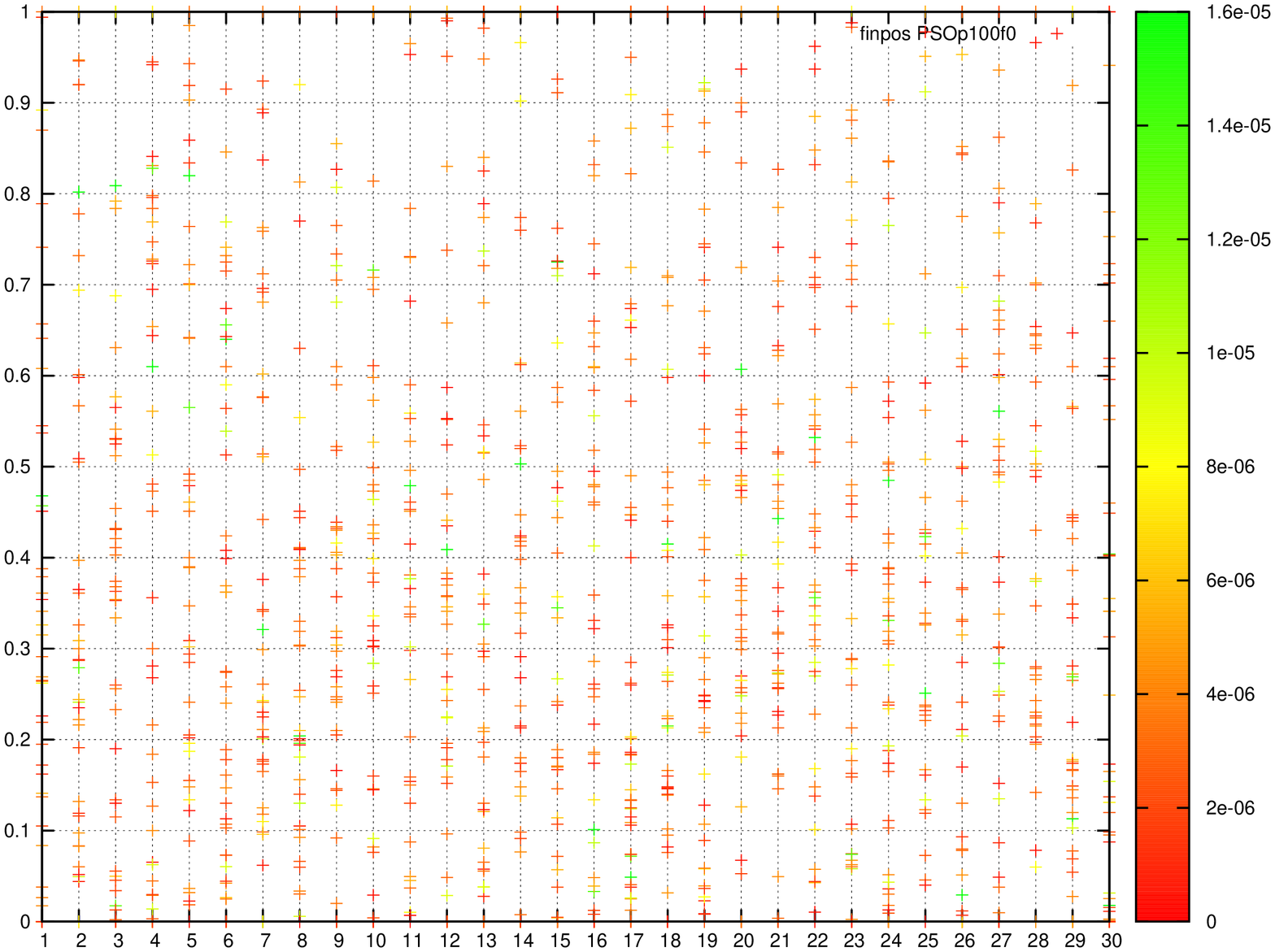}\label{fig:PSOcoors_finp100}}
\caption{Positions of points with the best fitness values in the first (left column) and the last (right column) populations of 50 runs of the considered PSO for different population sizes in parallel coordinates; horizontal axis shows the ordinal number of coordinate, vertical axis shows the range of this coordinate; fitness value of each point is shown in colour. A more complex type of dependency of the population size on the bias induced is present for the top and bottom values of the population size.}\label{fig:PSOpcoors}
\end{figure*}%
The same techniques as in Section \ref{sect:numresGA}, applied to the analysis of PSO, reveal a rather different situation as shown in Fig. \ref{fig:PSOpcoors}. As expected, the distribution of initial positions of points with the best fitness shown in the left columns of the figures is close enough to uniform. But, as in the case of the typical genetic algorithm, instead of a near-uniform distribution, positions of the final points show a clear bias, albeit of a different nature. A more complex type of dependency of the population size on the bias induced is present for the highest and lowest values of the considered population sizes. In the case of $N=5$, positions of final points clearly group around the corners and avoid regions in the middle of the search domain. Meanwhile, for $N=100$, the final points tend to be positioned closer to one corner of the hypercube domain and avoid its opposite corner. The behaviour of PSO with $N=20$ is also not ideal in terms of the distribution of final points, as they appear slightly further apart compared to the initial distribution. These anomalies clearly demonstrate the presence of structural bias in the considered version of PSO. Moreover, since none of the operators that makes up this PSO clearly predispose its population to cluster in such manner, this effect seems to be emerge from their combined dynamics.

\subsection{Remarks regarding the random generator}\label{sect:random}
The empirical results for the genetic algorithm in this paper are produced involving the use of a standard Java 48-bit random generator underpinning, where required, the generation of 'random' numbers within the implemented algorithm. This commonly-used pseudorandom generator is based on the linear congruential generator (LCG) with a period of $2^{48}\approx 2.8\cdot 10^{14}$, while the seed is automatically generated via the system routine \textit{System.currentTimeMillis()}. Meanwhile, our empirical PSO results are produced using the standard Unix function \textit{drand48} with the same parameters as above and the seed value is obtained via \textit{srand48}. It is well-known \cite{cit:Marsaglia1968} that when a series of consecutive values are obtained from this type of random generator to form multidimensional points, they end up lying on a finite number of hyperplanes intersecting the intended domain. This property is usually referred to as the Marsaglia effect. Clearly, unless the precision of the random generator is close to the precision used by the algorithm, this constitutes a problem as, even in the limit, these points cannot fill all of the domain. The number of such planes is bounded by $(n!m)^{1/n}$, where $n$ is the dimensionality and $m$ is the modulus of the LCG. For the case of $30$ dimensions, the bound on the number of planes is $36$. The quality of each version of LCG can be further assessed based on values of increment and multiplier via estimating the distance between the hyperplanes. However, such calculations are feasible reliably only for low dimensionalities \cite{cit:LEcuyer1999}.  

Another usual concern about random generators is how random their output actually is, in the sense of correlation between successive instances (as opposed to their coverage of the domain). There are two kinds of random generators which differ in how the numbers are produced: true random generators sample some source of entropy \cite{cit:Kenny2005}, whereas pseudorandom number generators use a deterministic algorithm to produce random looking numbers. True random generators measure some physical phenomenon that is expected to be random and then compensates for possible biases in the measurement process. Example sources include measuring atmospheric noise, thermal noise, and other external electromagnetic and quantum phenomena. Being truly non-deterministic and aperiodic, unfortunately, these generators are also slow, costly, inefficient and not reproducible which makes them a bad choice for practical sampling applications. It is still an open question as to whether it is possible in any practical way to distinguish the output of a well designed pseudorandom generator from a perfectly random source without knowledge of the generator's internal state \cite{cit:Kenny2005}. 

How should these observations concern us? Like virtually all implemented stochastic algorithms, our 'random' numbers are \textit{pseudorandom}. Intuitively, we might expect bias in the pseudorandom generator to be swamped by the combined action of the algorithmic operations and subsequently be invisible in the results -- this is, indeed, the common (implicit) approach. However, the design of $f_0$ explicitly removes one of the several dynamic forces that we would otherwise expect to contribute to this 'washing out' of any effects from the pseudorandom generator. To some, combining this with the perhaps-unexpected appearance of evidence for structural bias may lead to a suspicion that what we have observed could be artefacts of the pseudorandom generator. Intuition for the opposite conclusion is well-fuelled. For example, the Marsaglia effect is quickly obscured by aspects of the algorithm that distort the uninterrupted sequential mapping that the Marsaglia effect assumes, and (especially) are dense in operations that will move points away from the 'Marsaglia planes'. Also, as we discuss further below, modern pseudorandom generators are quite effective at avoiding periodic correlations. Nevertheless, in this section we place the pseudorandom generator under close scrutiny in order to uncover evidence as to whether it may have a bearing on our findings. 

To achieve this we have devised three tests, borrowing their design from the body of research that has gone into designing new classes of pseudo-random generators and testing their properties from various angles \cite{cit:NIST_random}, \cite{cit:Walker}, \cite{cit:DieHard}, \cite{cit:CryptX}, \cite{cit:Kenny2005}. Each test within these test suites is aimed at finding a different kind of non-randomness, but as yet no specific finite set of tests is deemed complete to guarantee that some generator is foolproof \cite{cit:Kenny2005}. For the purposes of this paper, the aspects mostly of interest are true uniformity and the absence of correlations in a long sequence of random values. There is no problem with uniformity as generators employed for this paper are among the most popular implementations used and tested widely\footnote{This is also supported by our tests.}. Regarding cross-correlations in the sequence of random values, apart from the aforementioned Marsaglia effect investigated for low dimensionalities, little in the way of theoretical results is available. Values of cross-correlation lag (or 'effective period' as we refer to it here) which need to be studied usually exceed the dimensionality of the objective function, since the majority of algorithms use random values for altering various parameters throughout the run. Careful examination of the pseudocode provided in Sections \ref{sect:GA} and \ref{sect:PSO} shows that both the genetic algorithm and PSO start with initialisation of their populations which require $(dim+1)N_{pop}$ random numbers for the genetic algorithm\footnote{in this and the next three formulas, one extra random number is added to account for evaluation of $f_0$} and $(dim+1+dim)N_{pop}$ random numbers for PSO, where $dim=30$ is the dimensionality of the domain and $N_{pop}$ is population size set to $5$, $20$ and $100$ for both algorithms. Subsequent functioning of the algorithm is periodic in the following sense: producing every new point to be examined by the algorithm requires the same amount of random numbers - $2dim+5$ for the genetic algorithm and $2dimN_{pop}+1$ for PSO. In other words, if $i$ is an index of the element of the pseudorandom sequence which is used to generate the position of the new point in dimension $j$, then $i+p_e$ is the index of the next pseudorandom element which will be used to generate the position of the subsequent new point in dimension $j$, where $p_e$ is the effective period\footnote{For reference, the Marsaglia effect bounds for these effective periods are the following: $41$ for $65$ dimensions for GA with $N_{pop}=5$, $125$ for $301$ dimensions for GA with $N_{pop}=20$, $455$ for $1201$ dimensions for GA with $N_{pop}=100$ and $2221$ for $6001$ dimensions in all three PSO implementations.}. 

\begin{figure*} \centering
\subfigure{\includegraphics[width=46mm]{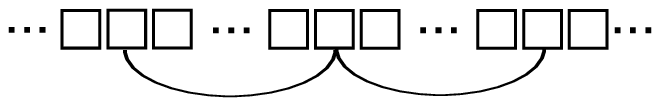}}\\
\subfigure{\includegraphics[width=30mm]{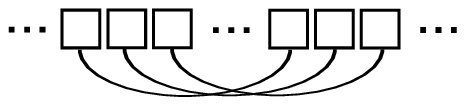}}\\
\subfigure{\includegraphics[width=20mm]{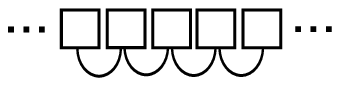}}
\caption{Schematic explanation of tests $1$--$3$ examining correlations between elements of random sequences. Squares represent consecutive elements of random sequence and loops denote considered correlations where length of the loop is constant for each test and referred to as period of this test. Such types of correlations can potentially induce patterns similar to those produced by structural bias.}\label{fig:rnd_tests_scheme}
\end{figure*}
To eliminate the possibility that structural bias observed in algorithms considered in this paper originates from the nature of the pseudorandom number generation rather than being inherent to the algorithm, let us suppose the opposite: there is a correlation between random numbers that are used to generate the values of some coordinate of two subsequent points examined by the algorithm. To examine any such correlations between elements of the pseudorandom sequences, we propose three tests. \textit{Test 1} selects some dimension and examines the correlation between consecutive pairs of random values used to generate points in this dimension. \textit{Test 2} replicates Test $1$ for all dimensions simultaneously. \textit{Test 3} tracks the correlation between consecutive values in the pseudorandom string or, in other words, replicates Test $1$ with period $1$. Schematically, these tests are explained in Fig. \ref{fig:rnd_tests_scheme}, where squares represent consecutive elements of the pseudorandom sequence and loops denote the considered correlations; the length of the loops is constant for each test and referred to as the period of the test.  

\begin{figure*} \centering
\subfigure[Test $1$ for pseudorandom sequence]{\includegraphics[width=.31\linewidth,angle=270]{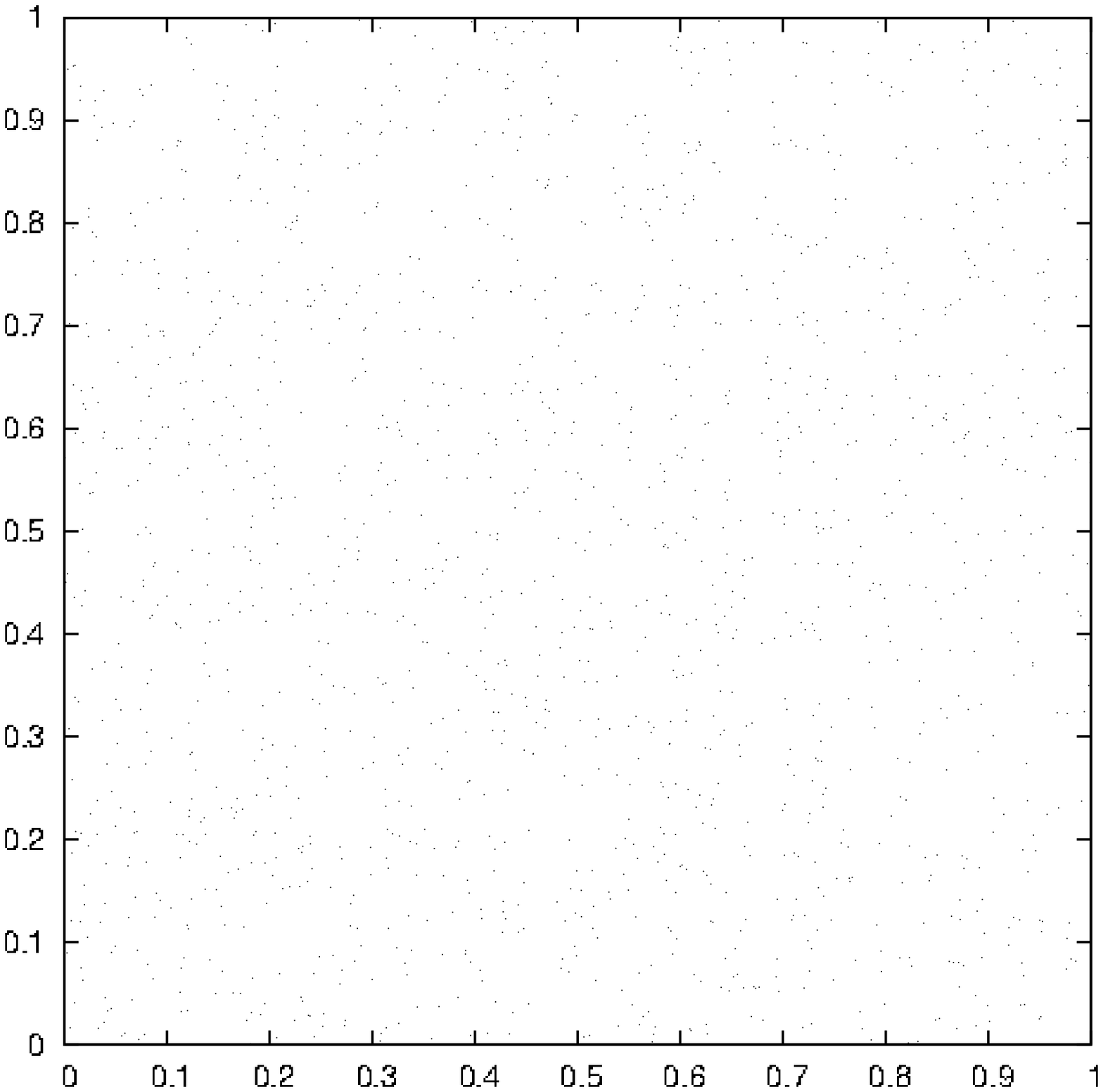}}
\subfigure[Test $2$ for pseudorandom sequence]{\includegraphics[width=.31\linewidth,angle=270]{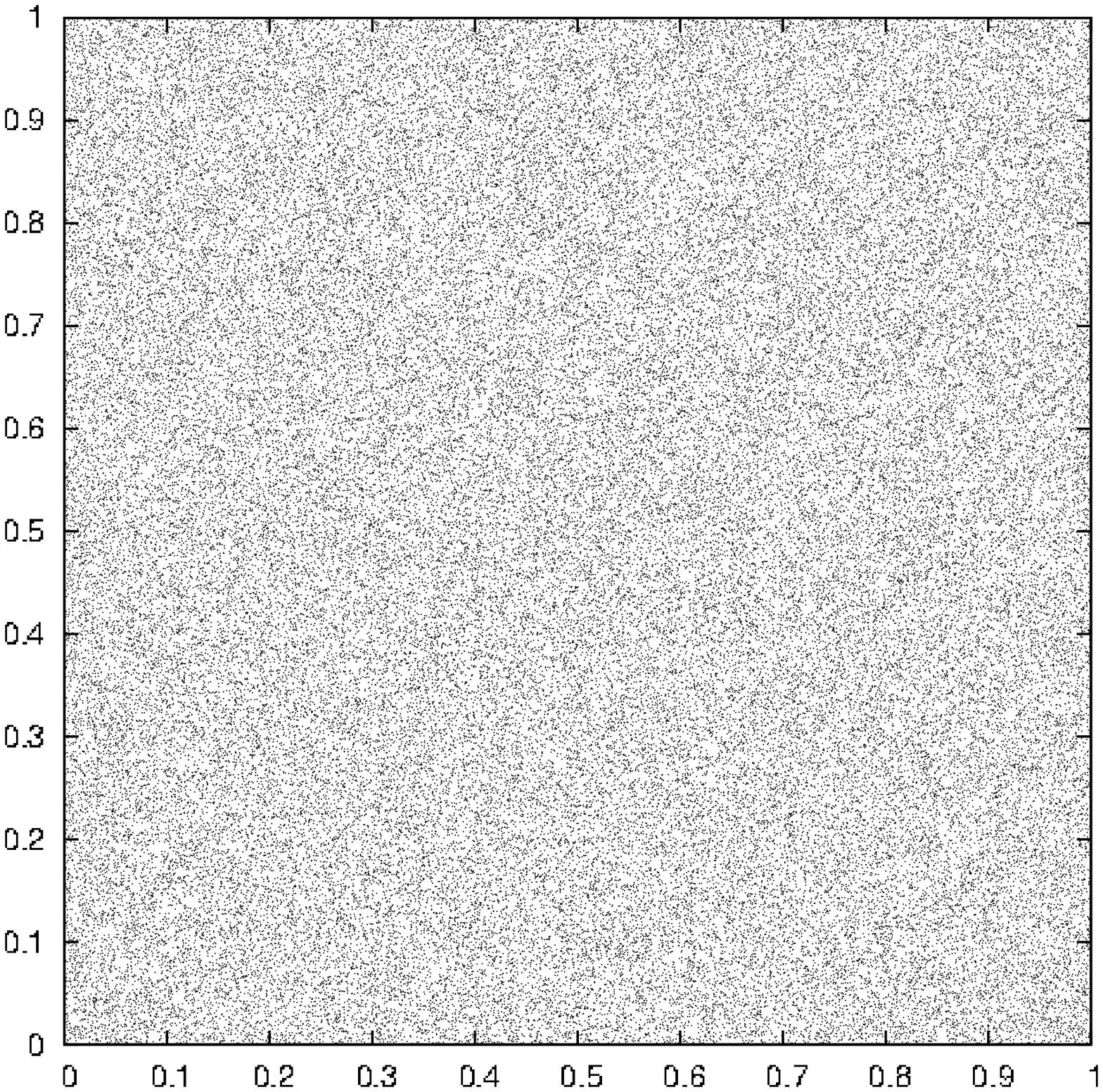}}
\subfigure[Test $3$ for pseudorandom sequence]{\includegraphics[width=.31\linewidth,angle=270]{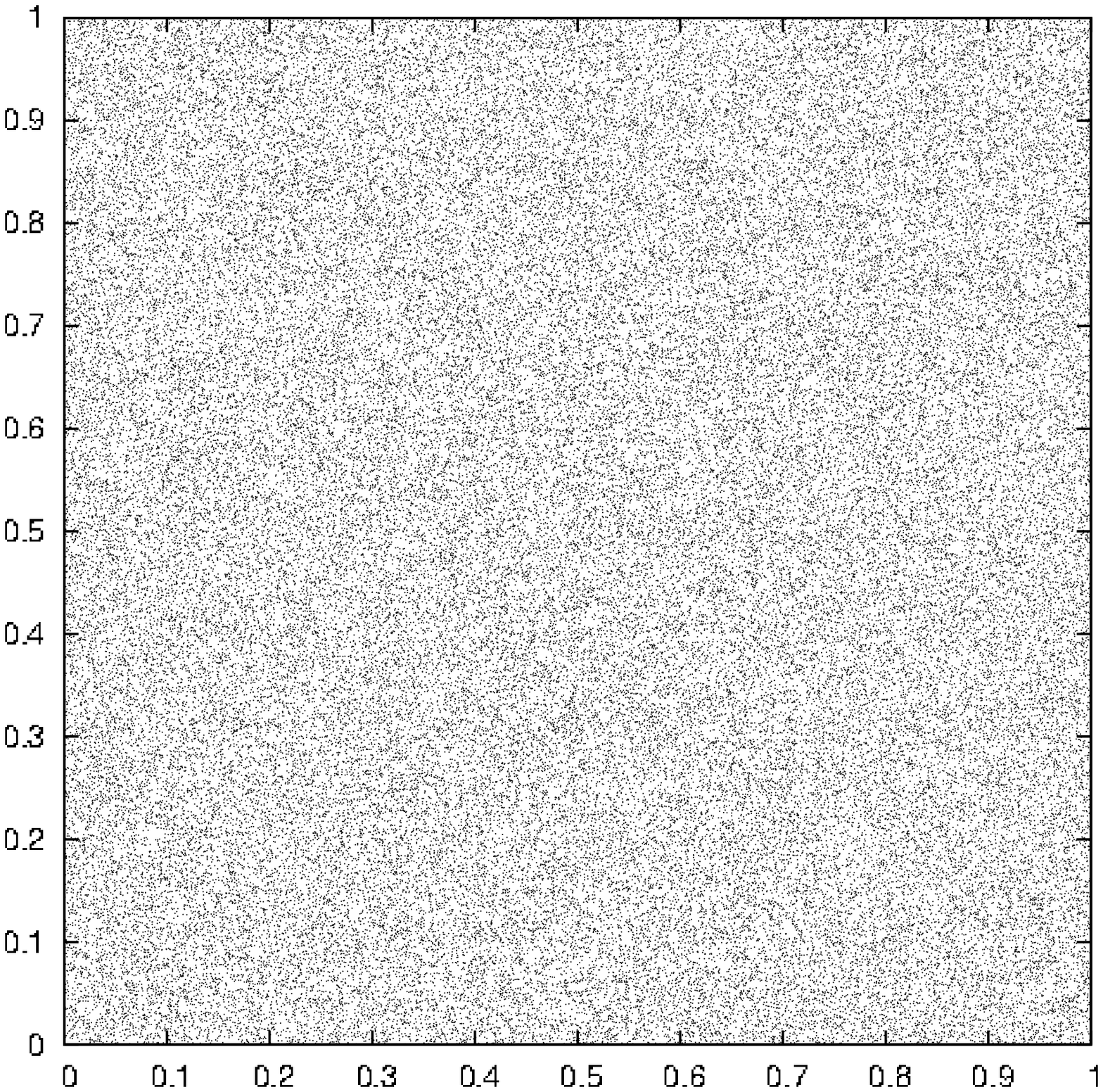}}\\
\subfigure[Test $1$ for true random sequence]{\includegraphics[width=.31\linewidth,angle=270]{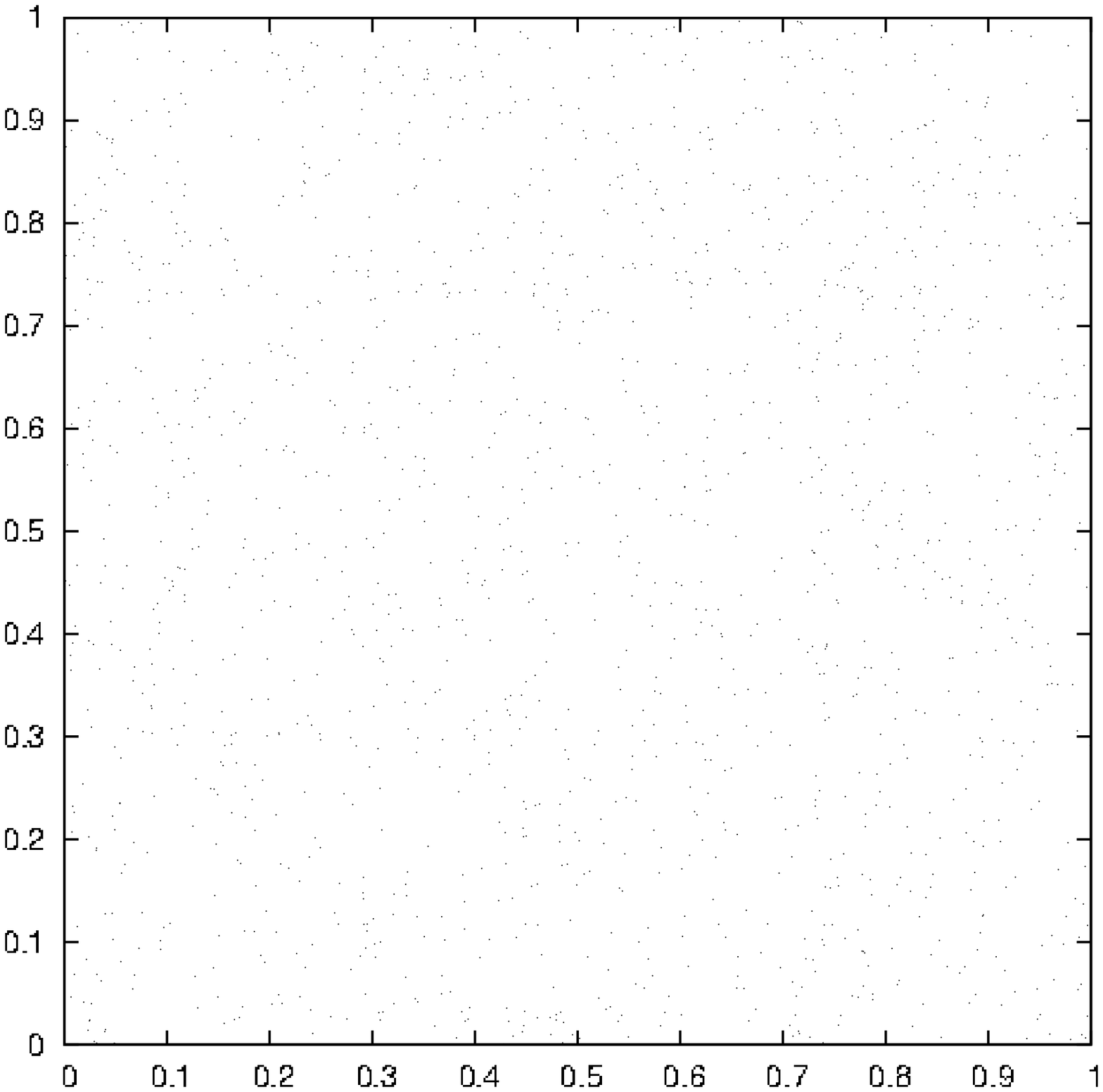}}
\subfigure[Test $2$ for true random sequence]{\includegraphics[width=.31\linewidth,angle=270]{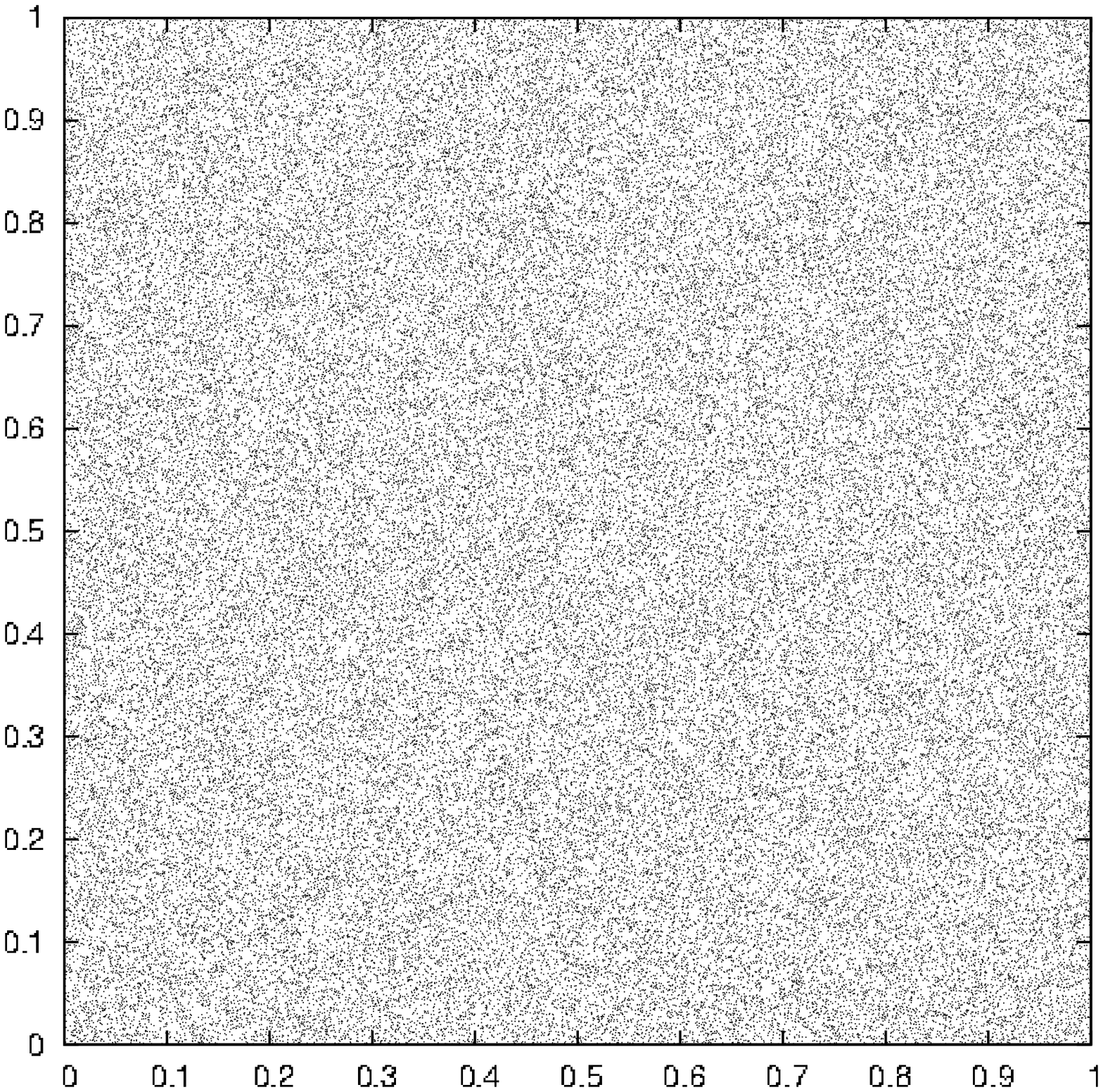}}
\subfigure[Test $3$ for true random sequence]{\includegraphics[width=.31\linewidth,angle=270]{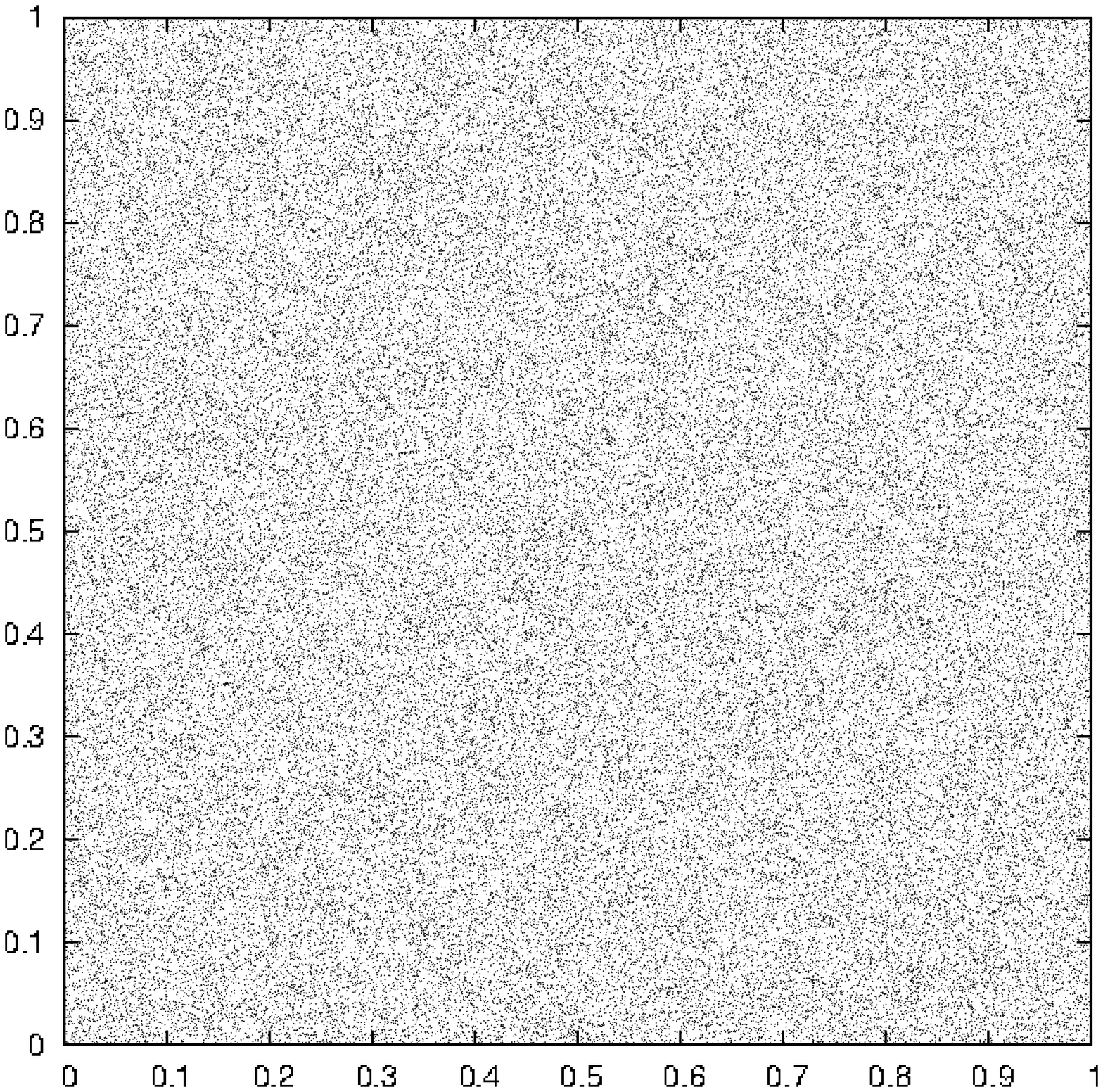}}
\caption{Correlations between elements of pseudorandom sequences of different nature. Results of tests $1$-$3$ for two sequences obtained from pseudorandom (first line) and true random (second line) generators each made up of $100000$ numbers from $[0,1]$. Period for tests $1$ and $2$ is set to $65$ which is a value of effective period of GA implementation considered in this paper for population size $5$; period for test $3$ is equal to $1$. Visual comparison does not reveal any significant differences between true and pseudorandom sequences. Results for other period values and random sequences with different seeds are of identical nature. This suggests that structural bias cannot originate from random generator but rather represents artefacts from the iterative application of algorithmic operators.}\label{fig:rnd_tests}
\end{figure*}

We apply these three tests to each of two kinds of long sequences, one coming from a true random generator and another from a pseudorandom generator used to produce results in Section \ref{sect:numres}. Our 'true random' sequence uses data from a reputable online service \textit{random.org} which generates randomness via atmospheric noise picked up by a radio. This service is subject to a battery of daily tests which confirm that it maintains all of the randomness properties claimed \cite{cit:Kenny2005}. In addition, a series of sequences has been produced via the standard Java generator discussed above for a selection of realistic values of seeds. Lengths of all sequences is set to 100000 elements. Results of tests $1$-$3$ for these two sequences are shown in Fig. \ref{fig:rnd_tests}, where the period for tests $1$ and $2$ is set to $65$, which is the effective period of our genetic algorithm implementation for population size $5$. The period for test $3$ is $1$, as explained above. Visual inspection does not reveal any significant differences between the true and pseudorandom sequences. Results for other period values and for random sequences with different seeds are of identical nature. \textit{This suggests that our observations of structural bias do not originate from the random generator but rather represent artefacts from the iterative application of algorithmic operators.}

In practice, the situation is slightly more complicated, since for some algorithms (like the genetic algorithm considered in this paper), not all examined points enter the population. This means that some of the dots shown in Fig. \ref{fig:rnd_tests} are sieved out and others are moved via a series of trivial parallel projections depending on whether or not the point constructed with these dots has entered the population based on its fitness values and particulars of the algorithm. At the current stage of our research, simulations involving such tracking is impractical and deemed of no particular value. 

Finally, while investigating known properties of pseudorandom generators, we stumbled upon a good example of a highly (structurally) biased algorithm. In what was a serious attempt by a skilled algorithm designer to design a "super-random" number generator, Donald Knuth came up with \textit{Algorithm K}, which turned out to have unexpected properties \cite{cit:Knuth1982}. Given a 10-digit decimal number, the algorithm functions as follows\footnote{The pseudocode is given here to demonstrate how complicated the algorithm is; there is no need to follow it in detail.}: 
\begin{enumerate}
\item Choose number of iterations. Set $Y \leftarrow \lfloor \frac{X}{10^9} \rfloor$, the most significant digit of $X$ (Steps 2 to 13 are executed exactly $Y+1$ times, that is randomizing transformations are applied a random number of times.)
\item Choose random step. Set $Z\leftarrow \lfloor X/10^8 \rfloor \mbox{ mod }10$, the second most significant digit of $X$. Go to step $(3+Z)$ (\ie jump to a random step).
\item Ensure $\geq 5\cdot10^9$. If $X<5000000000$, set $X\leftarrow X+5000000000$.
\item Middle square. Replace $X$ by $\lfloor X^2/10^5 \rfloor \mbox{ mod }10^{10}$.
\item Multiply. Replace $X$ by $1001001001X\mbox{ mod } 10^{10}$.
\item Pseudo-compliment. If $X<100000000$, then set $X\leftarrow X+9814055677$; otherwise set $X\leftarrow 10^{10}-X$.
\item Interchange halves. $X\leftarrow 10^5(X\mbox{ mod }10^5)+\lfloor X/10^5\rfloor$ \ie interchange the low-order five digits of $X$ with the high-order five digits.
\item Multiply. Same step as 5.
\item Decrease digits. Decrease each nonzero digit of the decimal representation of $X$ by one.
\item $99999$ modify. If $X<10^5$, set $X\leftarrow X^2+99999$; otherwise set $X\leftarrow X-99999$.
\item Normalize. (At this point $X$ cannot be zero.) If $X<10^9$, set $X\leftarrow 10X$ and repeat this step.
\item Modified middle square. Replace $X$ by $\lfloor X(X-1)/10^5\rfloor\mbox{ mod }10^{10}$.
\item Repeat? If $Y>0$, decrease $Y$ by $1$ and return to step 2. If $Y=0$, the algorithm terminates with $X$ as the desired "random" value${}_{\square}$
\end{enumerate}
Initial tests of this generator revealed that, depending on the starting value, the output of this algorithm is far from being "super-random": it either converges to the 10-digit value $6065038420$, or the sequence begins to repeat itself after $7401$ values, in a cyclic period of length $3178$ \cite{cit:Knuth1982}. This example strengthens the view that thoughtlessly assembled overcomplicated algorithms have elevated chances of possessing undesirable and intractable properties. 

\section{Structural analysis of a simplified genetic algorithm}\label{sect:sbiasGA}
In this section we are going to look at a simplified version of a genetic algorithm, and theoretically analyse this algorithm with a view to uncovering dynamics that may cause structural bias. Our simplifications will allow us to analyse changes in the sample variance of the positions of points in the population when we generate a new point (and replace one from the current population). However, the simplifications, though necessary to facilitate analysis, do not materially change the performance of the algorithm on a function such as $f_0$, as we explain later with both heuristic arguments and numerical experiments.

Consider a genetic algorithm, as in Section \ref{sect:GA}, with the following amendments to its operation.
\begin{enumerate}
\item Selection is uniformly random -- i.e. there is a purely random choice of parents;
\item the child replaces a randomly chosen member of the population.
\end{enumerate}

More precisely, we define a process $\{\underline{X}(t)\}_{t \in \mathbb{Z}_+}$, where $\underline{X}(t) \in \mathbb{R}^N$ for each $t$ and $X_i(0)$ is uniformly distributed in $[0,1]$ for each $i=1,..,N$. The change from time $t$ to time $t+1$ is as follows:

\begin{itemize}
\item Pick two numbers from $1$ to $N$ at random (with replacement). Let these numbers be $j$ and $k$.
\item Generate a new coordinate
\begin{equation} \label{eq:y}
Y = \left( \min\left(\alpha X_j + (1-\alpha)X_k + Z, 1\right)\right)^+,
\end{equation}
where $x^+ = \max(x,0)$, $\alpha$ is a random variable uniformly distributed on $(-d, 1+d)$ for a positive $d$ and $Z$ is a Normal random variable with mean $0$ and variance $\sigma^2$.

This represents a choice of a new point which is absorbed at the boundaries.

\item Pick a number from $1$ to $N$ at random; let this number be $i$.
\item Replace $X_i$ by $Y$.
\end{itemize}

Let $S^2(t)$ denote the sample variance of the vector $\underline{X}(t)$
$$
S^2(t) = \frac{1}{N-1}\left(\sum X_i(t)^2 - \frac{\left(\sum X_i(t)\right)^2}{N}\right).
$$

We can prove the following theorem.

\begin{theorem} \label{thm:absorb}
If
$$
d < \frac{-1+\sqrt{\frac{3N+9}{N-1}}}{2},
$$
then there exist $0 < K < \infty$ and $\varepsilon > 0$ such that if $S^2(t) > K$, then
$$
\mathbf{E} \left(S^2(t+1) - S^2(t)|\underline{X}(t)\right) < -\varepsilon.
$$
\end{theorem}

{\bf We prove} the theorem by bounding the sample variance of the real next step by the sample variance of the next step without absorption at the boundaries. Indeed, it is immediate to check that the sample variance of the non-absorbed values at the next step is an upper bound for the sample variance of the original (possibly absorbed) values, and the difference between (non-absorbed) sample values at two subsequent steps is equal to
\begin{eqnarray*}
(N-1) \mathbf{E} \left(S^2(t+1) - S^2(t)|\underline{X}(t)\right) = &\\
\frac{1}{N^3}\sum_{j,k,l} (S_2-X_l^2+\mathbf{E}Y^2) \quad - & \frac{1}{N}\mathbf{E} \left(\frac{1}{N^3}\sum_{j,k,l} (S_1-X_l+Y)^2 \right) - S_2 + \frac{1}{N}S_1^2,
\end{eqnarray*}
where $Y$ is defined in \eqref{eq:y}, $S_2 = \sum X_i(t)^2$ and $S_1 = \sum X_i(t)$. The remainder of the argument consists in re-arranging terms and noting that $\mathbf{E} U = 1/2$ and $\mathbf{E} U^2 = \dfrac{1+d+d^2}{3} {}_{\square}$

{\bf Note} that
$$
K = \frac{\sigma^2(1-1/N)}{N+1-2 \frac{1+d+d^2}{3} (N-1)}.
$$

The theorem implies that if the sample variance of the points' locations is larger than $K$, then on average it will decrease. This, heuristically, means that the points will tend not to spread over the entire interval $[0,1]$. We conjecture that there is a stronger result showing that points' locations converge to a strict subset of $[0,1]$. This is supported by our numerical results but so far we have not demonstrated it theoretically. 

For vectors with $N$ components all taking values in $[0,1]$, it is clear that the largest value of the sample variance is $\frac{N}{2(N-1)}$ and is always bounded away from $0$ (in fact, converges to $1/2$ as $N \to \infty$). One can easily see, however, that $K \to 0$ as $N \to \infty$. This means that for a sufficiently large number of points in the population, the range of configurations in which the average change in variance is negative (i.e. configurations from which the points tend to become spread less at the next time instance than at the previous one) is not empty and becomes larger as the number of points increases.

Finally, before closing this section with a brief empirical test of the simplified algorithm that we have analysed, we note certain observations that follow from the theorem, and that we will refer to subsequently. First, numerical exploration of the expression of Theorem 1 with typical and reasonable values suggests that the implied 'reducing variance' dynamics may be commonplace in genetic algorithm designs. Further, and interestingly, as $N$ (population size) increases, the 'burden' on $d$ to be small increasingly relaxes, which suggests that structural bias will become more prominent at larger population sizes, despite perhaps high levels of exploration (larger $d$) among the algorithm's operators. We note that this expectation, for more prominence in structural bias at higher population sizes, resonates strongly with our empirical findings in Section \ref{sect:GA}. Next, considering that 'difficult' landscapes may tend to keep a population scattered across multiple local optima (at least in early or middle stages of a genetic algorithm's search), the theorem indirectly suggests that the consequent high position variance in such circumstances will exacerbate structural bias. In other words, the theorem provides a theoretical root suggesting that many kinds of 'difficult' landscapes (where we might expect high positional variance during search) will be sensitive to an algorithm's structural bias, while 'easy' landscapes (for which a good algorithm can be expected to focus quickly around optimal areas, with consequent low variance) will be relatively insensitive to an algorithm's structural bias. In Section \ref{sect:conseq}, after first presenting approaches to visualise and quantify structural bias, we perform experiments that allow us to start to evaluate these suggestions.

\begin{figure*} \centering
\subfigure[simplified GA $N=5$ start]{\includegraphics[width=.34\linewidth,angle=270]{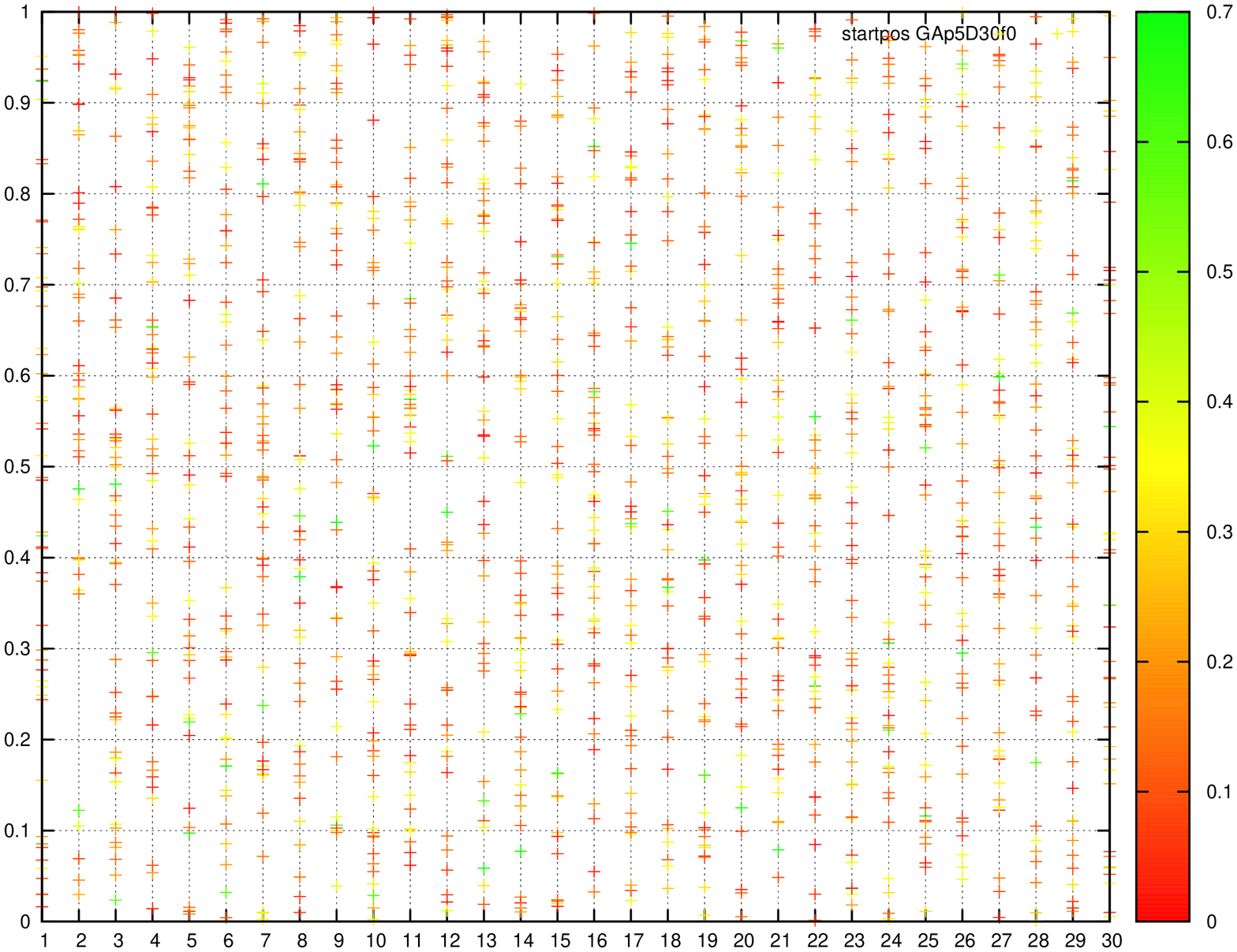}}
\subfigure[simplified GA $N=5$ end]{\includegraphics[width=.34\linewidth,angle=270]{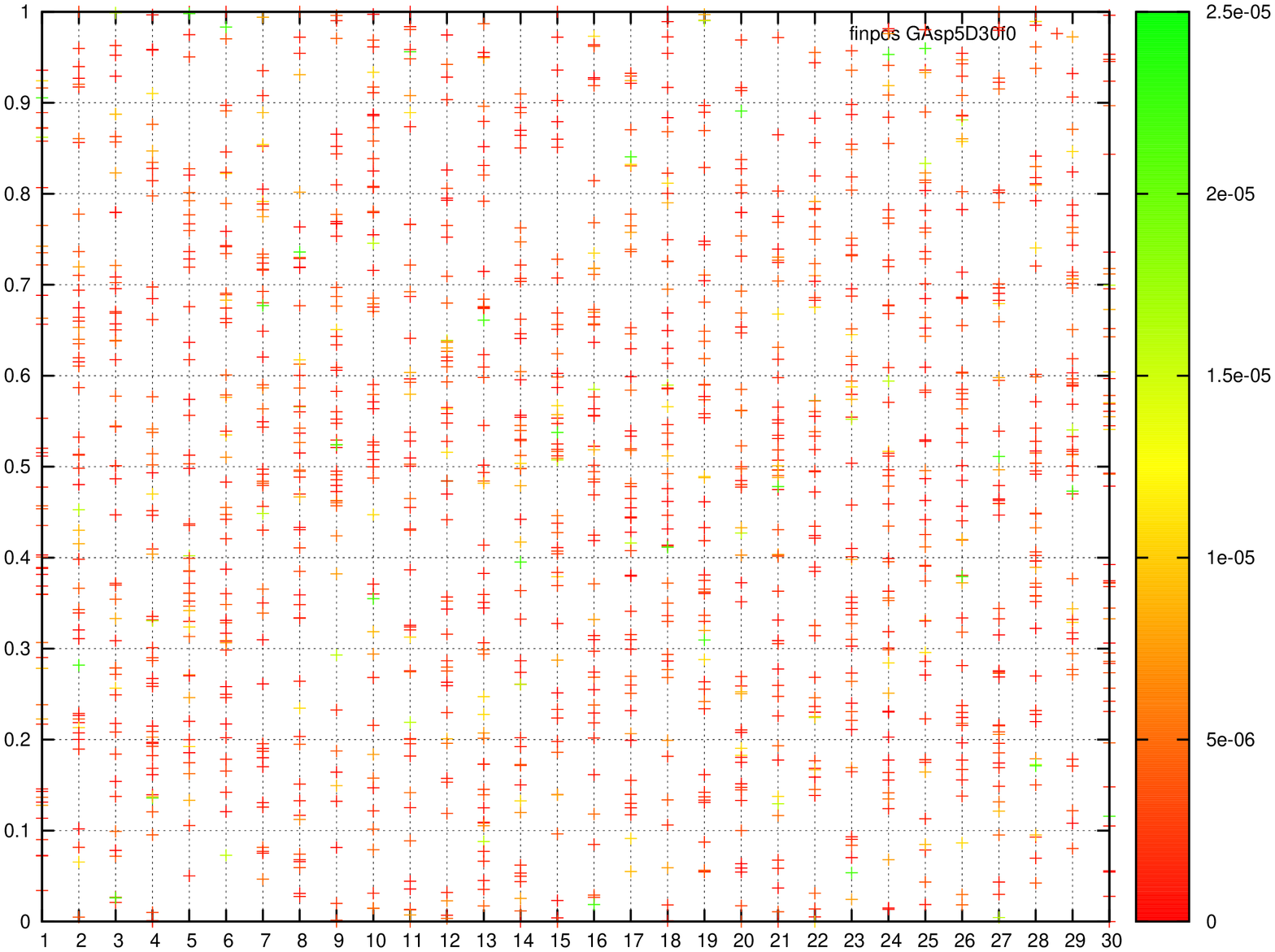}\label{fig:sGAcoors_finp5}}\\
\subfigure[simplified GA $N=20$ start]{\includegraphics[width=.34\linewidth,angle=270]{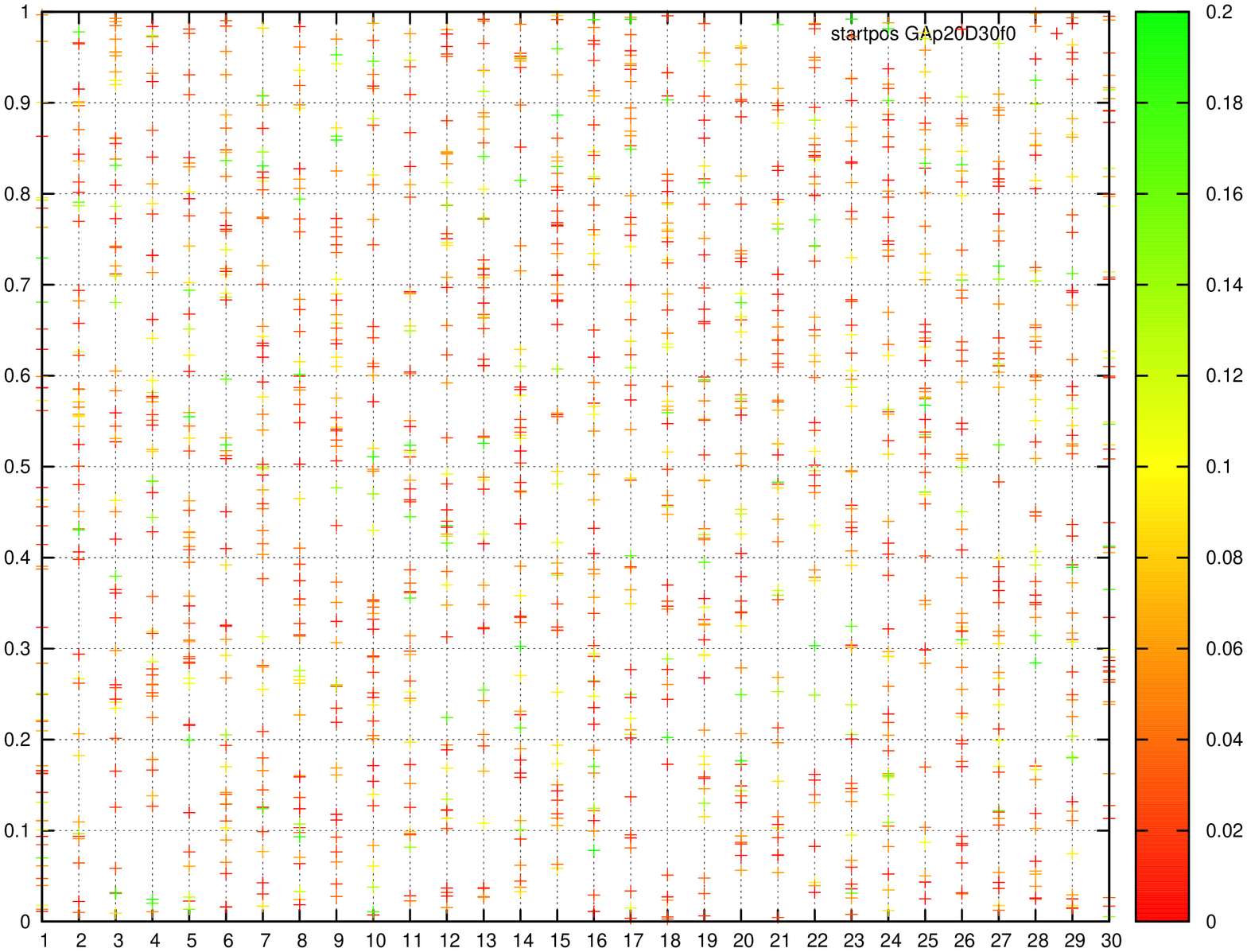}}
\subfigure[simplified GA $N=20$ end]{\includegraphics[width=.34\linewidth,angle=270]{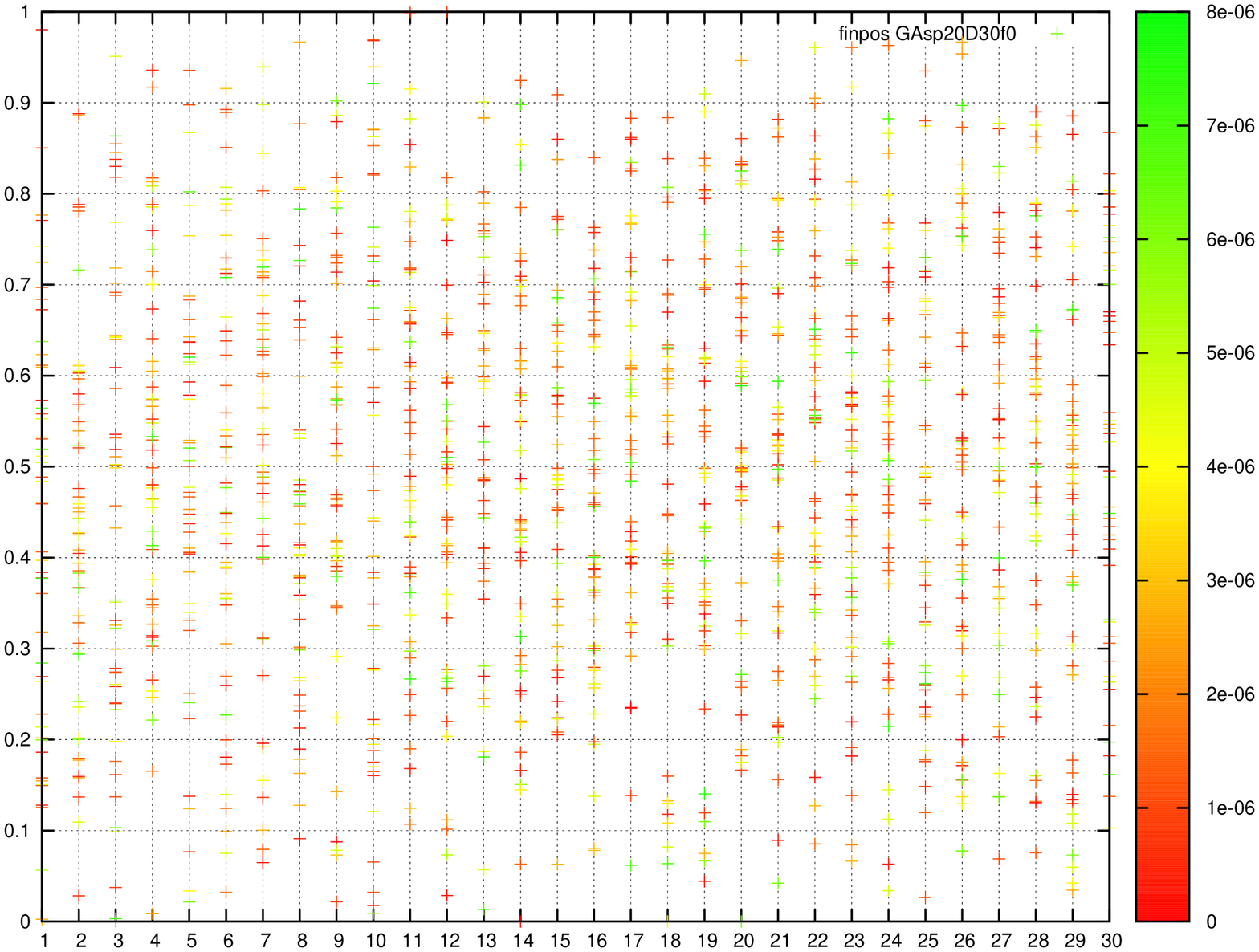}\label{fig:sGAcoors_finp20}}\\
\subfigure[simplified GA $N=100$ start]{\includegraphics[width=.34\linewidth,angle=270]{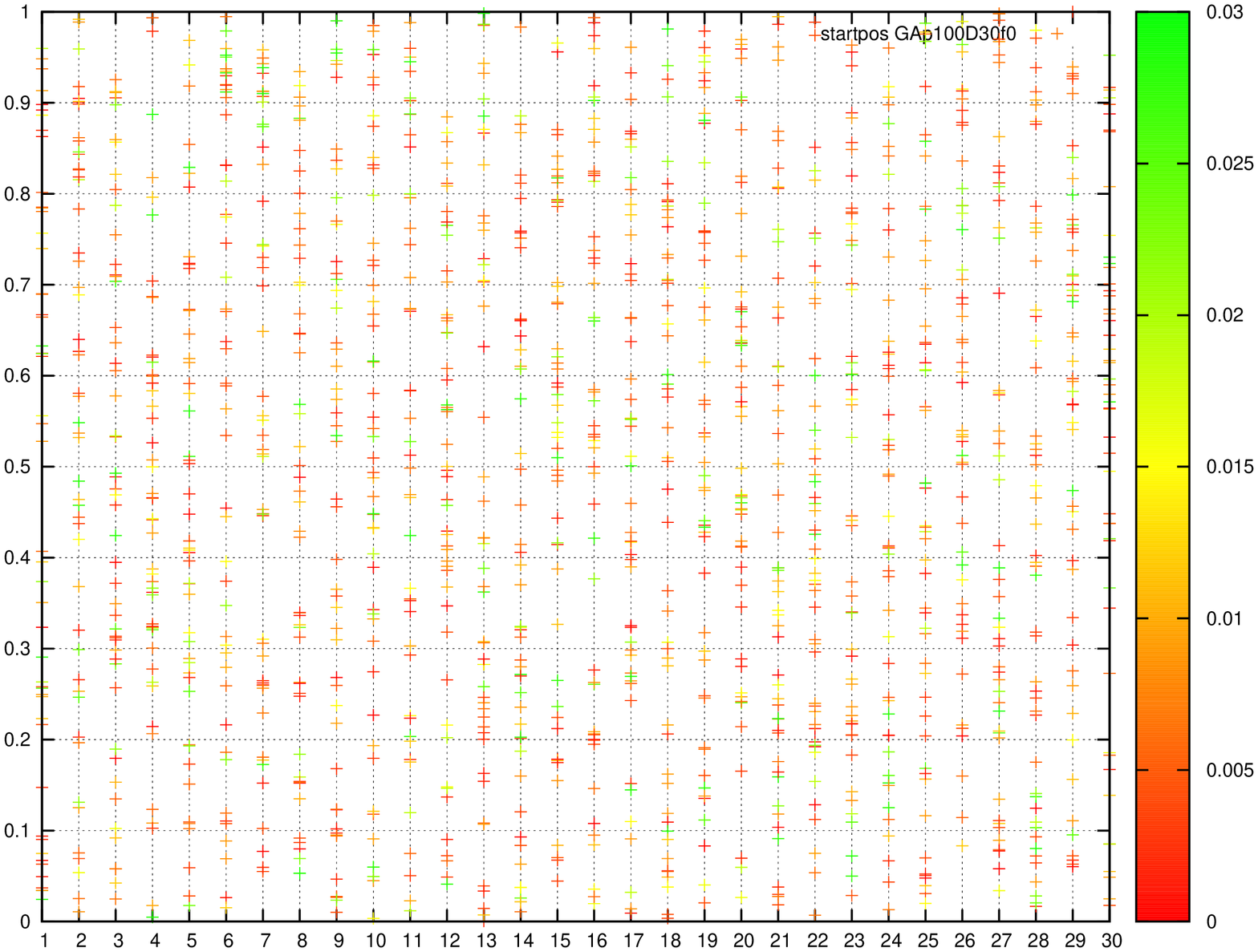}}
\subfigure[simplified GA $N=100$ end]{\includegraphics[width=.34\linewidth,angle=270]{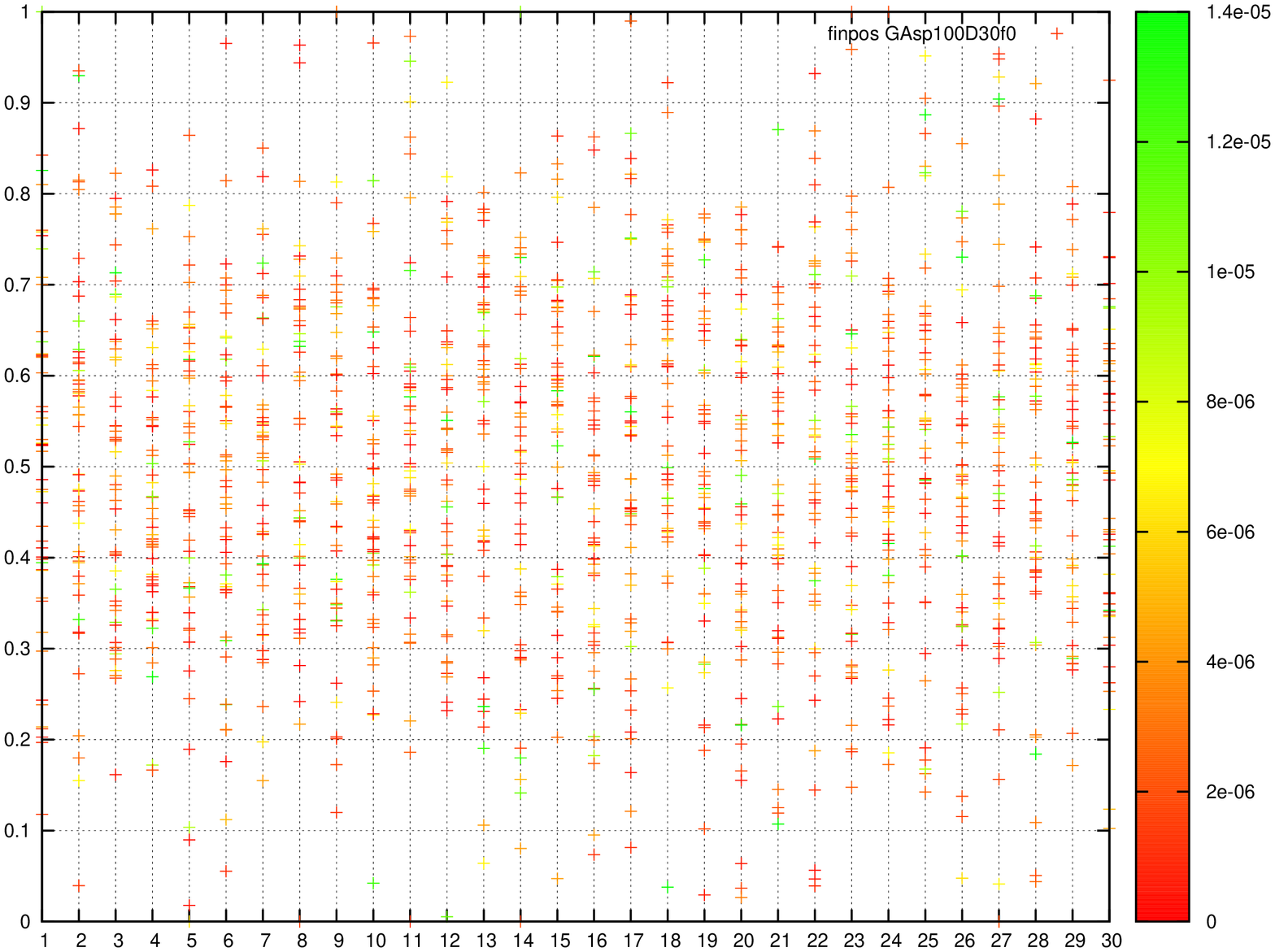}\label{fig:sGAcoors_finp100}}
\caption{Positions of points with the best fitness values in the first (left column) and the last (right column) populations of 50 runs of the \textit{simplified genetic algorithm} for different population sizes in parallel coordinates; horizontal axis shows the number of coordinate, vertical axis shows the range of these coordinate; the fitness value of each point is shown in colour. A clear bias towards the centre of the search space is visible in the last populations as population size increases.}\label{fig:sGApcoors}
\end{figure*}%
Numerically, the behaviour on $f_0$ of the simplified genetic algorithm is very similar to the behaviour of the typical genetic algorithm presented in Section \ref{sect:GA} as Figure \ref{fig:sGApcoors} shows. One can expect this due to simple heuristic arguments. Indeed, given that fitness at every step is chosen from a uniform distribution, independently of the fitness of all other points, and a point will only be accepted if its fitness is better than that of at least one existing point, the fitnesses of all points will converge to the optimal one. Therefore removing a random point instead of the worst one should not strongly influence the performance of the algorithm. The same concerns the choice of parents. Thus, this analysis approximately describes the typical genetic algorithm.

\section{Quantifying structural bias and observing its consequences}\label{sect:conseq}
Returning briefly to analogies, let us consider a football team running trials for a new goal-keeper. Imagine that the final choice is to be made between two persons: one talented but rather lazy keeper who prefers to stand still beside the left goalpost, no matter what the actions of the striker, and one very energetic keeper who can reach every part of the goal but occasionally fails. In this analogy, we intend the goal to represent the problem domain, the goal-keeper plays the role of the algorithm and the strategy of the striker, unknown to the goal-keeper, represents the objective function. It is then the duty of the goal-keeper to locate as close as possible a position where the ball is going to approach the goal, just as the algorithm needs to identify the region of the goal which contains an optimum of the current objective function. In life, it can happen that, by pure luck, the striker is equally limited and can hit only the region of the left goalpost. Clearly, our lazy goalkeeper will have no problem defending the goal from such a striker. However, as it usually happens that strikers tend to target different regions of the goal, a more flexible goalkeeper will end up being a better choice for the team regardless of his or her occasional shortcomings.

We propose, first of all, a simple visual test for structural bias that amounts to visualising the performance of the goal-keeper in such a trial. Our visual test is meant to identify whether or not an algorithm has any structural bias, or to compare the degrees of such bias among a suite of algorithms. In this test, conclusions can be made based on the distribution of coordinates of points with the best fitness values in the final populations of the algorithms under consideration running on $f_0$ for roughly the same fitness evaluation budget as intended for their deployment on real objective functions.

Application of this visual test to the algorithms explored in Section \ref{sect:numres} amounts to observing the parallel coordinates figures presented earlier, thereby inspecting the distributions of the positions of the $50$ best points (the best point from each of $50$ independent trials), for each of the algorithm configurations. Such inspection suggests that an appreciable level of structural bias is exhibited by the genetic algorithm with population size $100$ (Fig. \ref{fig:GAcoors_finp100}) , and by PSO with population sizes of both $N=5$ and $N=100$ (Figs. \ref{fig:PSOcoors_finp5}, \ref{fig:PSOcoors_finp100}). Meanwhile, the genetic algorithm with $N=20$ exhibits milder structural bias, see Fig. \ref{fig:GAcoors_finp20}. The remaining two cases -- the genetic algorithm with $N=5$ and PSO with $N=20$ -- seem to provide satisfactory performance in terms of structural bias, but are clearly more difficult to differentiate objectively based on a purely visual test of Figs. \ref{fig:GAcoors_finp5} and \ref{fig:PSOcoors_finp20}.

\begin{figure*} \centering
\includegraphics[width=0.7\linewidth]{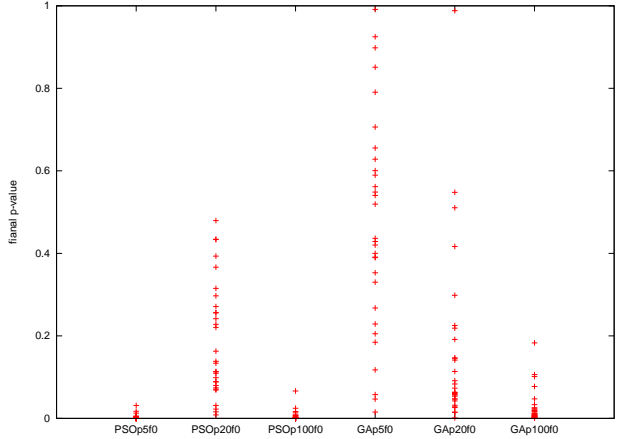}
\caption{Results of the Kolmogorov-Smirnov tests for the considered genetic algorithm and PSO, with three population sizes each. For each algorithm, the test is run independently for each dimension of the problem; the $p$-value returned by each test is shown with a marker. The $p$-values in the first, third and sixth columns are significantly lower than others, which translates into stronger structural bias present in results for these algorithms. $p$-values shown in the second column correspond to the case of milder structural bias, meanwhile the fourth and fifth columns characterise algorithms with the weakest structural bias observed in our series of experiments. These results support our conclusions regarding strength of structural bias based on purely visual analysis of Figs. \ref{fig:GApcoors} and \ref{fig:PSOpcoors}.}\label{fig:kstest_f0}
\end{figure*}%
For an objective test of the level of structural bias, we propose use of the Kolmogorov-Smirnov test \cite{cit:Kolmogorov1933}, \cite{cit:Smirnov1944}, \cite{cit:Stephens1974}, specifically to compare the empirical distribution function of the sample of coordinates and the cumulative distribution function of the uniform distribution. By using the Kolmogorov-Smirnov test in this way, we obtain a $p$-value that expresses the probability that the sample comes from a uniform distribution given the null-hypothesis is correct. Fig. \ref{fig:kstest_f0} summarises the results of this test, performed independently for each dimension, for the same sets of points discussed earlier in the context of visual tests. These results numerically support our aforementioned conclusions based on visual analysis of Figs. \ref{fig:GApcoors} and \ref{fig:PSOpcoors}. For example, the cluster of low $p$-values plotted against ''PSOp5f0'' corresponds to the observation of high levels of structural bias for PSO with $N=5$. 

Our proposed approach to quantifying an algorithm's inherent structural bias therefore comprises running repeated trials of the algorithm(s) in question using $f_0$ as the objective function, and subsequently applying one or both of: parallel-coordinates based visual inspection of the final points, and the Kolmogorov-Smirnov test to assess the uniformity of the distribution of those points. The proposed method is computationally highly efficient in comparison to approaches (such as that of \cite{cit:Davarynejad2014}) that require numerous optimization trials with the actual objective function (and also needing to be re-applied for every new objective function of interest). Our proposed approach is potentially suitable as an algorithmic design tool for general use. It is important to note that, on its own, a strategy of maintaining a more even coverage of the search space by the algorithm does not ensure a satisfactory algorithmic design capable of fast convergence to a near-optimum solution. The sole objective of such a strategy is to identify a combination of operators that forces the algorithm to explore the domain with more equal probability. This strategy is therefore complementary and should be used in conjunction with more comprehensive design strategies which ensure other favourable properties of optimisation algorithms such as those discussed in Section \ref{sect:intro} or other properties specific to a particular class of algorithms.

\subsection{Further numerical results: consequences of structural bias on a suite of test functions}\label{sect:further_numerical}
To investigate the consequences of structural bias when one aims to optimize a standard test function, in this subsection we perform experiments using the CEC 2005 test function suite, which is widely used to test and compare algorithms in the field of evolutionary computation \cite{cit:Suganthan2005}. 
Exact specifications of the CEC 2005 functions can be found in \cite{cit:Suganthan2005}. The benchmark suite comes along with source code that allows users to treat the individual functions in the test suite as 'black box' functions that simply return a fitness value when given an $n$-dimensional coordinate. No other information is provided to the optimization algorithm, except for specifications of the range of the search domain. The benchmark suite makes its functions available for specific dimensionalities (e.g. $10$, $30$ and $50$). 

For the purpose of illustration, we select a limited number of functions from the CEC 2005 benchmark suite for which the genetic algorithm considered in this paper:
\begin{itemize}
\item $f_8$ shifted rotated Ackley function in $[-32,32]^{30}$ with global optimum on the bounds
\item $f_9$ shifted Rastrigin function in $[-5,5]^{30}$,
\item $f_{13}$ shifted expanded Griewank and Rosenbrock function in $[-5,5]^{30}$,
\item $f_{14}$ shifted rotated expanded Scaffer F6 function in $[-100,100]^{30}$,
\item $f_{21}$ rotated hybrid composition function in $[-5,5]^{30}$,
\item $f_{24}$ rotated hybrid composition function in $[-5,5]^{30}$.
\end{itemize}

\begin{figure*} \centering
\subfigure[GA $N=5$ $f_8$]{\includegraphics[width=.34\linewidth,angle=270]{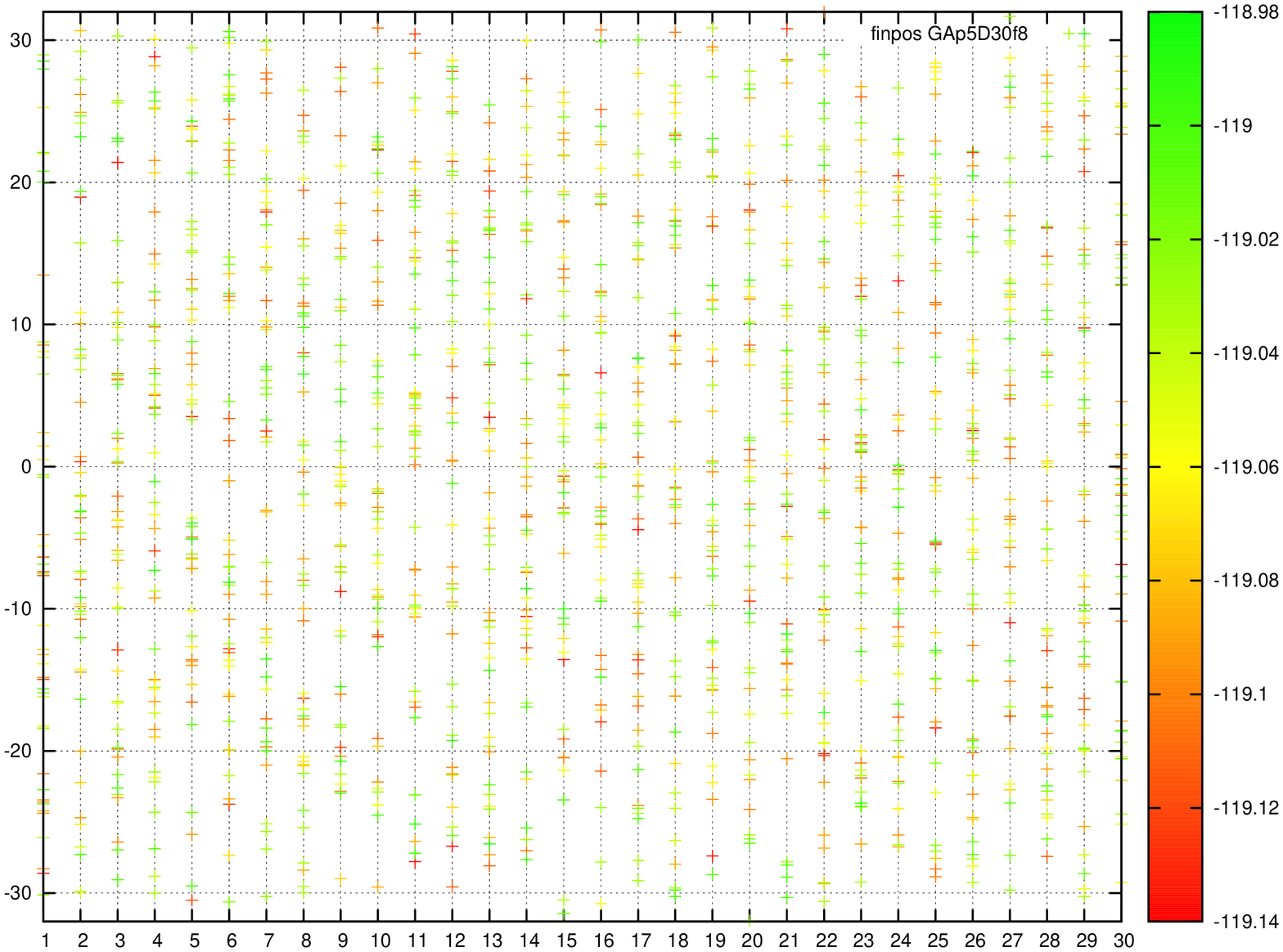}}
\subfigure[GA $N=5$ $f_9$]{\includegraphics[width=.34\linewidth,angle=270]{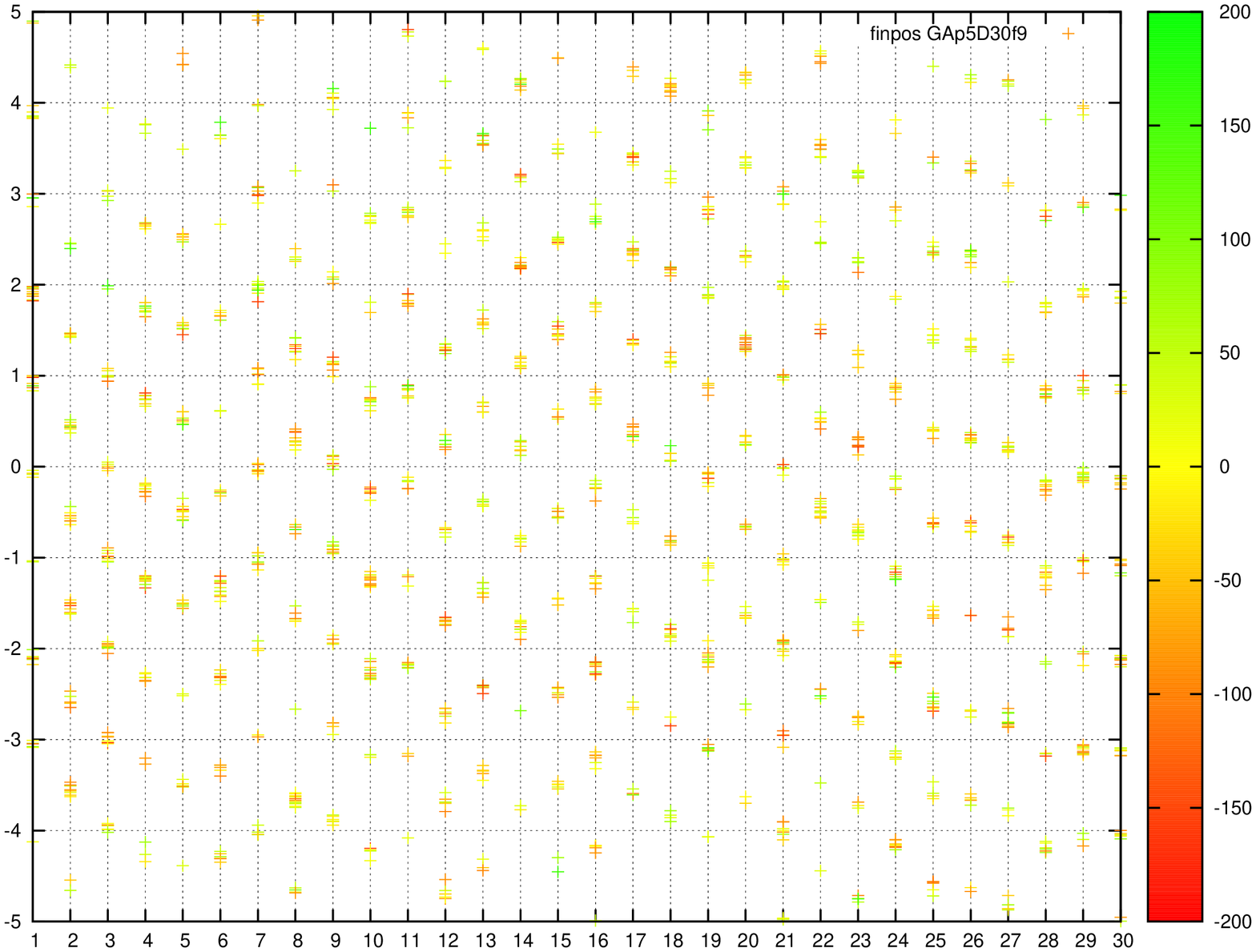}}\\
\subfigure[GA $N=20$ $f_8$]{\includegraphics[width=.34\linewidth,angle=270]{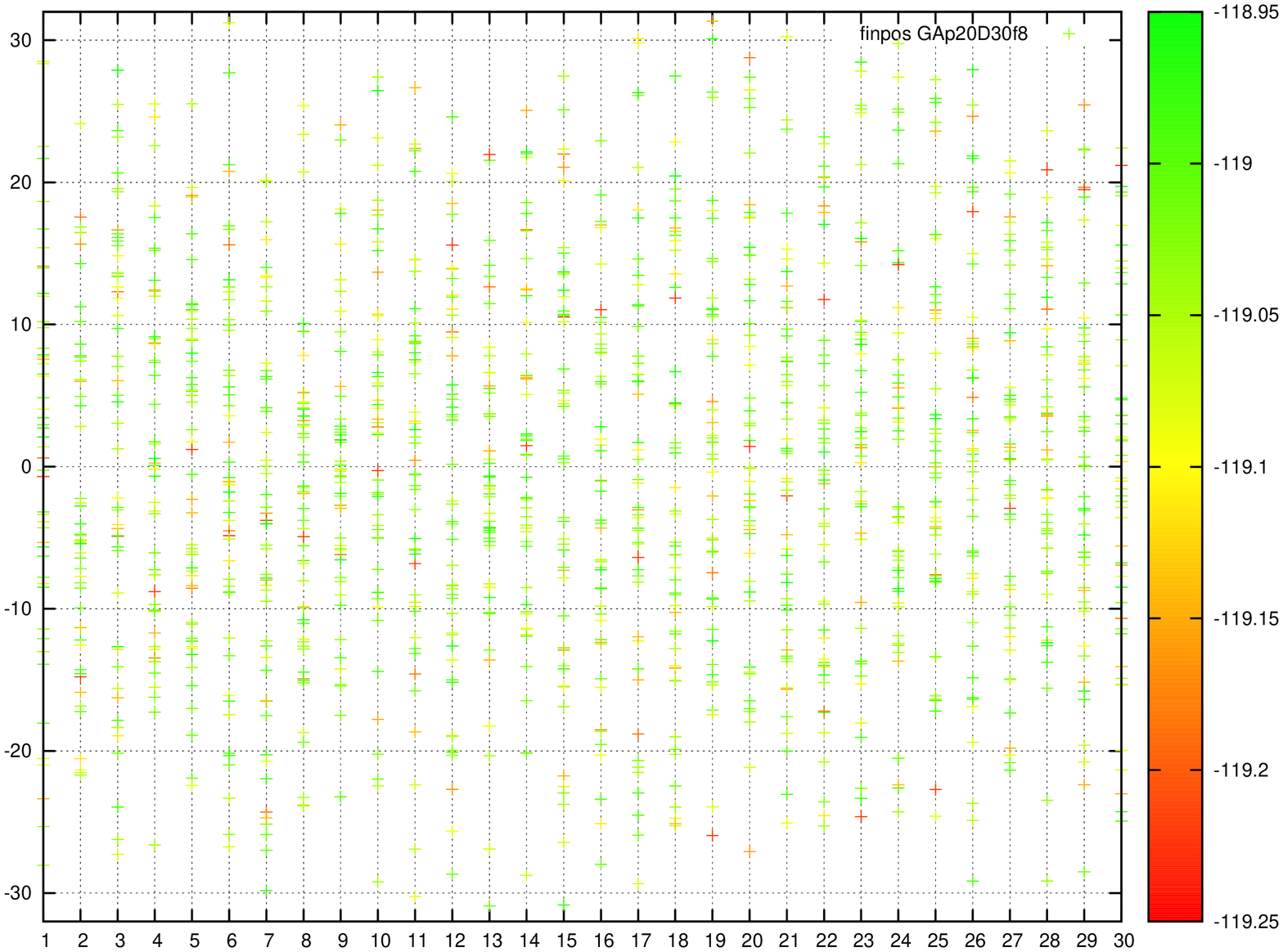}}
\subfigure[GA $N=20$ $f_9$]{\includegraphics[width=.34\linewidth,angle=270]{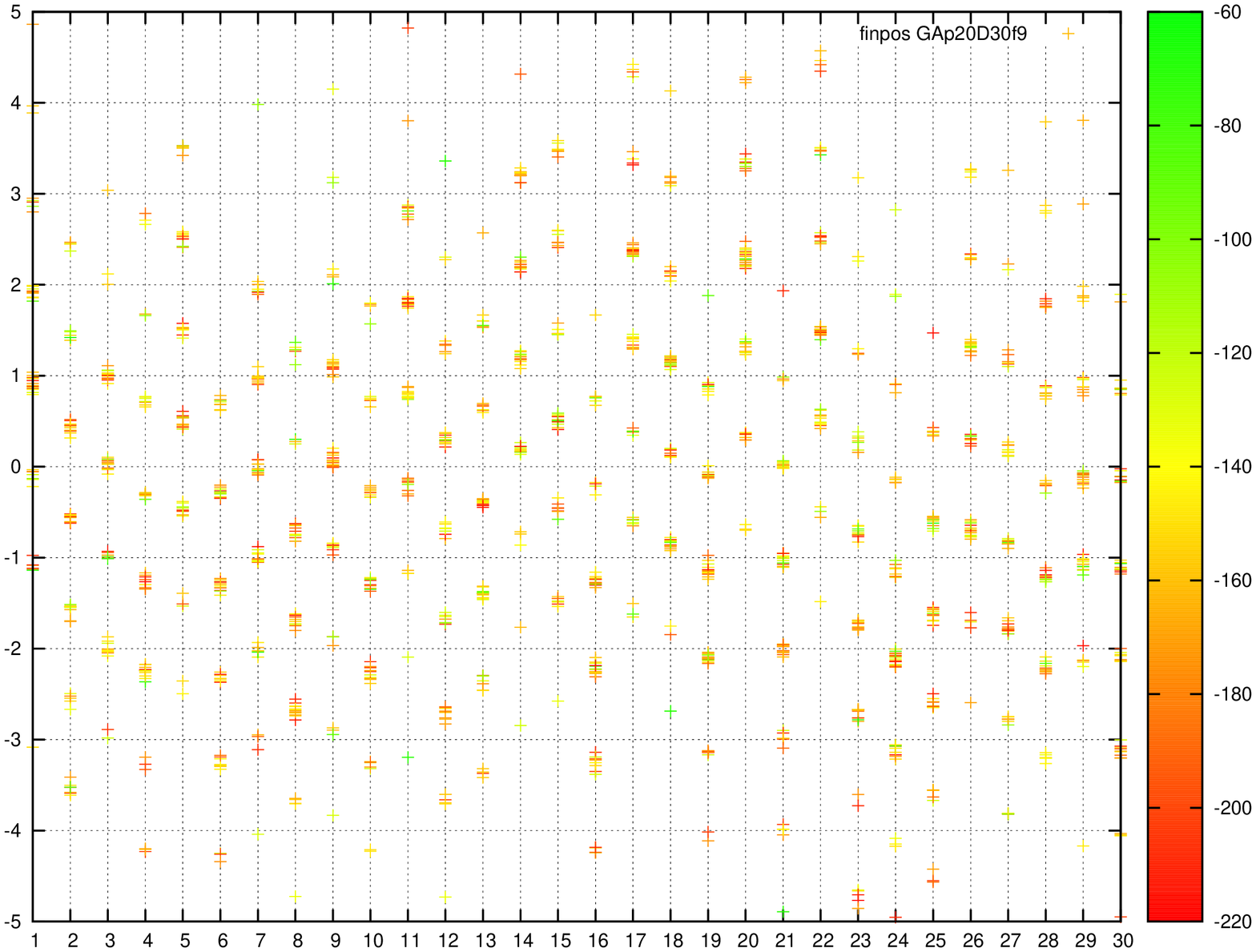}}\\
\subfigure[GA $N=100$ $f_8$]{\includegraphics[width=.34\linewidth,angle=270]{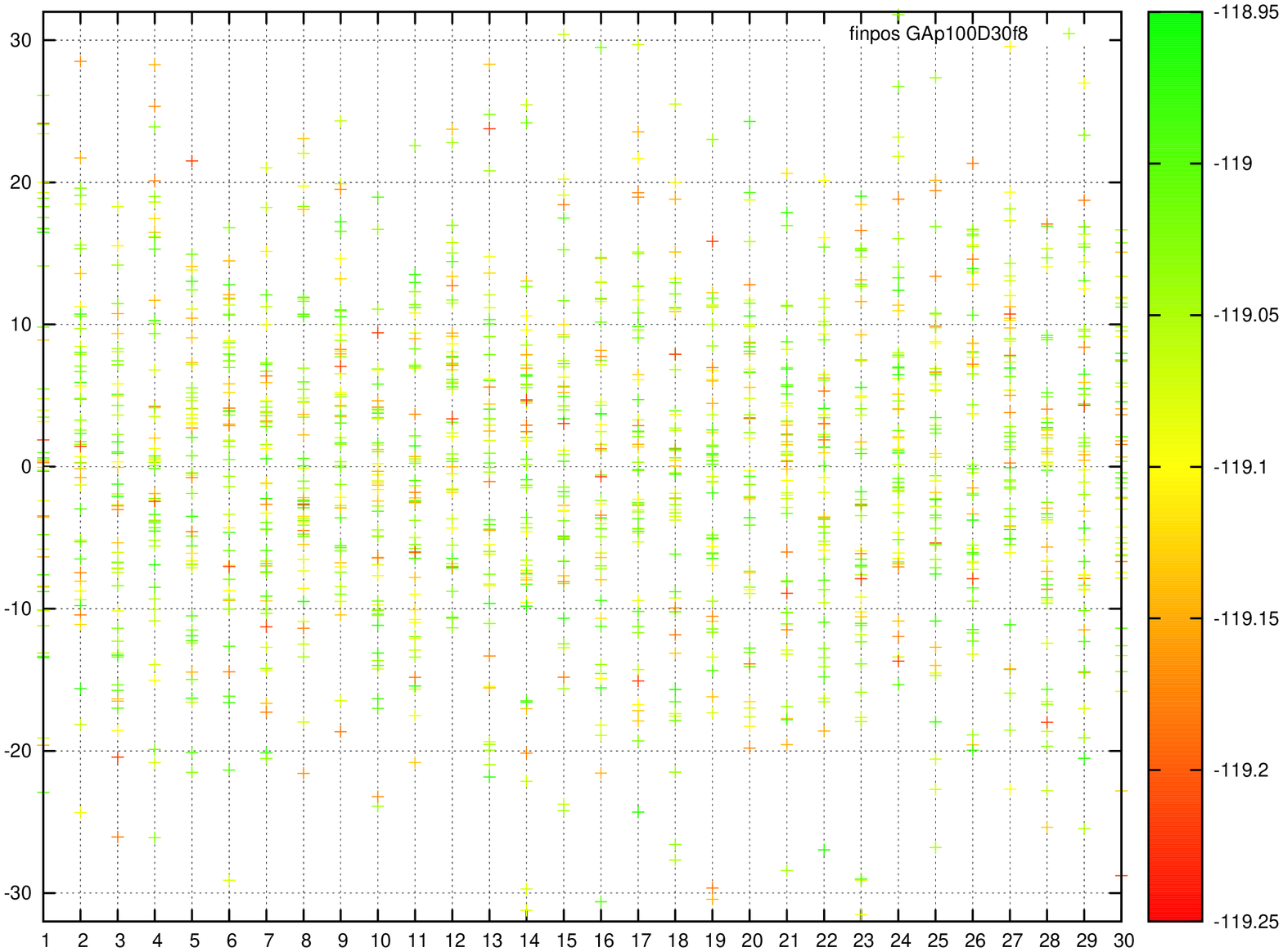}}
\subfigure[GA $N=100$ $f_9$]{\includegraphics[width=.34\linewidth,angle=270]{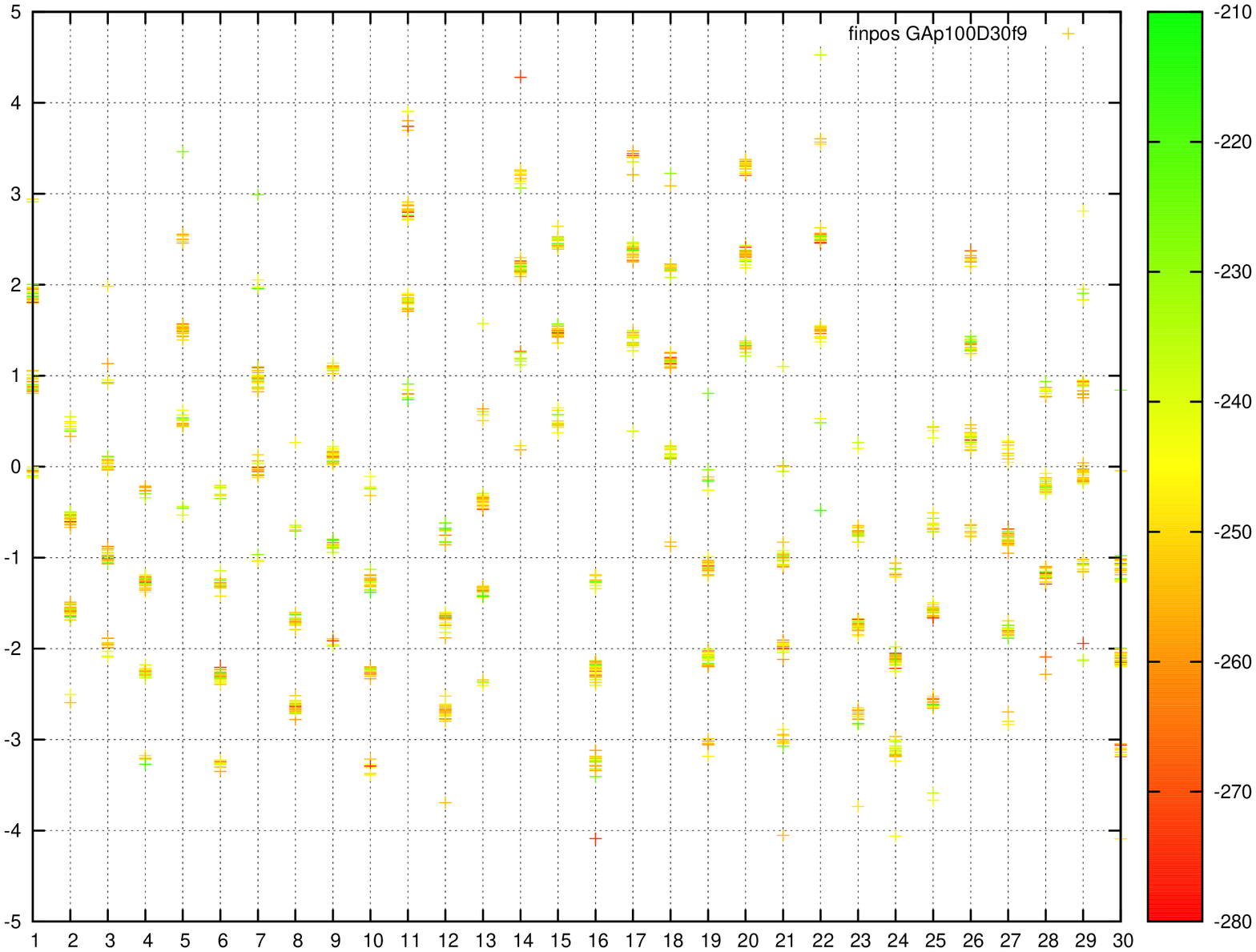}}
\caption{Positions of points with the best fitness in the last population of 50 runs of the considered GA for different population sizes in parallel coordinates - \textit{$f_8$ and $f_9$ are sensitive to structural bias of GA}. Horizontal axis shows the ordinal number of coordinate, vertical axis shows the full range of domain in this coordinate kept constant for each function; fitness value of each point is shown in colour.}\label{fig:GAf8f9pcoors}
\end{figure*}

\begin{figure*} \centering
\subfigure[GA $N=5$ $f_{13}$]{\includegraphics[width=.34\linewidth,angle=270]{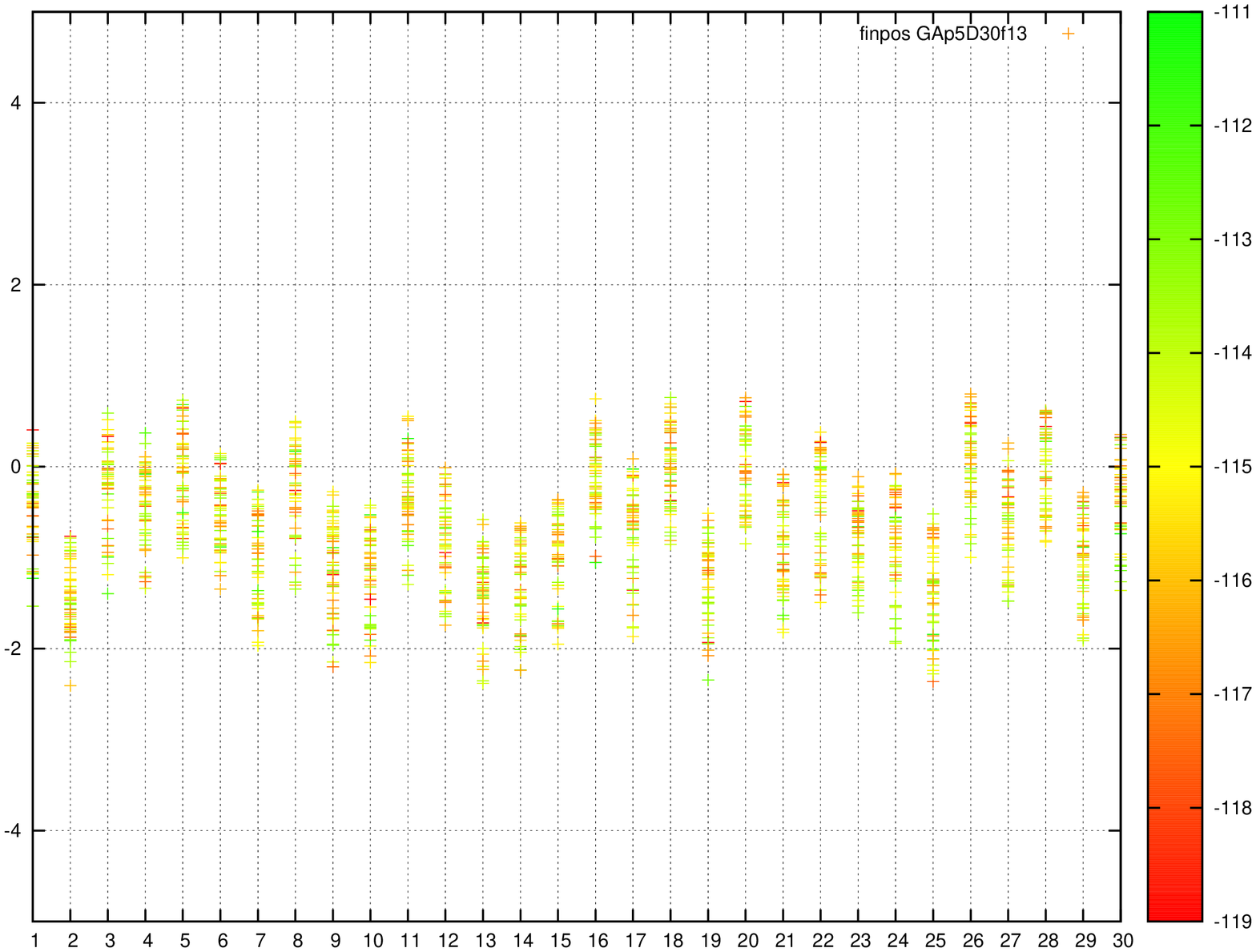}}
\subfigure[GA $N=5$ $f_{21}$]{\includegraphics[width=.34\linewidth,angle=270]{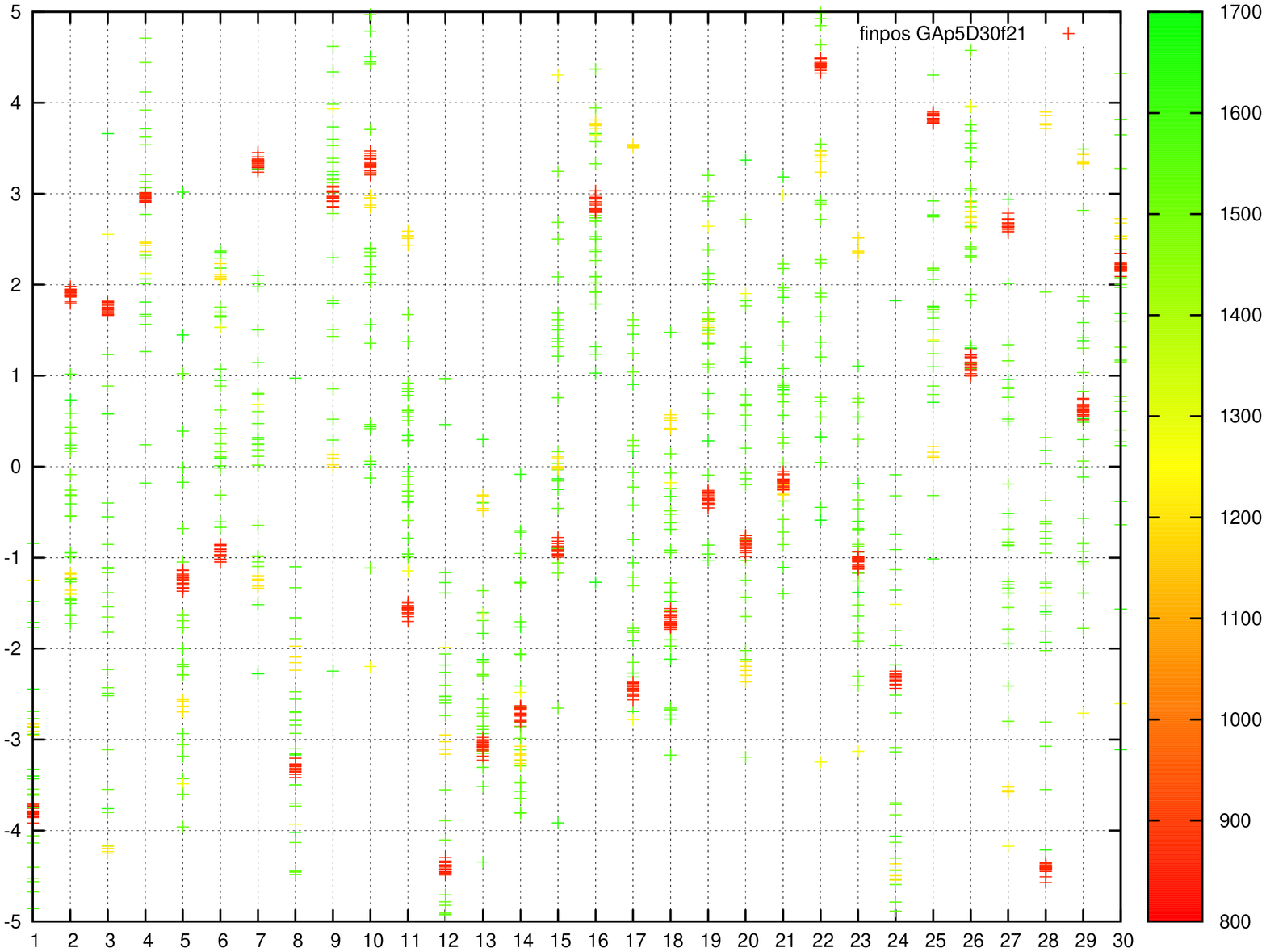}}\\
\subfigure[GA $N=20$ $f_{13}$]{\includegraphics[width=.34\linewidth,angle=270]{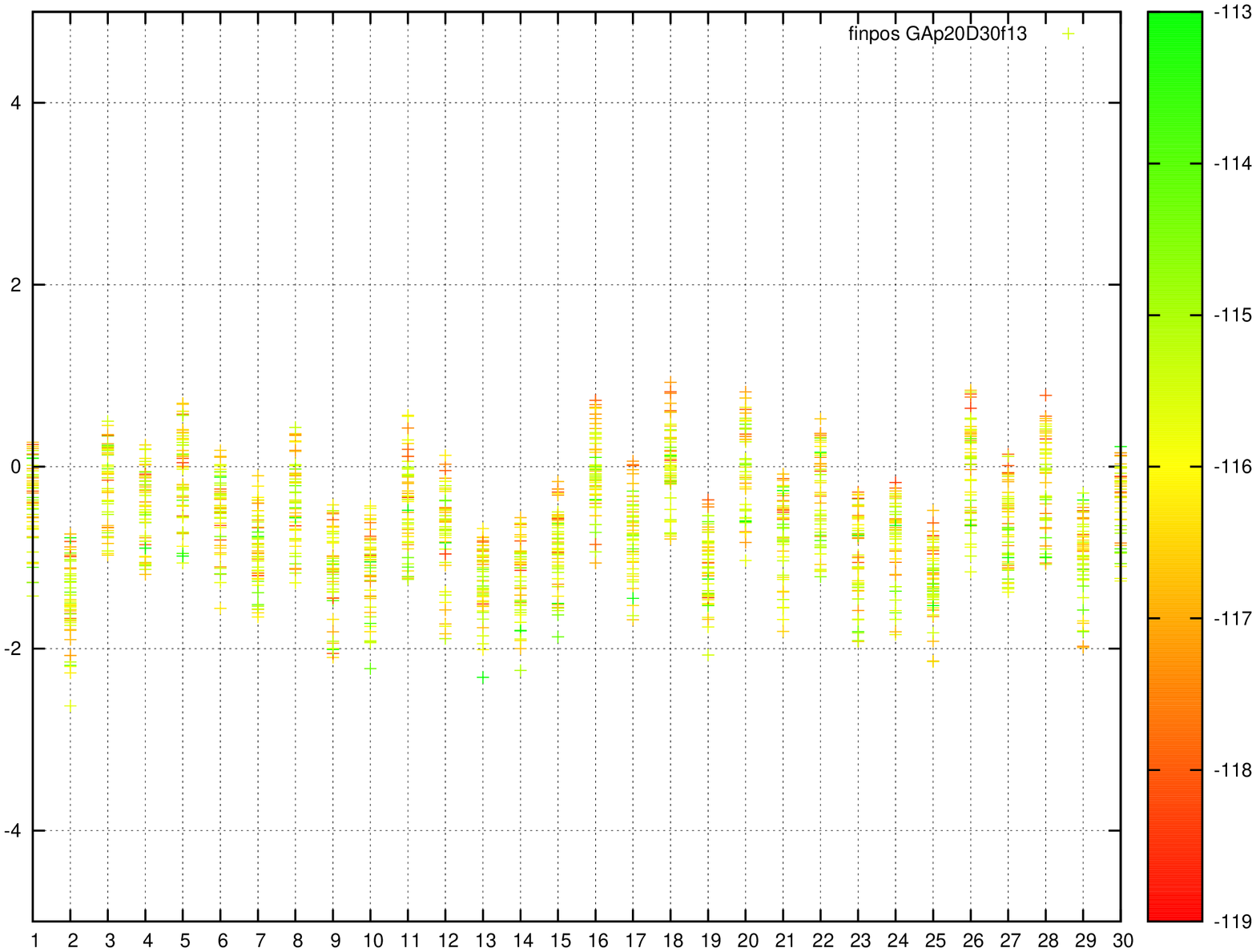}}
\subfigure[GA $N=20$ $f_{21}$]{\includegraphics[width=.34\linewidth,angle=270]{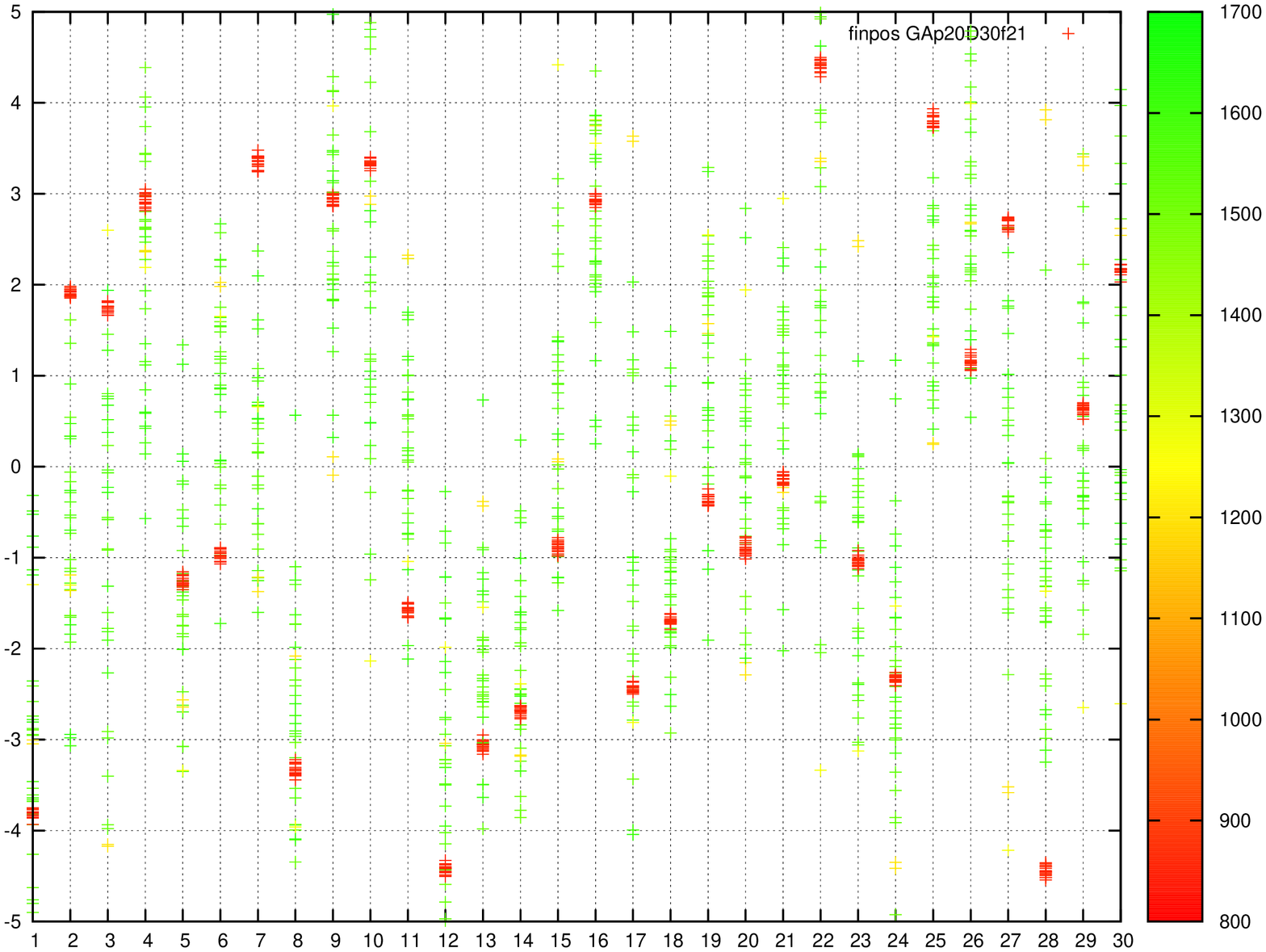}}\\
\subfigure[GA $N=100$ $f_{13}$]{\includegraphics[width=.34\linewidth,angle=270]{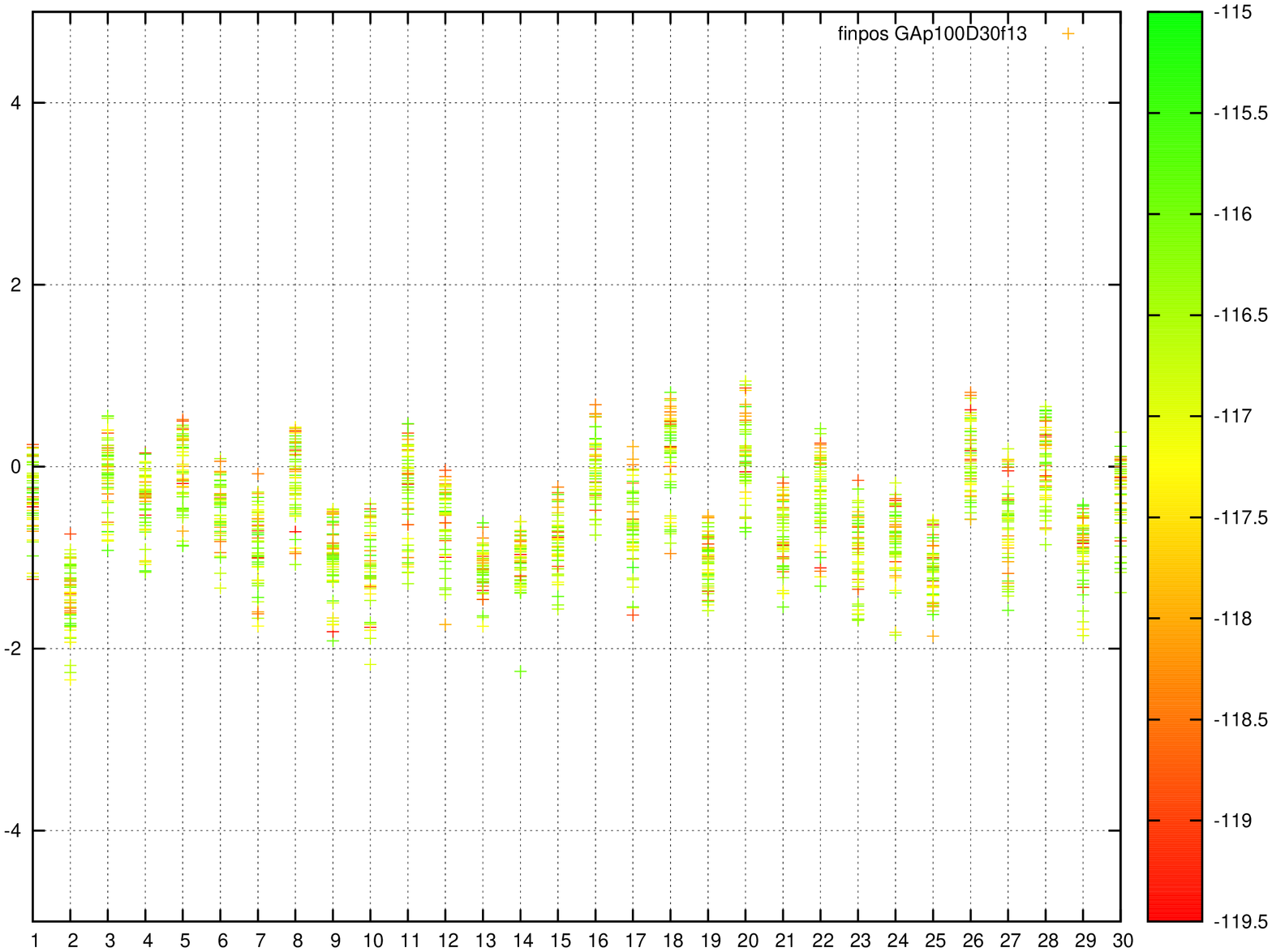}}
\subfigure[GA $N=100$ $f_{21}$]{\includegraphics[width=.34\linewidth,angle=270]{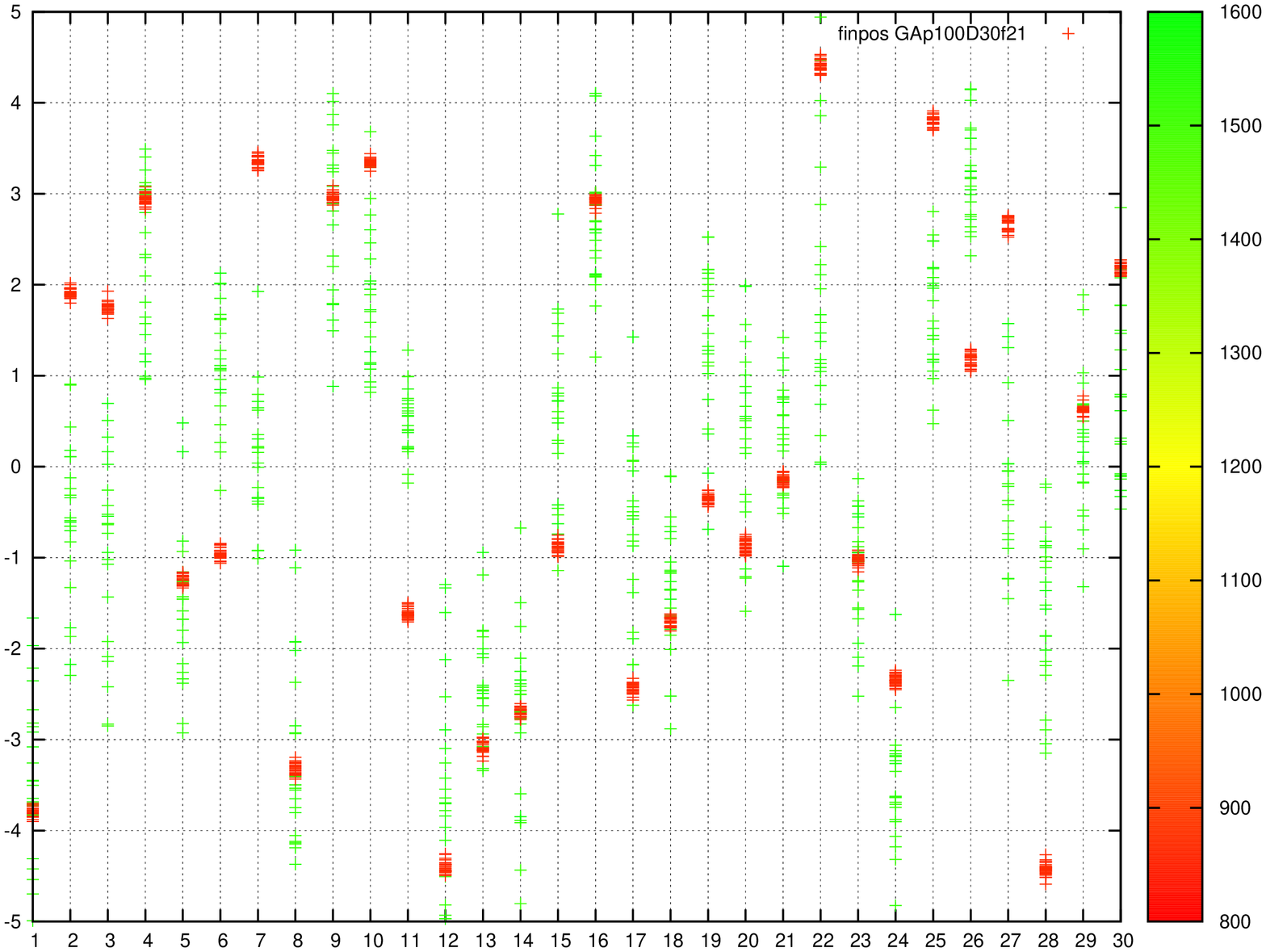}}
\caption{Positions of points with the best fitness in the last population of 50 runs of the considered GA for different population sizes in parallel coordinates - \textit{$f_{13}$ and $f_{21}$ are insensitive to structural bias of GA}. Horizontal axis shows the ordinal number of coordinate vertical axis shows the full range of domain in this coordinate kept constant for each function; fitness value of each point is shown in colour.}\label{fig:GAf13f21pcoors}
\end{figure*}

\begin{figure*} \centering
\subfigure[GA $N=5$ $f_{14}$]{\includegraphics[width=.34\linewidth,angle=270]{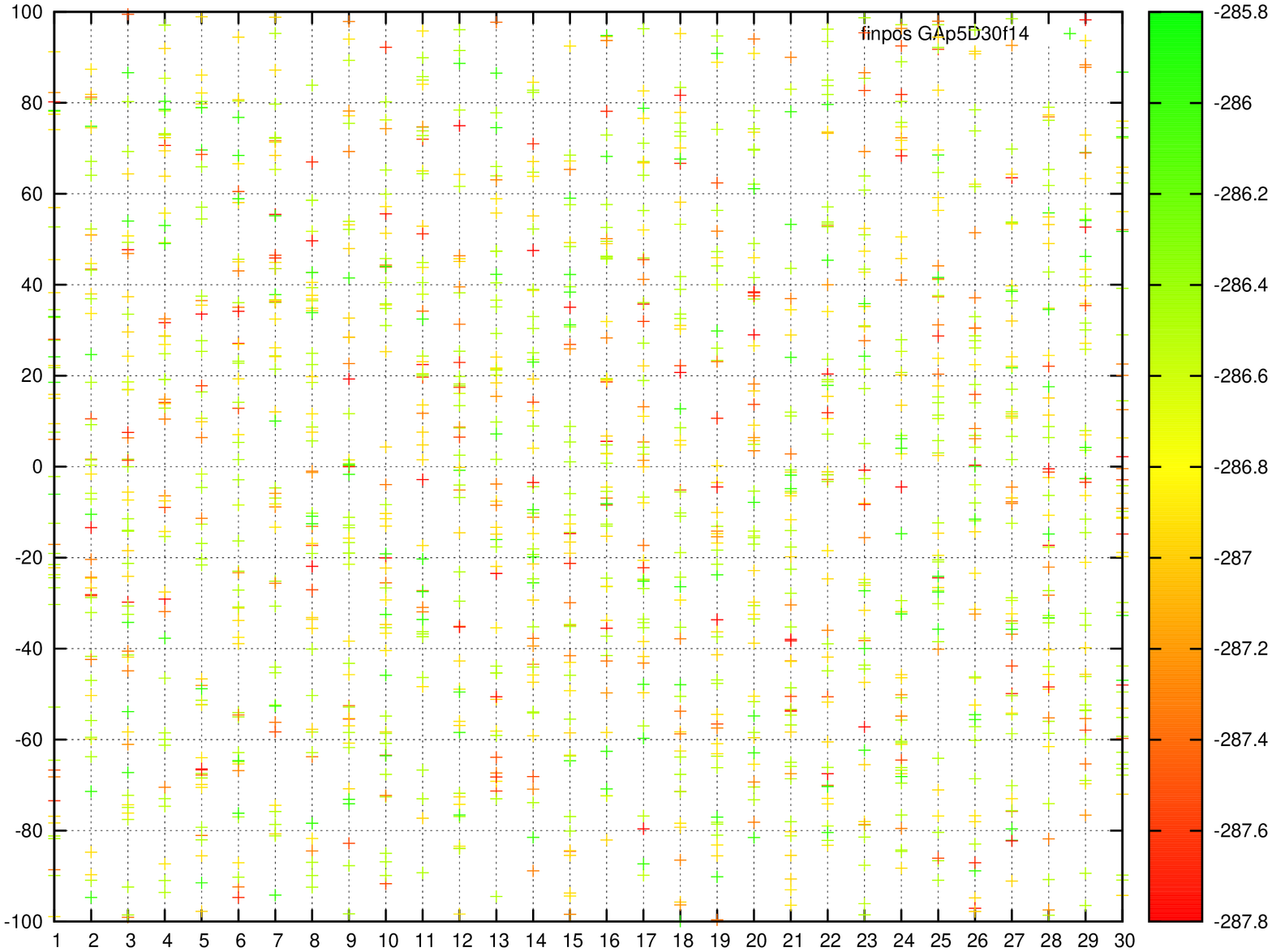}}
\subfigure[GA $N=5$ $f_{24}$]{\includegraphics[width=.34\linewidth,angle=270]{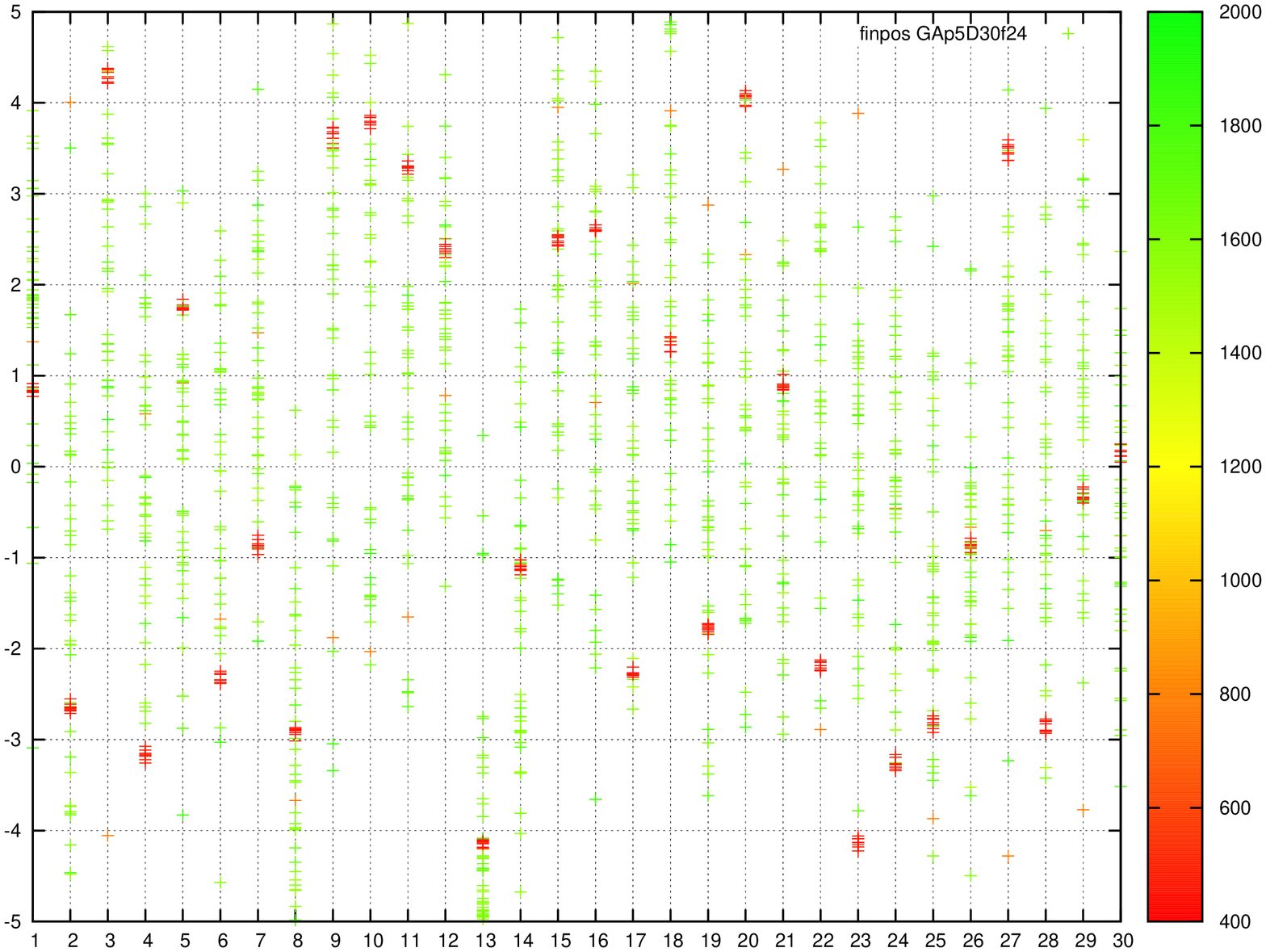}}\\
\subfigure[GA $N=20$ $f_{14}$]{\includegraphics[width=.34\linewidth,angle=270]{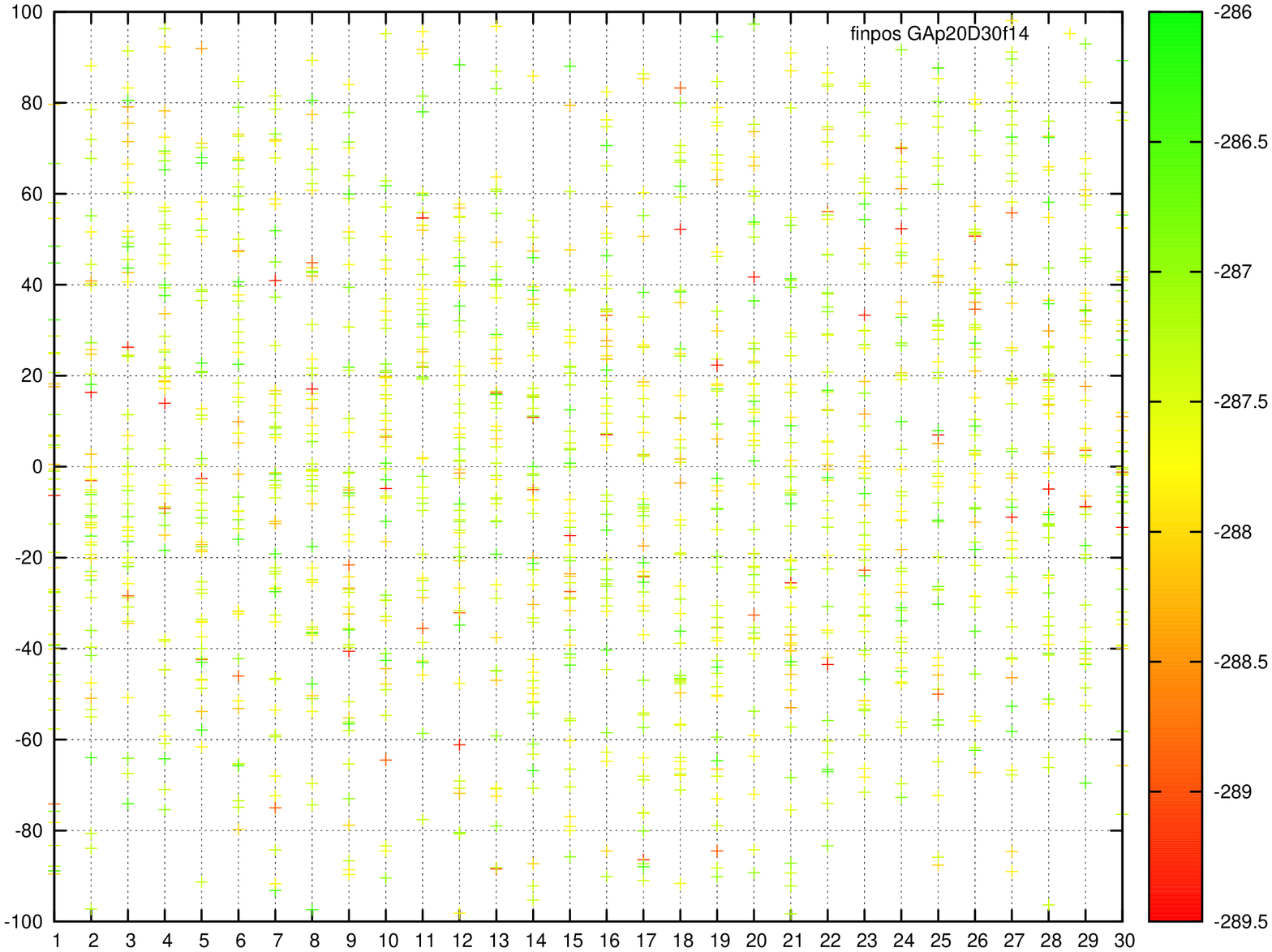}}
\subfigure[GA $N=20$ $f_{24}$]{\includegraphics[width=.34\linewidth,angle=270]{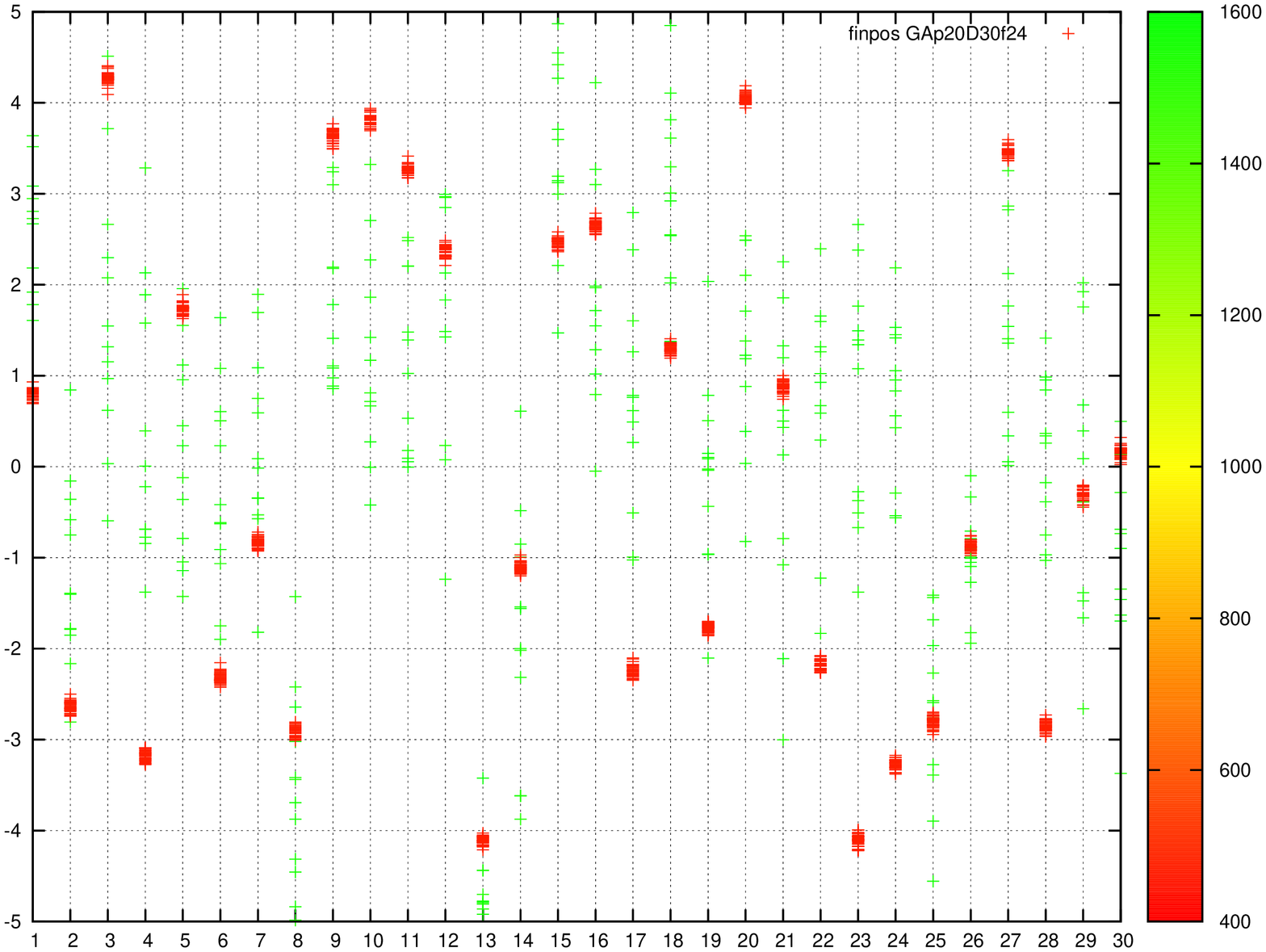}}\\
\subfigure[GA $N=100$ $f_{14}$]{\includegraphics[width=.34\linewidth,angle=270]{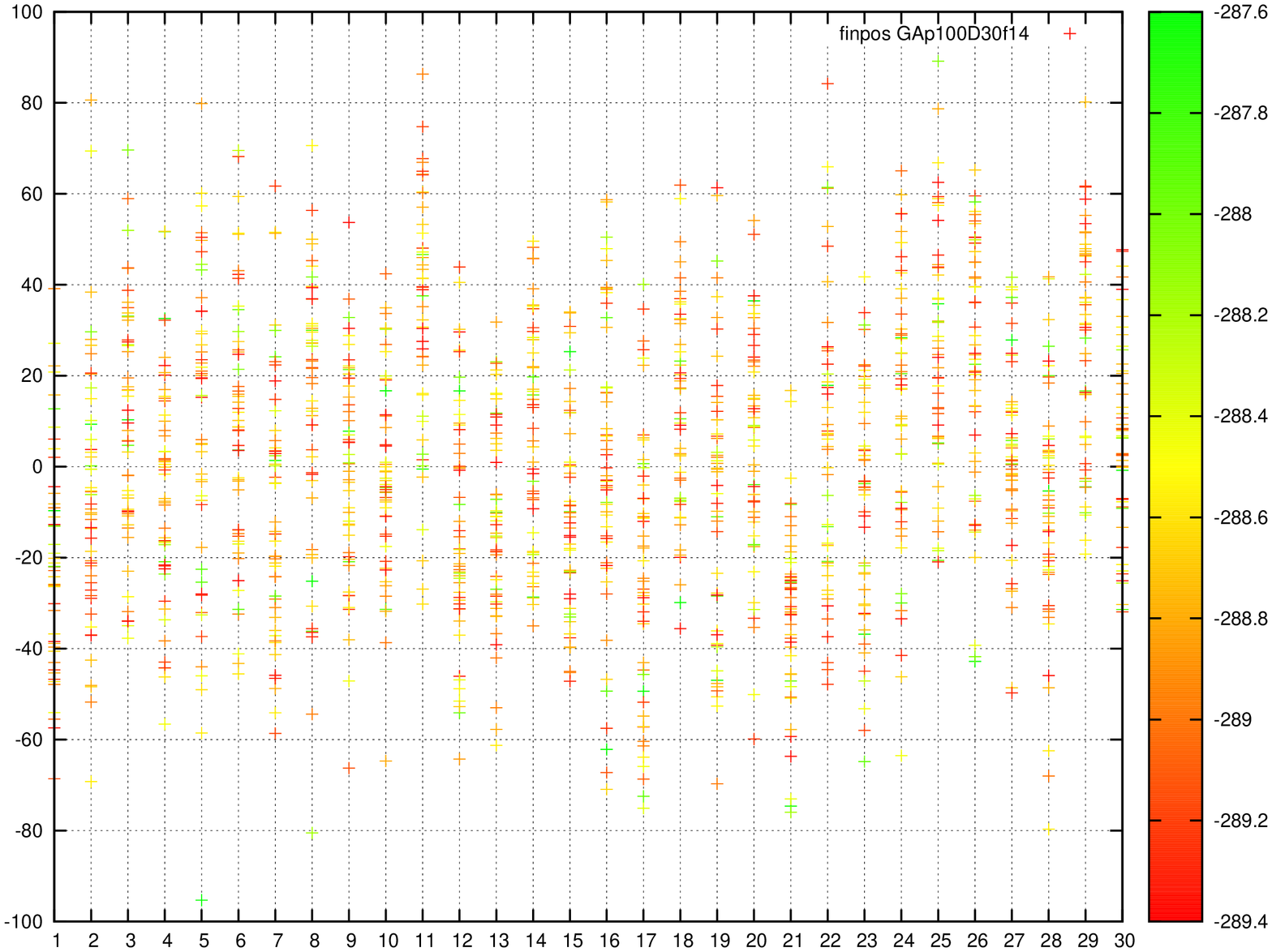}}
\subfigure[GA $N=100$ $f_{24}$]{\includegraphics[width=.34\linewidth,angle=270]{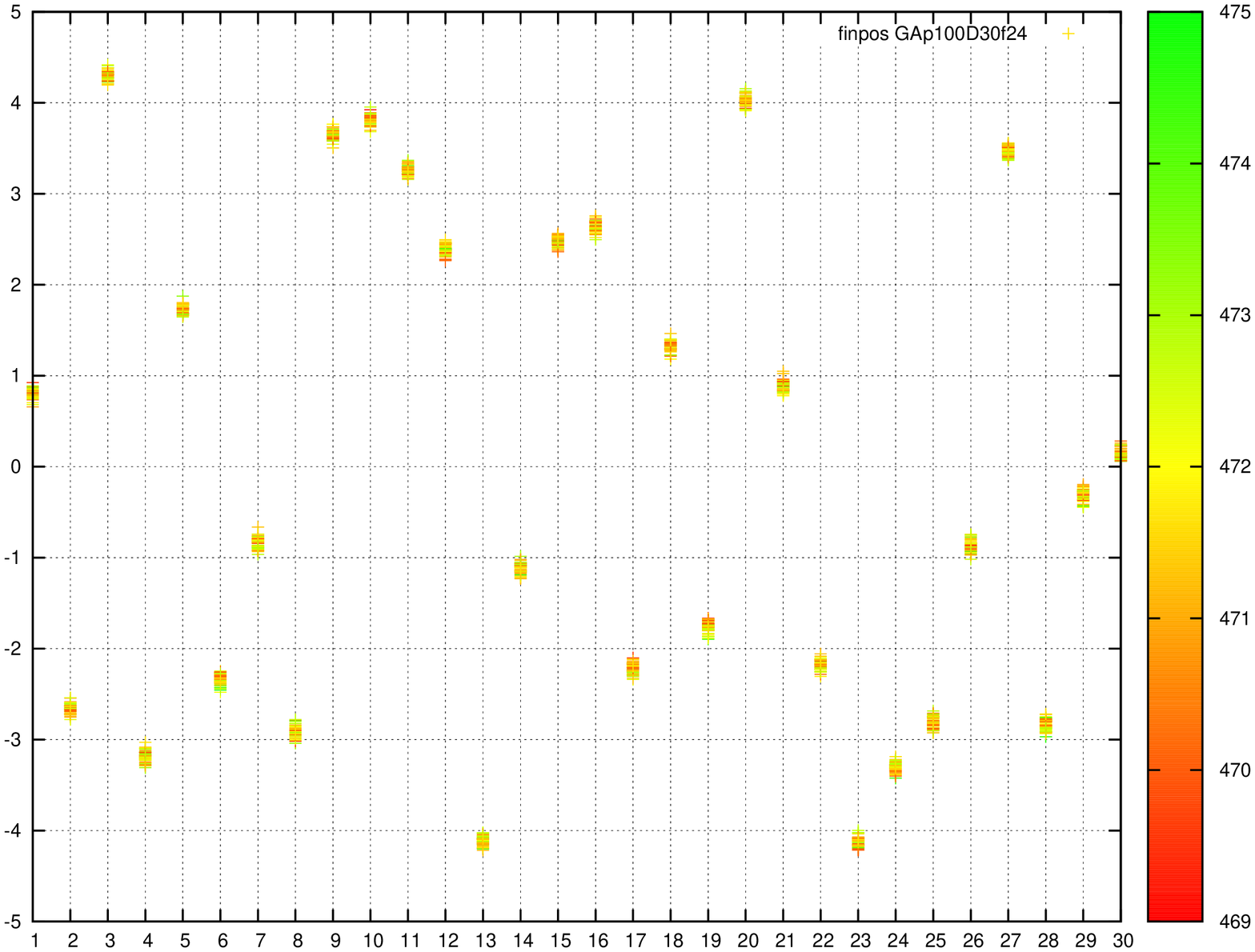}}
\caption{Positions of points with the best fitness in the last population of 50 runs of the considered GA for different population sizes in parallel coordinates - \textit{$f_{14}$ and $f_{24}$ are highly sensitive to structural bias of GA}. Horizontal axis shows the ordinal number of coordinate, vertical axis shows the full range of domain in this coordinate kept constant for each function; fitness value of each point is shown in colour.}\label{fig:GAf14f24pcoors}
\end{figure*}

Parallel coordinates visualisations of our genetic algorithm's results on these functions are shown in Figs. \ref{fig:GAf8f9pcoors}, \ref{fig:GAf13f21pcoors}, \ref{fig:GAf14f24pcoors}. Each individual plot summarises the results of 50 independent trials of the genetic algorithm on the function concerned, by showing, in parallel coordinates fashion, 50 30-dimensional points, comprising the best point reached in each trial. 
It is important to stress that interpretation of these figures should be \textit{entirely different} from interpretation of those shown in Figs. \ref{fig:GApcoors} and \ref{fig:PSOpcoors}. Unless specifically constructed so, it is not expected that the final distribution of positions of the best points in the final generation is close to uniform in the case of any function other than $f_0$. In sharp contrast to the $f_0$ results, we would naturally expect in these plots to see a strong effect due to the combined activity of selection and the shape of the landscape, resulting in the identification of regions at or close to optima of the objective function.  
 
At any point in time during the optimisation process, two forces can be conceptualized which simultaneously act on the population - landscape bias and structural bias. The first force pulls the population towards better values of the objective function, meanwhile the second force can be thought of as pulling the population towards 'attractors' in the domain (perhaps complex attractors) whose nature arises from the combination of algorithm design choices. Both of these forces are unknown and, therefore, their sum -- which defines population movement -- is also unknown. The use of $f_0$ to help quantify structural bias is precisely based on the idea of eliminating one of the unknowns, the 'landscape force', hence revealing any structural bias. It follows that, to interpret the visualisations of Figs. \ref{fig:GAf8f9pcoors}, \ref{fig:GAf13f21pcoors}, \ref{fig:GAf14f24pcoors}, we can proceed as follows. For a given objective function, we can observe how the distribution of final points varies as a function of the algorithm configurations considered, and consider how this correlates with the relative degrees of structural bias previously observed (via experiments with $f_0$) over the same set of configurations. Obviously, attention should also be paid to the final attained values of the objective function and their variances.

In such analysis of results over the CEC 2005 benchmark suite, we have observed three types of behaviour, which we conceptualise as resulting from the combination of structural bias in the algorithm configuration itself combined with more or less sensitivity to that bias inherent in the objective function at hand:
\begin{itemize}
\item \textit{Sensitivity} to structural bias, as exemplified by $f_8$ and $f_9$, see Fig. \ref{fig:GAf8f9pcoors}. For $f_8$, all three series of runs attained similar values of final fitness over $50$ runs, but runs with larger populations failed to find good solutions closer to the boundary of the domain. We attribute such failures to structural bias of the genetic algorithm, as also observed for $N=50$ on $f_0$. On $f_8$, the genetic algorithm exhibits behaviour overall similar to the case of $f_0$. Meanwhile, for $f_9$, final fitness values are quite different across the three series of runs, but the variance of positions of final best points demonstrates the pattern of sensitivity to structural bias.
\item \textit{Insensitivity} to structural bias, as exemplified by $f_{13}$ and $f_{21}$, see Fig. \ref{fig:GAf13f21pcoors}. All parameter settings considered lead to similar results in terms of the fitness values attained, positions of final best points and variances in their positions.
\item \textit{High sensitivity} to structural bias, as exemplified by $f_{14}$ and $f_{24}$, see Fig. \ref{fig:GAf14f24pcoors}. In the case of $f_{14}$, quite similar values of final fitness are attained over the three series of runs; however, drastic changes are clear from one series to another in terms of variances of positions of final best points. For $N=5$, these points fill the whole domain and better points, indicated on the figure with red markers, are uniformly spread out across the domain. The situation is to some extent similar for $N=20$ but all points start to shift towards the middle of the interval and those with better fitness values in particular. For $N=100$, no final best points are located in the outer regions of the domain, but their distribution in the centre of the domain is rather uniform. As for $f_{24}$, there are drastic changes both in terms of final fitness values and distribution of positions of final best points. It is interesting to note that for the genetic algorithm with $N=100$ on $f_{24}$, it is rather easy to find a region with low fitness values consistently over the series of $50$ runs suggesting that this particular function possesses a special property of some kind.
\end{itemize}

As regards other functions from the CEC2005 suite, it is worth mentioning that functions in the top of the list tend to be less sensitive to the structural bias of our genetic algorithm. These functions are known to be unimodal or close to unimodal \cite{cit:Suganthan2005}. This observation aligns well with our speculation in Section \ref{sect:sbiasGA}, and suggests that the theoretical analysis of the simplified genetic algorithm may have captured at least part of the essence of the factors that underpin structural bias and also the sensitivity to structural bias of any given objective function (via informed expectations of how the landscape may affect population variance). However, we are of course very much only at the beginning of a theoretical understanding of structural bias, in terms of both underpinning causes and of the effects of particular landscapes. This state of affairs goes hand in hand with a need for approaches to investigate and quantify inherent structural bias, such as proposed in this paper. Returning to the relative sensitivity to structural bias of different objective functions, we speculate that further work, involving analyses of particular collections of objective functions, might reveal similarities in the structure of basins of attraction in the landscapes might correlate with similarities in sensitivity. Empirically, we have seen that evolving populations seems to be less "confused" by structural bias in stronger regions of attraction which characterise unimodal optimisation as opposed to a weaker pull from multiple closer regions of attraction in the multimodal situation. This also points towards similarities between the effects of structural bias and noise in the objective function.  Just as a noisy objective function induces false optima in the landscape, structural bias deceptively pushes the evolving population towards regions potentially unremarkable in terms of objective function values.

\section{Discussion and Conclusions}\label{sect:conc}
A vast body of research in the field of population-based optimization algorithms deals with efficient exploitation of information already contained within the population, while little attention is paid to investigation of whether or not a specific combination of algorithmic operators and algorithm strategies is actually capable of reaching all parts of the search space with equal efficiency. When an algorithm is \textit{not} capable in the latter sense - that is, when an algorithm favours certain areas of the search space over others, independently of the fitness function, it is exhibiting 'structural bias'. 

In this paper we have argued, from both theoretical and empirical standpoints, that structural bias is likely to be common in practice, and amplified when we would least expect it (when we increase the population size in hope of a more exploratory search) and when it may cause most damage (on 'difficult' problems). When faced with the problem of optimising a given function, the amount of information usually available regarding its properties and landscape features is highly limited. Typically, for example, one has no prior information at all concerning where in the search space the optima may be. Therefore, one wishes to design an algorithm capable of locating the optima no matter where exactly they are in the search space. This implies that the variation operators of the algorithm must be able to, first of all, reach every region of the search space and, second, ideally, do so with no bias towards any particular region.

It is helpful to think of this issue in terms of intuitive 'forces' that act on the population. Population-based optimisation algorithms can all be regarded as sophisticated variants of 'generate-and-test' algorithms; the 'test' is effected by the fitness function, and provides information that the algorithm uses to guide its navigation of the landscape; meanwhile, 'generate' refers to the production of new candidate solutions, and is achieved by the algorithm's suite of operators. In very broad terms, we conceptualised population dynamics as the product of  a 'landscape bias force' and a 'structural bias force', respectively representing the influences of the fitness function itself, and the algorithm's design. To empirically investigate structural bias in this paper, we effectively neutralised the 'landscape force' by performing optimisation experiments with $f_{0}$ (as defined in Section \ref{sect:sbias}). This enabled us, by visualising the results of multiple experiments, to observe the 'structural bias' force in action by observing its effects on the distribution of the final best points. Since we would expect these distributions to be uniform in the absence of structural bias, the pattern of observed non-uniformity, in combination with other considerations, can be taken as informative of the structural bias inherent in the optimisation algorithm. 

To some extent, one can posit an alternative explanation for such effects by appeal to undesirable properties of the pseudo-random number generator used. Such an objection can be difficult to discount, since the effects of the complex ways in which pseudo-random number generators are indeed non-random are extremely difficult to predict. Nevertheless, our analysis in Section \ref{sect:random}, coupled with our theoretical findings and the pattern of empirical results, suggest that this approach reveals structural bias rather than pseudo-random artefacts.

Our approach to revealing and quantifying structural bias is easily replicated, and we recommend its use to investigate the structural bias that may be inherent in any instantiated optimization algorithm, prior to finalising the parameteric and design configuration of that algorithm to be deployed on real-world problems. Such investigation of structural bias can be seen as an additional 'validation' step, coupled with other investigations of the algorithm design which would normally be done to reveal the configuration that provides the best and/or fastest solutions to the class of problems of interest. It is interesting to speculate on the consequences of such validation in different circumstances. If the 'best' algorithm configuration also has minimal or no structural bias, then all is well. However this may very often not be the case. When optimisation is used for system identification problems (for example, determining the parameters of a function or model that best fit a set of empirically obtained points), it is usually deemed important to find, as far as possible, 'all' good solutions, or a fair representation thereof. If the otherwise preferred configuration exhibits structural bias, the 'true' system parameters may not be uniformly accessible to the algorithm. 

We have also contributed a theoretical argument that partially explains how structural bias can arise in a simple population-based algorithm. The analysed algorithm is simplified, but exhibits the primary strategies common to almost all population based optimization algorithms, including a parameter $d$ that controls the degree of exploration induced by the variation operator -- the larger $d$, the higher the chance and extent to which a new sample will extend beyond the region of search space occupied by its parents. The crucial step in the argument is to show that, on average, and under certain conditions, the population variance will decrease with time, despite the clear opportunities for search to extend beyond the current locations of the population. 

If we consider a truly random algorithm $RA$ in such circumstances, in which each new sample is generated uniformly at random in the search space and replaces a randomly chosen previous sample, we can expect unbiased coverage of the search space and maintenance of a constant variance over time, which (under the conditions of the theorem) would be $\frac{1}{12}$. For typical choices of parameters ($\sigma^2 = 0.1, N = 50, d = 0.2$), the value of $K$ in the theorem is much lower than $\frac{1}{12}$, suggesting that such an algorithm will rarely be able to maintain the levels of exploration required to eliminate structural bias without careful design. Algorithm $RA$ exhibits 'pure' exploration, however any effective optimization algorithm incorporates exploitation, which is invariably achieved by biassing samples towards the regions of previously visited points. The 'reducing variance' theorem suggests that such exploitation is intimately related to the emergence of structural bias, but it also suggests that the latter can be controlled by reducing the population size, or by raising $d$ (or, alternatively, by revisiting the algorithm's design to introduce mechanisms that introduce additional new samples in a way that is not tied to the locations of previous samples). By raising $d$, we (usually) increase the likelihood of structural bias but reduce the efficiency of exploitation; meanwhile, by reducing the population size we reduce the likelihood of structural bias (recall: in the context of a genetic algorithm not too distant from the theoretically analysed version) but reduce the level of exploration. 

The inverse relationship between structural bias and population size (strictly in the context of standard and simple genetic algorithms, which was the substrate of our theoretical analysis) that is at first counter-intuitive - to increase the population size would seem to inject more diversity, which we should expect to alleviate such bias. However, we believe this phenomenon can be explained by an effect akin to 'preferential attachment' in the evolution of complex systems. Structural bias, manifested as the concentration of search progress in ever narrower regions, builds on initial seed areas which begin to attract further points. In our context, if two parent solutions happen to be close together, their offspring will stay nearby and increase the density in this region, and the positive feedback dynamics of this process will exacerbate the non-uniform distribution. When the population size is increased, there is more opportunity for such initial seeds to be present. 

The results we have presented seem less surprising when iterative population-based algorithms are considered in relation to \textit{Iterated Function Systems} which, by definition, are finite sets of mappings of complete metric space or, symbolically, $\mathcal{F}=(\mathds{X},f_1,f_2,...,F_M), f_m:\mathds{X}\to\mathds{X}, m=1,2,...,M$. Depending on the properties of functions that make them up, IFSs exhibit a variety of behaviours. According to the collage theorem \cite{cit:Barnsley1988}, for any given set/image there exists a strictly contractive IFS whose attractor arbitrarily closely approximate this set/image. A linear IFS on $\mathbb{R}^n$ has a unique attractor located at the origin \cite{cit:Barnsley2011_algebra}. Any projective IFS has at most one attractor \cite{cit:Barnsley2012} but behaviour of such attractors appears to be more complicated than in the case of affine IFSs as they might not depend continuously on parameters \cite{cit:Barnsley2012}. Moreover, there are examples of non-contractive projective IFS with an attractor \cite{cit:Barnsley2012}. These results point to a potentially fruitful direction for the analysis of algorithms through studying the properties of their operators. 

Finally, it is instructive to speculate on the existence of structural bias in combinatorial optimization. Both the theoretical and practical investigations in this article are pinned to the context of real-valued vector optimization. In the empirical tests, we have observed structural bias in terms of the distribution of points in a continuous space, and in theory we have related its emergence to dynamics of positional variance in this space as a result of the operation of typical real-valued operators. So, at first sight, it is not at all obvious that structural bias may occur in the combinatorial case. However, it is trivial to see that it \textit{could} occur. For example, were we so inclined, we could purposely design operators to favour certain regions of the space independently of fitness. Imagine, say, a permutation space with an even number of objects, in which the only operator in use was to swap an item with its neighbour two steps away; this search is then confined to the cross-product of two subspaces, omitting most of the permutation space. Also, despite the 'real-value' focus of the theoretical argument, it is intuitively reasonable to speculate that a similar argument, couched in terms of suitable metrics, may be meaningful for combinatorial spaces. For example, the perturbation effect of a combinatorial operator on one or more points can be characterised as a distribution of edit distances from those points. Structural bias in combinatorial search algorithms might arise from the dynamics of the variance of this distribution in the context of other aspects of the algorithm's configuration.

\bibliographystyle{ieeetr}
\bibliography{refs}
\end{document}